\documentclass[12 pt]{article}
\usepackage{float}
\usepackage[pdftex]{graphicx}
\usepackage{amsmath}
\usepackage{psfig}
\usepackage{epic}
\usepackage{eepic}
\usepackage{headerfooter}
\usepackage{subfigure}
\usepackage{enumitem}
\usepackage{multicol}
\usepackage{color}
\newcommand{\debug}[1]{}              
\newcommand{\cmd}[1]{}

\newcommand{\daddcontentsline}[3]{\addcontentsline{#1}{#2}{#3}}

\newcommand{\iB}{\begin{itemize}}
\newcommand{\iE}{\end{itemize}}
\newcommand{\eB}{\begin{enumerate}}
\newcommand{\eE}{\end{enumerate}}
\newcommand{\dB}{\begin{description}%
}

\newcommand{\sname}{}

\newcommand{\tname}{}
\newcommand{\irefx}[1]{{\it \ref{#1}}\debug{[#1]}}
\newcommand{\TB}[1]{%
	\renewcommand{\tname}{\sname #1}%
	\daddcontentsline{lot}{table}{\debug{Theorem%
	\thetheorem \fbox{\sname #1}}}\begin{theorem}\slabelx{#1}%
	\cmd{TB} \  }
\newcommand{\TE}{\end{theorem}%
	\cmd{TE} }
\newcommand{\LB}[1]{%
	\renewcommand{\tname}{\sname #1}%
	\daddcontentsline{lot}{table}{\debug{Lemma \thelemma
	\fbox{\sname #1}}}\begin{lemma}\slabelx{#1}%
	\cmd{LB} \  }
\newcommand{\LE}{\end{lemma} %
	\cmd{ LE } }

\newcommand{\dE}{\end{description}}
\newcommand{\hB}{\begin{alphlist}}
\newcommand{\hE}{\end{alphlist}}
\newcommand{\hBa}{\begin{alphlista}}
\newcommand{\hEa}{\end{alphlista}}

\newcommand{\DefB}[1]{%
	\renewcommand{\tname}{\sname #1}%
    \daddcontentsline{lot}{table}{\debug{Definition%
    \thedefinition \fbox{\sname #1}}}\begin{definition}\slabelx{#1}%
	\cmd{DefB} \ }
\newcommand{\PB}{{\em Proof:\/\ }\cmd{ PB} }
\newcommand{\PE}{{\hfill $\Box$}{\cmd{PE}}}
\newcommand{\EB}{\begin{equation}\cmd{EB}}
\newcommand{\EE}[1]{ \debug{\fbox{\sname #1}}%
	\label{\sname #1} \end{equation}%
	\cmd{EE} }
\newcommand{\titem}[1]{\item  \cmd{titem} \tlabel{#1} }
\newcommand{\dlabel}[1]{\debug{\fbox{\tiny #1}}\cmd{dlabel}\label{#1}}
\newcommand{\dcite}[1]{\cite{#1}\debug{[#1]}\cmd{dcite}}
\newcommand{\dref}[1]{\ref{#1}\debug{[#1]}\cmd{dref}}
\newcommand{\tref}[1]{\irefx{\tname #1})\cmd{tref}}

\newcommand{\tlabel}[1]{\debug{\fbox{\tiny \tname #1}}%
	\label{\tname #1}\cmd{tlabel}}
\newcounter{ctr}
\renewcommand{\thectr}{\alph{ctr}}
\newenvironment{alphlist}{%
\begin{list}{\thectr)}{\usecounter{ctr}%
\topsep=0pt}%
}%
{\end{list}}
\newenvironment{alphlista}{%
\begin{list}{\thectr)}{\usecounter{ctr}%
\itemsep = -4.0pt \topsep=0pt}%
}%
{\end{list}}
\newcommand{\slabelx}[1]{\debug{\fbox{\tiny \sname #1}}%
	\label{\sname #1}}

\newcommand{\msec}[2]{\section[#1
	\debug{\fbox {#2}}]{#1 \cmd{msec} \dlabel{#2}}%
	\markboth{\today}{Sec. \thesection}}
\newcommand{\msubsection}[2]{\subsection[#1 \debug{\fbox {#2}}]
	{#1 \cmd{msubsection} \dlabel{#2}}%
	 \markboth{\today}{Sec. \thesection}}

\renewcommand{\PE}{{\hfill qed}{\cmd{PE}}}

\renewcommand{\dB}{\begin{description}}

\setlist{leftmargin=5.5mm}

\newtheorem{theorem}{Theorem}
\newtheorem{lemma}{Lemma}
\newtheorem{definition}{Definition}

\newcommand{\mathbb}{\textbf}

\begin{document}

\title{ Stochastic Broadcast Control of Multi-Agent Swarms}

\author{Ilana Segall  and Alfred Bruckstein }

\date{Center for Intelligent Systems\\MultiAgent Robotic Systems (MARS) Lab\\Computer Science Department\\
Technion, Haifa 32000, Israel}

\maketitle

\newpage
\tableofcontents


%


\newpage


\begin{abstract} We present a model for controlling swarms of  mobile agents via  broadcast control, assumed to be detected  by a random set of agents in the swarm. The agents that detect the control signal become ad-hoc leaders of the swarm.   The agents are assumed to be velocity controlled, identical, anonymous, memoryless units with limited capabilities of sensing their neighborhood.   Each agent is programmed to behave according to a linear local gathering process, based on the relative position of all its neighbors.
The detected exogenous control, which is a desired velocity vector, is added by the leaders to the local gathering  control. The graph induced by the agents adjacency is referred to as the \emph{visibility graph}. We show that for piecewise constant system parameters and a connected visibility graph,  the swarm asymptotically aligns in each time-interval on a line in the direction of the exogenous control signal, and all the agents move with identical speed. These results hold for two models of pairwise influence in the gathering process, uniform and scaled. The impact of the influence model is mostly evident when the visibility graph is incomplete.  These results are conditioned by the preservation of the connectedness of the visibility graph. In the second part of the report we analyze sufficient conditions for preserving the connectedness of the visibility graph. We show that if the visibility graph is complete then certain  bounds on the control signal suffice to preserve the completeness of the graph. However, when the graph is incomplete,  general conditions, independent of the leaders topology, could not be found.\end{abstract}


\section{Introduction}\label{Introduction}

We present a system composed of a group or swarm of autonomous agents and a controller. All the agents behave according to a distributed gathering process, ensuring cohesion of the swarm, and the controller sends desired velocity controls  to the cloud. The signal sent by the controller is received by a random set of agents. If all the agents receive the signal then the cloud will move with the desired velocity.  If only part of the agents receive the signal then the cloud will move in the desired direction but with a fraction of the desired speed, depending on the topology of the inter-agent visibility graph.  This can be viewed as representing the "inertia" or the "reluctance of the cloud to move" in the desired direction with the desired speed. We investigate two models of neighbors influence in the local control, uniform and scaled. We show that if the visibility graph is complete, then the ratio of the achieved collective speed to the desired speed, for both influence models, is the ratio of the \emph{number} of leaders to the total number of agents. However, when the graph is incomplete, the ratio of the achieved collective speed to the desired speed is a function of the influence model. If the influence is uniform the ratio stays as before, i.e.  the ratio of the number of leaders to the total number of agents. but if the influence is scaled then the ratio of the achieved collective speed to the desired speed depends not only on the number of leaders but also on the exact topology of the visibility graph and on the location of the leaders within the graph. Hence, for the same number of leaders in the same incomplete visibility graph, with scaled influence, different results can be obtained for different leaders.

\subsection{Statement of problem}\label{Model}
We consider a system composed of $n$  homogeneous agents evolving in $\mathbb{R}^2$.  The agents are assumed to be homogenous,  memoryless, with limited visibility (myopic) and are modeled by single integrators, namely are velocity controlled.
     The visibility (sensing) zone of agent $i$ is a disc of radius $R$ around its location. Agents within the sensing zone of agent $i$ are referred to as the \emph{neighbors} of $i$.  If $j$ is a neighbor of $i$, we write $i \sim j$.  The set of neighbors of $i$,  define the \emph{neighborhood} of $i$, denoted by $N_i$.
The emergent behavior of agents with unlimited visibility and stochastic broadcast control was discussed in \dcite{SB2016}

\begin{figure}
\begin{center}
\includegraphics[scale=0.4]{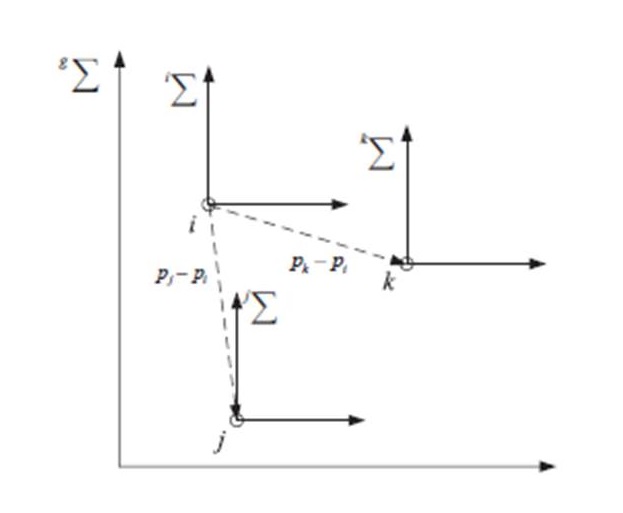}
\caption{Illustration of local and global reference frames alignment}\label{fig-frames}
\end{center}
\end{figure}

Each agent can measure only the \emph{relative position} of other agents in its own local coordinate system.  The orientation of all local coordinate systems is aligned to that of a global coordinate system, as illustrated in Fig.\dref{fig-frames}, i.e. agents are assumed to have compasses enabling them to align their local reference frames to a global reference frame.   Here $p_i=(x_i,y_i)$ represents the position of agent $i$ in the global reference frame, \emph{unknown to the agent itself}.

\begin{figure}
\begin{center}
\includegraphics[scale=0.5]{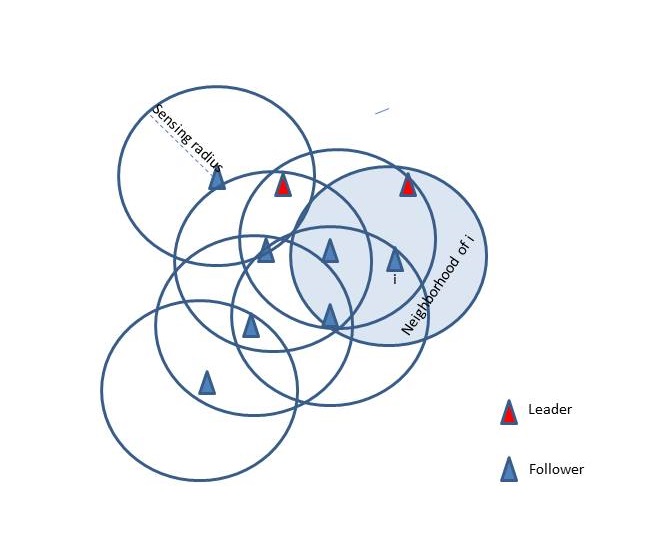}
\caption{Illustration of problem topology at a certain point in time}\label{topology}
\end{center}
\end{figure}

We assume that the agents do not have data transmission capabilities, but all the agents are capable of detecting an exogenous,  broadcast control.  At any time, a random set of agents detect the broadcast control.  These agents will be referred to as ad-hoc \emph{leaders}, while the remaining agents will be the \emph{followers}. The exogenous control, a velocity vector $u$, is common to all the leaders.  The agents are unaware of which of their neighbors are leaders.
 The setup of the problem  is illustrated in Fig. \dref{topology}.

 The sets of the leaders and of the followers are denoted by $N^l$, $N^f$ respectively.  The number of leaders and followers in the system is denoted by $n_l=|N^l|$, $n_f=|N^f|$ respectively.  The sum $n=n_f+n_l$ is the total number of agents. The agents are labeled $1,...,n$.

\msubsection{The dynamics of the agents}{DynModel}
In our model, the followers apply a local gathering control based on the relative position of all their neighbors and the leaders apply \emph{the same local control (\dref{gen-SelfDyn}) with the addition of the exogenous input $u$}.  In general, the strength of the influence of neighbor $j$ on the movement of agent $i$ is  some function $f(j,i)$, most often a function of the distance between $i$ and $j$, cf. \dcite{M-T}, \dcite{J-E}, \dcite{C-S}.
If we denote by $\sigma_{ji}$ the strength of the \emph{influence} of agent $j$ on the movement of agent $i$, then  we have:
\begin{itemize}
  \item for each $i \in N^f$
\begin{equation}\dlabel{gen-SelfDyn}
  \dot{p}_i(t)=\sum\limits_{j \sim i} \sigma_{ji}(t) (p_j(t) - p_i(t))
\end{equation}
where $p_i$ is the position of agent $i$
  \item  for $ i \in N^l$
\begin{equation}\dlabel{gen-LeadDyn}
  \dot{p}_i(t)=\sum\limits_{j \sim i} \sigma _{ji}(t) (p_j(t) - p_i(t)) + u
\end{equation}
\end{itemize}

\begin{figure}
\begin{center}
\includegraphics[scale=0.4]{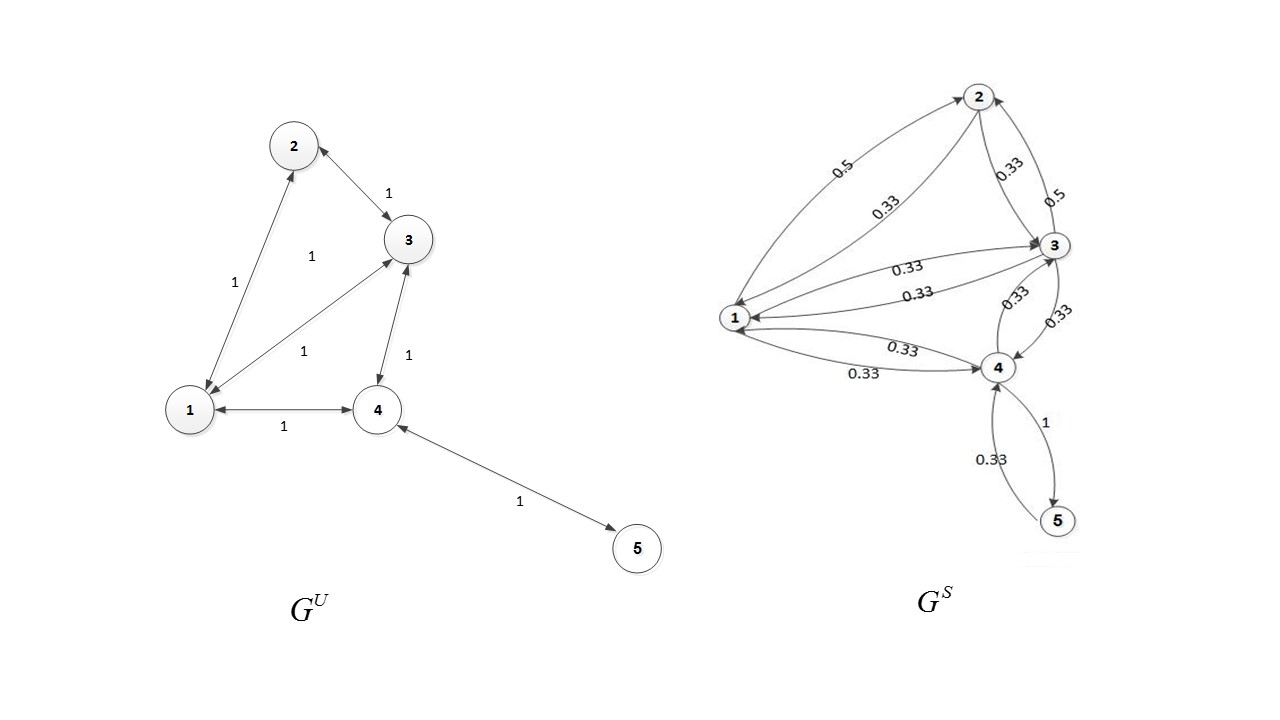}
\caption{Illustration of an interactions graphs with uniform vs scaled influence }\label{UniScaled}
\end{center}
\end{figure}

 We consider two cases of influence :
\begin{enumerate}
  \item \emph{Uniform} -  The influence of all neighbors on any agent is identical and time independent, i.e. $\sigma_{ji} (t) =1 \text{    } \forall j \in N_i (t)$.
  \item \emph{Scaled} - The influence of an agent $j \in N_i(t)$ on $i$ is scaled by the size of the neighborhood $N_i(t)$, i.e. for each $i$, we have $  \sigma_{ji}(t) = \frac{1}{|N_i (t)|} ; \forall j \in N_i (t)$. \end{enumerate}
 Fig. \dref{UniScaled} illustrates an example of pairwise interaction graph with uniform influences, denoted by $G^U$, vs the corresponding graph with scaled influences, denoted by $G^S$.

 In this report, we derive the emergent behavior of agents with any visibility graph, complete or incomplete, applying protocol (\dref{gen-SelfDyn}), (\dref{gen-LeadDyn}) for followers and leaders, for both influence models. We show that when  the visibility graph is complete (due to a very large $R$) the two influence models will move the swarm with the same velocity to the same asymptotic (moving) gathering point but when the graph is incomplete the two influence models affect differently the collective velocity and asymptotic state of the swarm.

 Since $p_i(t) = [x_i(t) \quad y_i(t)]^T$ and  $u =[u_x \quad u_y]^T$  and assuming that $x_i(t)$ and $y_i(t)$ are decoupled we can write

\begin{eqnarray}
  \dot{x}_i(t)& =  & \sum\limits_{j \in N_i}\sigma_{ji} (x_j(t) - x_i(t)) + b_i u_x\dlabel{x-1D}\\
  \dot{y}_i(t)& =  & \sum\limits_{j \in N_i}\sigma_{ji} (y_j(t) - y_i(t)) + b_i u_y\dlabel{y-1D}
\end{eqnarray}
and consider $x_i(t)$ and $y_i(t)$ separately, as one dimensional dynamics,  (cf. Section \dref{Gen-1D}).

\begin{equation}\label{bi}
  b_i = \begin{cases}1 ; \text{    if    } i \in N^l \\
  0 ; \text{   otherwise}
  \end{cases}
\end{equation}
where $N^l$ is the set of leaders.

In the piecewise constant case, when the time-line can be divided into intervals in which the  system evolves as a linear time-independent dynamic system, (\dref{x-1D}) can be written in vector form as
\begin{equation}\dlabel{x-1D-vec}
\dot{x}( t )=-L_k\cdot x(t)+ B_k u_x(t_k)
\end{equation}
and similarly for (\dref{y-1D}), where
\begin{itemize}
  \item $t \in [t_k \quad t_{k+1} )$
  \item $t_k$ is a switching point, i.e. the time when either the visibility graph , the leaders or the exogenous control change
  \item $L_k$ is the  Laplacian associated with the interactions graph $G_k$,  either uniform or scaled, in the interval $[t_k \quad t_{k+1} )$
  \item $B_k$ is a leaders indicator in the interval $[t_k \quad t_{k+1} )$ , i.e. a vector of dimension $n$ with $0$ entries in places corresponding to the followers and $1$ in those corresponding to the leaders
  \item $u_x(t_k)$ is the $x$ - component of the exogenous control $u$ in the interval $[t_k \quad t_{k+1} )$
  \item $L_k, B_k, u_k $ are constant
\end{itemize}
The emergent behavior in the interval $t \in [t_k, t_{k+1})$ is a function of the corresponding properties of $L_k$. We show in the sequel that if the influence is uniform then the corresponding Laplacian is symmetric and its properties are independent of the topology of the graph but if the influence is scaled then  the Laplacian corresponding to an incomplete graph is non-symmetric while the Laplacian corresponding to  a complete graph is symmetric with the corresponding change in properties.

 In the sequel we treat each such interval separately, and thus it is convenient to suppress the subscript $k$.
We first assume $G_k$ to be strongly connected for all $k$. In the second part of the report, we show scenarios and conditions for never losing friends, i.e. for  $G_k \subseteq G_{k+1};  \quad \forall k$. We show that if $G_k$ is complete then bounding $|u_k|$ suffices to ensure that it remains complete. However, if $G_k$ is incomplete, the conditions are tightly related to the graph topology and could be derived only for specific cases.

\textbf{\textit{Note}} that losing visibility to a neighbor does not necessarily mean losing connectivity. However, never losing neighbors ensures never losing connectivity.

\msubsection{Literature survey and contribution}{survey}
Many ways of controlling the collective behavior of self-organized multi-agent systems by means of one or more \textbf{special agents}, referred to as \emph{leaders} or \emph{shills}, have been investigated in recent years.   We will be grouping the surveyed work in several broad categories and indicate the novelty of our model as compared to each.

\begin{enumerate}
   \item \underline{Leaders that do not abide by the agreement protocol}\\ These leaders are pre-designated and their state value is fixed at a desired value. Jadbabaie et al. in  \dcite{JLM} consider Vicsek's discrete model \dcite{VC95}, and introduce a leader that moves  with a fixed heading  .  Tanner, Rahmani, Mesbahi and others in  \dcite{Tanner}, \dcite{Rahmani}, \dcite{Rahmani2}, \dcite{RM2}, etc. consider static leaders  (sometimes named "anchors") and show conditions on the topology that will ensure the controllability of the group. A system is controllable if for any initial state there exists a control input that transfers any initial state to any final state in finite time. Our model differs from the above in that the leaders are neither pre-designated nor static. The number of leaders and their identity is arbitrary. They do not ignore the agreement protocol, but rather add the received exogenous control to the computed local rule of motion and move accordingly.  We do not require the system to reach a pre-defined final state.  Our aim is to steer the swarm in a desired direction.  We show the emergent dynamics for a desired velocity sent by a controller and received by random  agents in the swarm.

\item \underline{Leaders combining the consensus protocol with goal attraction}\\In \dcite{DGEH}, \dcite{GDEH09}, the exogenous control is a goal position, known only to the leaders. The dynamics of all agents, leaders or followers, is based on the consensus protocol.  For leaders however, it includes an additional goal attraction term which aims at leading the team to the \textbf{pre-defined goal position}. The attraction term is a function of the leader's distance from the goal position, therefore varies from leader to leader.
This approach  is the closest to our model that we have found in the surveyed literature, but some major differences exist.
In our model the exogenous control is not a goal position but a velocity vector, $u$, \textbf{common to all leaders}. Moreover, agents are not aware of their own position, but only of their relative position to their neighbors.  We show that with our model, the agents, rather than gathering at a goal position, \emph{asymptotically align along  a line  in the direction of $u$ and move with identical speed}.

\item \underline{Shills} - Intelligent agents with on-line state information of regular units.
Han, Guo and Li, \dcite{HLG2006}, followed by Wang, \dcite{HW2013} introduced the notions of shill and soft control. Shills are special agents \textbf{added} to the swarm with the purpose of controlling the collective behavior. They are the exogenously controlled part of the system. The basic local rules of motion of the existing agents in the system are not changed.  The existing agents treat the special agent as an ordinary agent, thus enabling it to  "cheat" or "seduce" its neighbors towards the desired goals. These special agents are called "shills" \footnote{Shill is a decoy who acts as an enthusiastic customer in order to stimulate the participation of others}.
As opposed to the above, in our work we study the emergent collective behavior when probabilistically selected agents, out of the existing agents, receive an exogenous control $u$. These agents become the ad-hoc leaders. The number of leaders is not predetermined, hence can be any number from $1$ to $n$. Also, we do not design $u$ in order to obtain some desired final state. Moreover, in our model the leaders do not have an entirely stand-alone control rule. All agents follow the same rule of motion, with the addition of the exogenous control, when received, i.e. while leaders. Leaders do not have on-line state information of other agents. The only available information, for leaders and all other agents, is relative position to neighbors.

\item \underline{Broadcast control}\\
Recently Azuma, Yoshimura and Sugie \dcite{AYS2013} have proposed a broadcast control framework for multi-agent coordination, but in their model the control is assumed to be received by all units, i.e. there are \emph{no followers}. In this model the global controller observes the group performance, designs the information to be broadcast and sends a signal, received by all, to govern the group behavior. The agents set the local control, based on the received signal.
As opposed to the above, in our model the broadcast control is the goal velocity vector, aiming  to steer the swarm in some desired direction with desired speed. The detailed group performance is not directly observed by the controller, therefore the broadcast control does not depend on it.  Moreover, not all units necessarily receive the broadcast control, but at least one does.

\end{enumerate}

\msubsection{Paper outline}{Outline}
We derive the collective swarm behavior for piecewise constant systemד. We first treat each time interval separately, as a  \emph{time-independent  system } over an interval $[0,t)$.      Section \dref{Gen-1D} presents the one dimensional case which is readily extended to two dimensions in Section \dref{LTI-dyn}.
In Section \dref{SimEx} we show simulation results, illustrating the two dimensional swarm behavior over a single time interval.
In Section \dref{MultiIntEx} we extend the investigation of one interval to multiple intervals, where new intervals are triggered by changes in the exogenous control, $u$, in leaders or in the visibility graph.  We assume that $u$ and the leaders change randomly, but the visibility graph is state dependent, therefore, when the visibility is limited, the system may disconnect. In Section \dref{never-losing2} we derive conditions for a complete visibility graph to remain complete and in Section \dref{LimitedComplete-Ex} we illustrate the effect of the derived bounds. In Section \dref{LimitedIncomplete} we derive conditions for never losing friends, when the visibility graph is incomplete, and show that these depend on the exact, time-dependent, topology.
We conclude in section \dref{future} with a short summary and directions for future research.

\msec{One dimensional group dynamics}{Gen-1D}
In this Section we consider a one dimensional piecewise constant system, (\dref{x-1D-vec}). In the sequel we threat each time interval, $[t_k \quad t_{k+1} )$, separately. Thus, it is convenient to suppress the subscript $k$.  Moreover, it is convenient to denote by $t$  the relative time since the beginning of the interval ($t=0$) and  by $x(0)$  the state of the system at this time.

We then have (in each interval)
\begin{equation}\dlabel{1D-vec}
\dot{x}( t )=-L\cdot x(t)+ B u
\end{equation}

Eq. (\dref{1D-vec}) has the well known solution (ref. \dcite{TK})
\begin{equation}\dlabel{eq-pieceDyn}
x(t) = e^{-L t} x(0) + \int_{0}^{t} e^{-L(t-\tau)} B u \mathrm{d}\tau
\end{equation}

Eq. (\dref{eq-pieceDyn}) can be rewritten as
\begin{equation}\dlabel{eq-xdyn2}
x(t) = x^{(h)}(t) + x^{(u)}(t)
\end{equation}
where
 \begin{itemize}
   \item $x^{(h)}(t)= e^{-L t} x(0)$ represents the zero input solution
   \item $x^{(u)}(t)=\int_{0}^{t} e^{-L (t-\tau)} B u \mathrm{d}\tau$ represents the contribution of the exogenous input to the group dynamics
 \end{itemize}

\msubsection{Definitions}{Def}

\begin{itemize}
  \item $G^U$ an undirected graph of uniform  interactions, with vertices labeled $1,...,n$.
  \item ${{d}_{i}}$ the number of neighbors of  vertex $i \in G^U$, i.e. the degree of $i$
  \item $\Delta $ the degree matrix of the graph $G^U$, a diagonal matrix with elements $\Delta_{ii}={{d}_{i}}$,
  \item $A^U$ the adjacency matrix of $G^U$, a symmetric matrix with 0,1 elements, such that

\begin{equation*}
A^U_{ij} =
\begin{cases}
1 & \text{if } i \sim{\ } j
\\
0 & \text{otherwise }
\\
\end{cases}
\end{equation*}
  \item $L^U = L(G^U)$ the Laplacian representing $G^U$, is defined by
\begin{equation}\dlabel{def-Lu}
  L^U=\Delta -A^U
\end{equation}\dlabel{Def-Gamma}
  \item $\Gamma$ the normalized Laplacian of $G^U$, is defined by
  \begin{equation}\dlabel{def-Gamma}
  \Gamma= \Delta^{-1/2} L^U \Delta^{-1/2}
\end{equation}
  \item $G^S$ the directed graph of scaled interactions corresponding to $G^U$
  \item $L^S=L(G^S)$ the Laplacian representing $G^S$
  \begin{equation}\dlabel{def-Ls}
    L^S=\Delta^{-1} L^U
  \end{equation}
\end{itemize}

\emph{\underline{Note}} that Eq. (\dref{1D-vec}) and its general solution (\dref{eq-pieceDyn}) hold for both  $L^U$ or  $L^S$.

In the following Sections, we  develop explicit solutions for each case and investigate their properties.

\msubsection{Zero input group dynamics }{ZeroInp}
Denote by $L$ the Laplacian associated with the time-independent visibility graph, in the time interval.
 The zero input group dynamics is given by
\begin{equation}\dlabel{eq-agree}
  \dot{x}^{(h)}( t )=-L\cdot x^{(h)}(t)
\end{equation}
We will show that for both $L=L^U$ and $L=L^S$, representing Laplacians of connected graphs and strongly connected digraphs respectively, the solution of eq. (\dref{eq-agree}) converges asymptotically to a  consensus state, namely $x^{(h)}_i=x^{(h)}_j=\alpha; \forall i,j, i\neq j$ (cf. Proposition 2 in \dcite{OS-M2003}).\\
Since we consider each interval separately and $t$ is the time elapsed from the beginning of the interval, by "asymptotic state" we mean here the value of the state for large $t$.
The value of the consensus state $\alpha$, in each interval, is obtained by explicitly calculating $exp(-Lt)$, for large $t$, as described below.

\subsubsection{Uniform influence - Symmetric Laplacian  $L^U$}
\LB{L-uniConverge}

The value of the consensus state for an undirected, connected, interactions graph with corresponding  Laplacian, $L^U$, is the average of the initial states.
\LE
\PB\\
Using the properties of $L^U$ (cf. Appendix \dref{Graphs}), namely that:
\begin{itemize}
  \item $L^U$ is a real symmetric positive semi-definite matrix
  \item all the eigenvalues of $L^U$, denoted by $\lambda^U_i$ are real and non-negative.
  \item if $G^U$ is connected then there is a single zero eigenvalue, denoted by $\lambda^U_1$  and the remaining eigenvalues are strictly positive.
  \item we can always select $n$ real orthonormal eigenvectors of $L^U$, denoted by $V^U_i$, where $V^U_i$ is the (right) eigenvector  corresponding to eigenvalue $\lambda^U_i$ (cf. Theorem \dref{T-properties}d).
  \item the normalized eigenvector corresponding to  $\lambda^U_{1}=0$ is ${V^U_{1}}\text{=}\frac{1}{\sqrt{n}}\mathbf{1}_n$.
\end{itemize}
it follows that $L^U$  can be diagonalized, with
\begin{equation*}
  L^U=V^U\Lambda^U {V^U}^T
\end{equation*}
 where $V^U$ is the  matrix of (right) orthonormal real eigenvectors  of $L^U$ and  $\Lambda^U$ is the diagonal matrix  of eigenvalues of $L^U$ (see Appendix \dref{App-decomp}).

Therefore we have
\begin{eqnarray*}
  e^{-L^U \cdot t}&=&e^{-\left( V^U \Lambda^U (V^U)^T \right) t} \\& =& V^U e^{-\Lambda^U t} (V^U)^T=e^{-\lambda^U_1 t}V^U_1 (V^U)_1^T+e^{-\lambda^U_2 t}V^U_2 (V^U_2)^T+.......+e^{-\lambda^U_n t} V^U_n (V^U_n)^T
\end{eqnarray*}

Since ${V^U_{1}}\text{=}\frac{1}{\sqrt{n}}\mathbf{1}_n$ we can write:

\begin{equation*}
  x^{(h)}(t)= e^{-L t}x(0) )=\frac{1}{n}{{\mathbf{1}_n}^{T}}x(0)\mathbf{1}_n + \sum_{i=2}^n e^{-\lambda^U_i t}((V^U_i)^T x(0)) V^U_i
\end{equation*}
or
\begin{equation}\dlabel{eq-xh}
   x^{(h)}(t)=\alpha \mathbf{1}_n + \sum_{i=2}^n e^{-\lambda^U_i t}((V^U_i)^T x(0)) V^U_i
\end{equation}
Since $\lambda^U_i>0$ forall $i>1$ we have
\begin{equation}\dlabel{eq-x1}
  x^{(h)}_\infty=\underset{t\to \infty }{\mathop{\lim }}\,x^{(h)}(t)=\frac{1}{n}{{\mathbf{1}_n}^{T}}x(0)\mathbf{1}_n=\alpha \mathbf{1}_n
\end{equation}
with
\
\begin{equation}\dlabel{eq-alphaG}
  \alpha =\frac{1}{n}\sum_{i=1}^n x_i(0)
\end{equation}
the \emph{average of the initial states}.

\PE

\subsubsection{Scaled influence}

Let $G^S$ be a strongly connected interactions (visibility) graph with scaled influences corresponding to $G^U$.
Then the Laplacian  $L^S$ has the following properties (cf. Appendix \dref{ScaledGraph}).
\begin{itemize}
  \item  The eigenvalues of $L^S$ are also the eigenvalues of $\Gamma$, the normalized Laplacian of $G^U$, a real symmetric matrix, as defined in section \dref{Def}, equation (\dref{Def-Gamma})
  \item All eigenvalues of $L^S$, denoted by $\lambda^S_i$, are real and non-negative
  \item There is a single zero eigenvalue, $\lambda^S_1=0$, and all remaining eigenvalues are strictly positive
  \item The eigenvectors of $L^S$ relate to the the eigenvectors of the real symmetric matrix $\Gamma$ by:
  \begin{equation*}
    V_i^S=\Delta^{-1/2} V_i^\Gamma
  \end{equation*}
  where $V_i^S$ and $V_i^\Gamma$ correspond to the eigenvalue $\lambda_i^S=\lambda_i^\Gamma$ and $\Delta$ is the degree matrix associated with the undirected graph $G^U$
  \item Since $\Gamma$ is real and symmetric, one can select $V_i^\Gamma$, for all $i$, s.t $V^\Gamma$ is real and orthonormal, where $V_i^\Gamma$ is the $i'th$ column of $V^\Gamma$  (cf. Theorem \dref{T-properties}d). Since $\Delta$ is real and invertible it follows that the corresponding $V^S$ is a matrix of normalized real right eigenvectors of $L^S$
  \item $V_1^S$ corresponding to $\lambda^S_1=0$ is  $\displaystyle \frac{1}{\sqrt{n}}\mathbf{1}_n$
  \item $L^S$ is diagonizable, thus it can be written as
  \begin{equation*}
    L^S=V^S \Lambda^S (V^S)^{-1}
  \end{equation*}
  where $\Lambda^S $ is a diagonal matrix, s.t. $\Lambda^S_{ii}=\lambda^S_i $ and $V^S$ is the matrix of normalized real right eigenvectors of $L^S$
  \item If we denote  $(V^S)^{-1}$ by $(W^S)^T$, i.e. $(W^S)^T=(V^S)^{-1}$, then

\begin{itemize}
  \item Each row of $(W^S)^T$ is a left eigenvector of $L^S$
  \item The first row of $(W^S)^T$, denoted by $(W^S_1)^T$, is a left eigenvector of $L^S$ corresponding to $\lambda_1^S=0$ satisfying  $(W^S_1)^T V^S_1=1$, where $V^S_1$ is the normalized right eigenvector corresponding to $\lambda_1^S=0$
\end{itemize}
\item According to Theorem \dref{T-WsT1} in  Appendix \dref{ScaledGraph}
\begin{equation}\dlabel{eq-Ws1T}
  (W^S_1)^T= \frac{\sqrt{n} \cdot \mathbf{d}^T}{\sum_{i=1}^n d_i}
\end{equation}
where $\mathbf{d}$ is a vector of degrees in the graph $G^U$, $\mathbf{d}_i=d_i$, and $d_i$ is the degree of vertex $i$ in $G^U$.
\end{itemize}

\LB{L-scaledConverge}
The value of the \emph{asymptotic consensus state} $\alpha$ for a strongly connected digraph $G^S$ representing scaled influences,  is in the convex hull of the initial states $x(0)$ and is given by
\begin{equation*}
  \alpha=  \frac{ \mathbf{d}^T x(0)}{\sum_{i=1}^n d_i}
\end{equation*}
where $\mathbf{d}$ is the vector of degrees in the undirected graph $G^U$ corresponding to $G^S$.
\LE

\PB
We have
\begin{equation*}
    L^S=V^S \Lambda^S (W^S)^T
  \end{equation*}
  where $(W^S)^T=(V^S)^{-1}$
  Thus
  \begin{eqnarray*}
x^{(h)}(t)&=&e^{-L^S t} x(0)\\ & = & e^{-\lambda^S_1 t}(W^S_1)^{T}x(0)V^S_1+e^{-\lambda^S _2t}(W^S_2)^T x(0) V^S_2+.......+e^{-\lambda^S_n t}(W^S_n)^{T} x(0)V^S_n\\
& = & (W^S_1)^{T}x(0)V^S_1 + \sum_{i=2}^N e^{-\lambda^S_i t}(W^S_i)^{T}x(0)V^S_i\\
& = & \frac{ \mathbf{d}^T x(0)}{\sum_{i=1}^n d_i} \mathbf{1}_n+ \sum_{i=2}^N e^{-\lambda^S_i t}(W^S_i)^{T}x(0)V^S_i
\end{eqnarray*}

where we used
\begin{itemize}
  \item $\lambda^S_1=0$
  \item  $\displaystyle V^S_1 = \frac{1}{\sqrt{n}}\mathbf{1}_n$
  \item $(W^S_1)^T$ from eq. (\dref{eq-Ws1T})
\end{itemize}
Since $ \lambda^S_i>0 \quad \forall i\geq 2$ we have for $t \rightarrow \infty$
\begin{equation}\dlabel{eq-x1s}
  x^{(h)}_\infty=\underset{t\to \infty }{\mathop{\lim }}\,x^{(h)}(t)=\frac{ \mathbf{d}^T x(0)}{\sum_{i=1}^n d_i} \mathbf{1}_n =\alpha \mathbf{1}_n
\end{equation}
Thus, $\displaystyle \alpha=  \frac{ \mathbf{d}^T x(0)}{\sum_{i=1}^n d_i}$ is the asymptotic consensus value for dynamics with scaled influences and no external input.
\PE

Lemma \dref{L-scaledConverge} holds for any visibility graph, $G^U$. If we let the graph be complete, then we have $d_i=n-1; \quad i=1,..,n$ and thus $\displaystyle \alpha=  \frac{ 1}{n} {\sum_{i=1}^n x(0)}$, i.e. the asymptotic consensus value, for complete graphs with scaled influence, is the average of the initial states. The above results can be summarized by the following theorem:

\TB{T-ConsensusValue}
The value of the \emph{asymptotic consensus state}, $\alpha$, of $n$ agents with a connected visibility (interactions) graph is
\hB
\titem{a} the average of the initial states if the influence is uniform or  if the influence is scaled and the visibility graph is complete.
\titem{b} the weighted average of the initial states, $\displaystyle \alpha=  \frac{ \mathbf{d}^T x(0)}{\sum_{i=1}^n d_i}$, if the influence is scaled and the visibility graph is incomplete, where
\begin{itemize}
  \item $d_i$ is the degree of vertex $i$ in $G^U$
  \item $\mathbf{d}^T$ is the vector of degrees in $G^U$, i.e. $\mathbf{d}^T=(d_1 d_2 ....d_n)$
\end{itemize}
\hE

\TE

\msubsection{Input induced group dynamics}{MoveAgreement}

Next, consider the general form of the input-related part of the group dynamics,  $x^{(u)}(t)$, given by eq.  (\dref{gen-xu}), where $L, B$ and $u$ are constant in the time interval $[0,t]$.

 \begin{equation}\dlabel{gen-xu}
   x^{(u)}(t)= \int_{0}^t e^{-L(t-\tau)}B u d\tau = \int_0^{t} e^{-L\nu}B u d\nu
\end{equation}
Eq. (\dref{gen-xu}) holds for both the uniform and the scaled influence, i.e $L=L^U$ or $L=L^S$.
\begin{itemize}
  \item For the \textbf{uniform influence case}, since $L^U$ is symmetric we can use again the Spectral theorem and decompose (\dref{gen-xu}) into
\begin{equation}\dlabel{Sym-xu}
  x^{(u)}(t)=
 \sum_{i=1}^n\left [ \int_0^{t} e^{-\lambda^U_i \nu}V^U_i (V^U_i)^T d\nu \right ] B u
\end{equation}
  \item For \textbf{any} visibility graph with \textbf{scaled influence }, using the properties of $L^S$, we can  write
\begin{equation}\dlabel{Scaled-xu}
  x^{(u)}(t)=
 \sum_{i=1}^n \left [ \int_0^t e^{-\lambda^S_i \nu}V^S_i (W^S_i)^T d\nu \right ] B u
\end{equation}
\end{itemize}
Since for both $L^U$ and $L^S$, representing connected graphs,  there is a single zero eigenvalue and the remaining eigenvalues are positive,  we can decompose  $x^{(u)}(t)$  in two parts:
\begin{equation}
  x^{(u)}(t)= x^{(a)}(t)+ x^{(b)}(t)
\end{equation}\dlabel{eq-xu}
where
\begin{itemize}
  \item $x^{(a)}(t)$ is the zero eigenvalue dependent term, representing the movement in the agreement space
  \item $x^{(b)}(t)$ is the remainder, representing the deviation from the agreement space
\end{itemize}

\subsubsection{Movement along the agreement subspace }

\paragraph{The uniform case}\mbox{}

We have

  \begin{equation}\dlabel{Uni-xua}
    x^{(a)}(t) = \int_0^t e^{-\lambda^U _1  \nu} \text{} V^U_1 (V^U_1)^T B u d\nu = V^U_1 (V^U_1)^T B u  t = \frac{n_l}{n} \text{} u \text{} t\mathbf{1}_n
  \end{equation}
where $n_l$ is the number of leaders and we have used $V^U_1=\frac{1}{\sqrt{n}} \mathbf{1}$ and $\mathbf{1}^T B=n_l$.  Therefore:

 \LB{L-firstSym}
    Consider a group of $n$ agents, forming a connected interactions graph, and moving according to (\dref{x-1D})  with uniform influences.  If there are $n_l$ agents that receive an exogenous velocity control $u$,  the entire group will move collectively with a velocity $\displaystyle \frac{n_l}{n} u $.
\LE

\paragraph{Scaled case}\mbox{}

For the scaled case we have

  \begin{equation}\dlabel{Scaled-xua}
    x^{(a)}(t) = \int_0^t e^{-\lambda^S _1  \nu} \text{} V^S_1 (W^S_1)^T B u d\nu = V^S_1 (W^S_1)^T B u t = \frac{\sum_{i \in N^l} d_i}{\sum_{i=1}^n d_i}\text{} u \text{} t \text{} \mathbf{1}_n
  \end{equation}

  Substituting in (\dref{Scaled-xua}) $V^S_1= \displaystyle \frac{1}{\sqrt{n}} \mathbf{1}_n$ and $(W^S_1)^T$ from (\dref{eq-Ws1T}) we obtain

  \begin{equation}\dlabel{Scaled-beta}
    x^{(a)}(t) =  \frac{\sum_{i \in N^l} d_i}{\sum_{i=1}^n d_i} \text{} u \text{}  t \text{} \mathbf{1}_n
  \end{equation}

 \LB{L-firstScaled}
 Consider a group of $n$ agents, forming a strongly connected interactions graph with scaled influences, with some of the agents being leaders, i.e. detecting the exogenous velocity control $u$.    If each  agent  moves according to (\dref{x-1D}) then  the entire group will move collectively with a velocity $\displaystyle \frac{\sum_{i \in N^l} d_i}{\sum_{i=1}^n d_i} u$, where $N^l$ is the set of leaders and $d_i$ is the number of edges entering $i$.
\LE
We see from (\dref{Scaled-beta}) that if the the influences are scaled and
\begin{enumerate}
  \item the visibility graph is complete then the collective velocity of the group reduces to  $\displaystyle \frac{n_l}{n} u $, same as for the uniform case.
  \item the visibility graph is incomplete then the collective velocity of the group is a function not only of the number of leaders but also of the number of links connecting the leaders to followers
\end{enumerate}

\textbf{\textsf{Example}}:

 We illustrate the impact of leader selection on the collective velocity, when the visibility graph is incomplete and scaled influence is used, by considering  the two configurations shown in Fig. \dref{ChangeLeader}, with identical $G^U$, but different leader.
 \begin{figure}
\begin{center}
\includegraphics[scale=0.5]{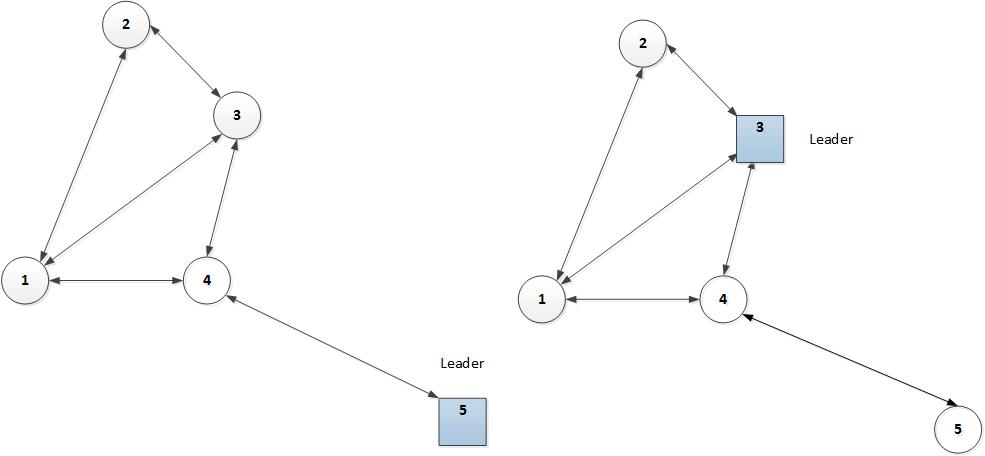}
\caption{Same $G^u$ with different leader }\dlabel{ChangeLeader}
\end{center}
\end{figure}

Based on Lemma \dref{L-firstScaled}, when agent 5 is the leader the group will move with velocity $\displaystyle \frac{1}{12} u$, while when the leader is agent 3 the collective velocity increases to $\displaystyle \frac{1}{4} u$. Note that when uniform influence is employed, the collective velocity depends only on the \emph{number of leaders}. Thus, in both above configurations, the collective velocity is $\displaystyle \frac{1}{5} u$

\subsubsection{Deviations from the agreement subspace }\label{DevProp}

Consider now the remainder $x^{(b)}(t)$ of the input-related part, i.e. the part of $x^{(u)}(t)$ containing all eigenvalues of $L$ other than the zero eigenvalue and representing the agents' state deviation from the agreement subspace. The geometric meaning of deviations is elaborated in section \dref{Dev-mean}.

In the sequel we will need the following definitions:
\begin{definition}Two agents $i,j$ in a network $G$ are said to be \emph{equivalent} if there exists a  Leaders-Followers Preserving Permutation $\Pi$ such that $\Pi(i) = j$,$\Pi(j) = i$ and $\Pi(G) = G$   \end{definition}
\begin{definition} A Leaders-Followers Preserving Permutation $\Pi$ is a permutation of  agents labeling such that
$\Pi(leader)$ is a leader and $\Pi(follower)$ is a follower for all leaders and followers.   \end{definition}

\paragraph{Uniform case}\mbox{}

We have

\begin{equation*}
  x^{(b)}(t)=
 \left [ \sum_{i=2}^n \int_0^t \left (e^{-\lambda^U _i \cdot \nu} \right ) V^U_i (V^U_i)^T d\nu \right ] B u
\end{equation*}
Thus
\begin{equation}\dlabel{Uni-xub}
  x^{(b)}(t)= \left[ \sum_{i=2}^n \frac{1}{\lambda^U_i}(1-e^{-\lambda^U_i t})V^U_i (V^U_i)^T \right ] B u
\end{equation}
Since all eigenvalues $\lambda^U_i \text{ for } i \geq 2$ are strictly positive, $x^{(b)}(t)$ converges  asymptotically to a time independent vector, denoted by $\varrho$, given by:
\begin{equation}\dlabel{Uni-Delta}
  \varrho = \left[ \sum_{i=2}^n \frac{1}{\lambda^U_i}V^U_i (V^U_i)^T \right ] B u
\end{equation}

The quantity $\varrho$ represents the vector of asymptotic deviations of the agents from the agreement subspace.

\TB{T-DevSum}
The asymptotic deviations of all agents, with  uniform interactions,  sum to zero

\begin{equation}\dlabel{eq-sumb}
\sum_{i=1}^n \varrho_i = 0
\end{equation}
where $\varrho_i$ is the deviation of agent $i$.
\TE

\PB
Consider eq. (\dref{1D-vec}) with $L=L^U$ and multiply it from the left by $\mathbb{1}^T$.
Recalling that  $L^U$ has  a left eigenvector $\mathbb{1}^T$ corresponding to $\lambda^U_1=0$, we obtain
\begin{equation*}
\sum_{i=1}^n \dot{x}_i(t) = n_l  u
\end{equation*}
and thus, for all $t$
\begin{equation}\dlabel{eq-sumt}
\sum_{i=1}^n x_i(t) = n_l  u t + \sum_{i=1}^n x_i(0)
\end{equation}
On the other hand, recalling that
\begin{equation}\dlabel{eq-xt}
  x(t)= x^{(h)}(t)+ x^{(a)}(t) + x^{(b)}(t)
\end{equation}
 multiplying  (\dref{eq-xt}) from the left by $\mathbf{1}_n^T$ and letting  $t\rightarrow \infty$, we have:
\begin{equation}\dlabel{eq-sumtt}
\sum_{i=1}^n x_i(t \rightarrow \infty) = n \alpha + n_l u t + \sum_{i=1}^n \varrho_i
\end{equation}
Substituting for $\alpha$ its value from eq. (\dref{eq-alphaG}) and comparing equations (\dref{eq-sumt}) and (\dref{eq-sumtt}) we obtain the required result (\dref{eq-sumb}).
\PE

In general, agents have non-equal deviations, but there are some special cases, detailed in Theorem \dref{T-SpecialDevUni}.

\TB{T-SpecialDevUni}
\hB
\titem{a} Equivalent agents have the same deviation
\titem{b} In a fully connected network, all followers  have the same asymptotic deviation and all leaders have the same asymptotic deviation, with opposite sign to followers' deviation.
\titem{c} If all agents are leaders, i.e. $n_l=n$,  then all asymptotic deviations are zero, i.e. $\varrho_i=0, \text{  } \forall i$
\hE
\TE
\PB\\
\tref{a} Equivalent agents follow the same equation, therefore have the same deviation.\\
\tref{b} In a fully connected network, all followers are equivalent to each other and all leaders are equivalent to each other. Thus all followers have the same asymptotic deviation and all leaders have the same asymptotic deviation (different from the followers).  Since the sum of all asymptotic deviations is zero, eq. (\dref{eq-sumb}), the deviations of the followers and of the leaders have opposite signs.\\
\tref{c} If $n_l=n$, then $B=\mathbf{1}_n$. Since $V^U_k$ is an eigenvector of the Laplacian $L^U$ with eigenvalue $\lambda^U_k$, we have $L^U V^U_k = \lambda^U_k V^U_k$ or
\begin{equation}\label{eq-Vuk}
  V^U_k = \frac {1}{\lambda^U_k}L^U V^U_k
\end{equation}
Substituting (\dref{eq-Vuk}) in equation (\dref{Uni-Delta}) we obtain:
\begin{equation*}
  \varrho = \left( \sum_{k=2}^n \frac{1}{(\lambda^U_k)^3} (L^U V^U_k) ((V^U_k)^T (L^U)^T) \right) \mathbf{1}_n u = \mathbf {0}_n
\end{equation*}
since $(L^U)^T \mathbf{1}_n  = \mathbf {0}_n$.
\PE

\paragraph{Scaled case}\mbox{}\\
Following the same procedure as above, with the corresponding decomposition of $L=L^S$, we obtain
the following expression for $x^{(b)}(t)$, in the scaled case:
\begin{equation}\dlabel{eq-xbtScaled}
  x^{(b)}(t)= \left[ \sum_{i=2}^n \frac{1}{\lambda^S_i}(1-e^{-\lambda^S_i t} ) V^S_i (W^S_i)^T \right ] B u
\end{equation}
and since $\lambda^S_i; \quad i \geq 2$ are positive
\begin{equation}\label{eq-devScaled}
  \varrho=\underset{t\to \infty }{\mathop{\lim }}\,x^{(b)}(t)=\left [ \sum_{i=2}^n \frac{1}{\lambda^S_i} V^S_i (W^S_i)^T \right ] B u
\end{equation}
Thus, here again  $x^{(b)}(t)$,  converges asymptotically to a time-independent vector,   $\varrho$, given by (\dref{eq-devScaled}) and representing asymptotic deviations from the agreement subspace.
\TB{T-DevSumScaled}
The \textbf{weighted sum} of the asymptotic deviations of all agents, with scaled pair-wise interactions,  is zero

\begin{equation}\dlabel{eq-sumbs}
\sum_{i=1}^n d_i \varrho_i = 0
\end{equation}
where $\varrho_i$ is the deviation of agent $i$ and $d_i$ is the number of edges entering $i$.
\TE

\PB
Multiplying eq. (\dref{1D-vec}), where $L=L^S$, from the left by $(W^S_1)^T$ and integrating, we obtain \emph{for any $t$}
\begin{equation}\dlabel{eq-xtScaled}
          \mathbf{d}^T \cdot x(t) = \mathbf{d}^T \cdot B u t +\mathbf{d}^T \cdot x(0)
 \end{equation}
        where $\mathbf{d}$ is the vector of degrees in the corresponding $G^U$ and we used
        \begin{itemize}
          \item $(W^S_1)^T L^S = \mathbf{0}_n^T$
          \item $\displaystyle (W^S_1)^T=\frac{\sqrt{n} \mathbf{d}^T}{\sum_{i=1}^{n} d_i}$
        \end{itemize}
     Considering now  $t \rightarrow \infty$,  we can write $x(t \to \infty)$ from eq. (\dref{eq-xt}) as
\begin{equation}\label{xt-inf}
  x(t \to \infty) = \frac{\mathbf{d}^T x(0)}{\sum_{i=1}^{n} d_i} \mathbf{1}_n + \frac{\sum_{i \in N^l} d_i}{\sum_{i=1}^n d_i} u t \mathbf{1}_n + \varrho
\end{equation}
     where we used Lemma \dref{L-scaledConverge} and Lemma \dref{L-firstScaled}.\\
 Multiplying (\dref{xt-inf}) from the left  by $\mathbf{d}^T$  we obtain

\begin{equation}\label{xt-inf2}
\begin{split}
  \mathbf{d}^T \cdot x(t) & =  \frac{\mathbf{d}^T x(0)}{\sum_{i=1}^{n} d_i} \sum_{i=1}^{n} d_i + \frac{\sum_{i \in N^l} d_i}{\sum_{i=1}^n d_i} u t \sum_{i=1}^{n} d_i + \mathbf{d}^T \cdot \varrho \\
  & = \mathbf{d}^T x(0) + \sum_{i \in N^l} d_i  u t + \mathbf{d}^T \cdot \varrho\\
  & = \mathbf{d}^T \cdot x(0) + \mathbf{d}^T \cdot B u t + \mathbf{d}^T \cdot \varrho
\end{split}
\end{equation}

Comparing now (\dref{xt-inf2}) with (\dref{eq-xtScaled}) for $t\to\infty$ we immediately obtain the required result (\dref{eq-sumbs}).

\PE

Theorem \dref{T-DevPropertiesScaled} shows properties of the asymptotic deviations of agents with scaled influences in some special cases. These properties for scaled influences are identical to the corresponding ones for uniform influence.

\TB{T-DevPropertiesScaled}
$n$ agents with scaled interaction, out of which $n_l$ agents are leaders, satisfy the following:
\hB
\titem{a} All equivalent agents have the same asymptotic deviation
\titem{b} In a fully connected network  all followers have the same asymptotic deviation and all leaders have the same asymptotic deviation, with opposite sign to followers' deviation.
\titem{c} If all agents are leaders, i.e. $n_l=n$, then $\varrho_i=0, \text{  } \forall i$

\hE
\TE
\PB\\
\tref{a} Equivalent agents follow the same equation, therefore have the same deviation.\\
\tref{b} In a fully connected network, with scaled influences,
\begin{itemize}
  \item $d_i=n-1; \quad \forall i$, thus substituting in  (\dref{eq-sumbs}) we obtain $\displaystyle \sum_{i=1}^n \varrho_i=0$
  \item all leaders are equivalent and all followers are equivalent,  thus all leaders have the same asymptotic deviation, $\varrho_l$, and all followers have the same asymptotic deviation, $\varrho_f$
 \end{itemize}
 Thus
  \begin{equation*}
      n_l \varrho_l + n_f \varrho_f  =  0
  \end{equation*}
Thus,    $sign(\varrho_l) = - sign(\varrho_f)$ \\
\tref{c} If $n_l=n$, then $B=\mathbf{1}_n$. Since $V^S_k$ is a right eigenvector of the Laplacian $L^S$ with eigenvalue $\lambda^S_k$ and $(W^S_k)^T$ is a left eigenvector with the same eigenvalue, we have
\begin{itemize}
  \item $\displaystyle V^S_k = \frac {1}{\lambda^S_k}L V^S_k$
  \item $\displaystyle (W^S_k)^T = \frac {1}{\lambda^S_k} (W^S_k)^T L^S$
\end{itemize}
Substituting in equation (\dref{eq-devScaled}) we obtain:
\begin{equation*}
  \varrho = \left( \sum_{k=2}^n \frac{1}{(\lambda^S_k)^3} (L^S V^S_k) ((W^S_k)^T L^S) \right) \mathbf{1}_n u = \mathbf {0}_n
\end{equation*}
since $L^S\mathbf{1}_n  = \mathbf {0}_n$.\\
\PE

\paragraph{Illustration of asymptotic deviations for various cases}\mbox{}

\begin{figure}
\begin{center}
\includegraphics[scale=0.6]{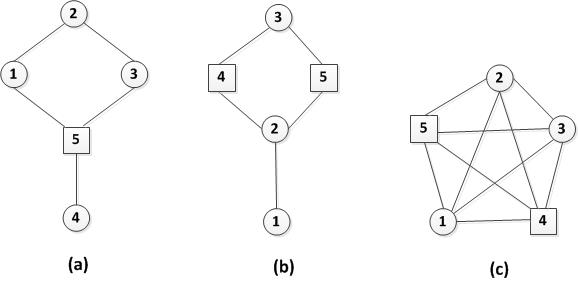}
\caption{Several Networks (Leaders are squares)}\label{fig-NewExamples}
\end{center}
\end{figure}

In this section we illustrate by a few examples the impact of the influence model as well as of the equivalence on the obtained deviations. Due to the construction of the interaction graph with scaled influence, $G^S$, out of the the interaction graph with uniform influence, $G^U$, equivalent nodes in $G^U$ are also equivalent in $G^S$.
Consider the graphs in Fig. \dref{fig-NewExamples}.

Denote by $A^U$ the adjacency matrix corresponding to $G^U$ and by $A^S$ the adjacency matrix corresponding to $G^S$. Then the adjacency matrices  for each interactions graph depicted  in  Fig. \dref{fig-NewExamples}, uniform or scaled, are:
\newpage
\begin{description}

  \item[(a)]
 \begin{multicols}{2}
  \begin{equation*}
    A^U=\left [ \begin{matrix}
    0 & 1 & 0 & 0 & 1\\
    1 & 0 & 1 & 0 & 0\\
    0 & 1 & 0 & 0 & 1\\
    0 & 0 & 0 & 0 & 1\\
    1 & 0 & 1 & 1 & 0 \end{matrix} \right ]
  \end{equation*}

  \begin{equation*}
    A^S=\left [ \begin{matrix}
    0 & \frac{1}{2} & 0 & 0 & \frac{1}{2}\\
    \frac{1}{2} & 0 & \frac{1}{2} & 0 & 0\\
    0 & \frac{1}{2} & 0 & 0 & \frac{1}{2}\\
    0 & 0 & 0 & 0 & 1\\
    \frac{1}{3} & 0 & \frac{1}{3} & \frac{1}{3} & 0 \end{matrix} \right ]
  \end{equation*}
\end{multicols}

  \item[(b)]
  \begin{multicols}{2}
  \begin{equation*}
    A^U=\left [ \begin{matrix}
    0 & 1 & 0 & 0 & 0\\
    1 & 0 & 0 & 1 & 1\\
    0 & 0 & 0 & 1 & 1\\
    0 & 1 & 1 & 0 & 0\\
    0 & 1 & 1 & 0 & 0 \end{matrix} \right ]
  \end{equation*}

  \begin{equation*}
    A^S=\left [ \begin{matrix}
    0 & 1 & 0 & 0 & 0\\
    \frac{1}{3} & 0 & 0 & \frac{1}{3} & \frac{1}{3}\\
    0 & 0 & 0 & \frac{1}{2} & \frac{1}{2}\\
    0 & \frac{1}{2} & \frac{1}{2} & 0 & 0\\
    0 & \frac{1}{2} & \frac{1}{2} & 0 & 0 \end{matrix} \right ]
  \end{equation*}
\end{multicols}

  \item[(c)]
  \begin{multicols}{2}
  \begin{equation*}
    A^U=\left [ \begin{matrix}
    0 & 1 & 1 & 1 & 1\\
    1 & 0 & 1 & 1 & 1\\
    1 & 1 & 0 & 1 & 1\\
    1 & 1 & 1 & 0 & 1\\
    1 & 1 & 1 & 1 & 0 \end{matrix} \right ]
  \end{equation*}

  \begin{equation*}
    A^S=\left [ \begin{matrix}
    0 & 0.25 & 0.25 & 0.25 & 0.25\\
    0.25 & 0 & 0.25 & 0.25 & 0.25\\
    0.25 & 0.25 & 0 & 0.25 & 0.25\\
    0.25 & 0.25 & 0.25 & 0 & 0.25\\
   0.25 & 0.25 & 0.25 & 0.25 & 0 \end{matrix} \right ]
  \end{equation*}
\end{multicols}
\end{description}

Denoting now by $\varrho^U$ the deviations vector for the uniform case, by $\varrho^S$ the deviations vector for the scaled case and using the input $u =1$ in all examples we obtain:
\begin{description}

  \item[(a)] Node 5 is the leader, nodes 1 and 3 are equivalent, the others have no equivalents.
 \begin{multicols}{2}
  \begin{equation*}
    \varrho^U=\left [ \begin{matrix}
    -0.06\\
    -0.16\\
    -0.06\\
    0.04\\
    0.24 \end{matrix} \right ]
  \end{equation*}

  \begin{equation*}
   \varrho^S=\left [ \begin{matrix}
   -0.2526\\
   -0.5142\\
   -0.2526\\
    0.2952\\
    0.5812 \end{matrix} \right ]
  \end{equation*}
\end{multicols}
We see in this example that
\begin{itemize}
  \item In both cases, uniform and scaled influence, the asymptotic deviations of the equivalent agents' 1 and 3, are identical
  \item $\displaystyle \sum_{i=1}^n \varrho^U_i = 0$
  \item $\displaystyle \sum_{i=1}^n \varrho^S_i \neq 0$
  \item $\displaystyle \sum_{i=1}^n d_i \varrho^S_i = 0$ where $d_i$ is the $i'th$ element of $\mathbf{d}^T=[2 \quad 2\quad 2 \quad 1 \quad 3]$
\end{itemize}

  \item[(b)] The leaders, nodes 4 and 5 are equivalent, the others have no equivalent
 \begin{multicols}{2}
  \begin{equation*}
    \varrho^U=\left [ \begin{matrix}
   -0.5200\\
   -0.1200\\
    0.0800\\
    0.2800\\
    0.2800 \end{matrix} \right ]
  \end{equation*}

  \begin{equation*}
   \varrho^S=\left [ \begin{matrix}
   -0.6857\\
   -0.3368\\
    0.0284\\
    0.4098\\
    0.4098
 \end{matrix} \right ]
  \end{equation*}
\end{multicols}

In this example again
\begin{itemize}
  \item In both cases, uniform and scaled influence, the asymptotic deviations of the equivalent agents' 4 and 5, are identical
  \item $\displaystyle \sum_{i=1}^n \varrho^U_i = 0$
  \item $\displaystyle \sum_{i=1}^n \varrho^S_i \neq 0$
  \item $\displaystyle \sum_{i=1}^n d_i \varrho^S_i = 0$ where $d_i$ is the $i'th$ element of $\mathbf{d}^T=[1 \quad 3 \quad 2 \quad 2 \quad 2]$
\end{itemize}

\item[(c)] Nodes 4 and 5 are leaders.  Clearly they are equivalent, and so are nodes 1, 2, 3.
 \begin{multicols}{2}
  \begin{equation*}
    \varrho^U=\left [ \begin{matrix}
   -0.0800\\
   -0.0800\\
   -0.0800\\
    0.1200\\
    0.1200 \end{matrix} \right ]
  \end{equation*}

  \begin{equation*}
   \varrho^S=\left [ \begin{matrix}
   -0.3200\\
   -0.3200\\
   -0.3200\\
    0.4800\\
    0.4800 \end{matrix} \right ]
  \end{equation*}
\end{multicols}
In this example, as before:
 \begin{itemize}
   \item equivalent nodes have identical deviations, for both uniform and scaled influences
   \item $\displaystyle \sum_{i=1}^n \varrho^U_i = 0$,
 \end{itemize}
  but also $\displaystyle \sum_{i=1}^n \varrho^S_i = 0$. This is due to the completeness of the graph, as stated in Theorem \dref{T-DevPropertiesScaled}. Moreover, we note that in this example, where the visibility graph is complete, the dispersion of agents along the alignment line, i.e. the distance between the position of leaders to the position of followers, is 4 times larger in the scaled case than in the uniform case. This is due to the following holding \textbf{for complete graphs};
  \begin{itemize}
    \item     $\lambda^U_i=n; \quad i=2,....,n$ and thus $\displaystyle \varrho^U = \frac{1}{n}\left [ \sum_{i=2}^n V^U_i (V^U_i)^T \right ] B u$
    \item     $\displaystyle \lambda^S_i=\frac{n}{n-1}; \quad i=2,....,n$; and thus $\displaystyle \varrho^S=\frac{n-1}{n}\left [ \sum_{i=2}^n V^S_i (W^S_i)^T \right ] B u$.\\ Since $V^S=V^U$ and $(V^S_i)^T=(W^S_i)^T$ we obtain, when the same $u$ is used in both cases, \textbf{$ \varrho^S =(n-1) \varrho^U $}

  \end{itemize}

\end{description}

\msec{Two dimensional group dynamics }{LTI-dyn}

In this section we derive the asymptotic dynamics of a two-dimensional group of agents, in a time interval $[0,t)$, where the system is time independent and the visibility graph is connected.
Denote by $p_i = (x_i,y_i)^T $ the position of agent $i$ at time $t$.  Let $x(t)$ denote the $n$-dimensional vector $x(t) = (x_1(t) ... x_n(t))^T$ and similarly for $y(t)$.  Let $p(t)$ be the $2n$-dimensional vector $(x^T(t)\quad y^T(t))^T$.

Assuming that the two dimensions are decoupled,    Eq. (\dref{1D-vec}) holds for each component and thus:

\begin{eqnarray*}
   \dot{x} ( t ) &=& -L\cdot x(t)+ Bu_x  \\
   \dot{y}( t ) &=& -L\cdot y(t)+ Bu_y
  \end{eqnarray*}

Applying the results derived in section  \dref{Gen-1D}, for one time interval, we can write:
\begin{equation}\label{p-decoupled}
  p(t)= p^{(h)}(t)+p^{(a)}(t)+p^{(b)}(t)= \left [\begin{matrix}
 x^{(h)}(t)+x^{(a)}(t)+x^{(b)}(t)\\
  y^{(h)}(t)+y^{(a)}(t)+y^{(b)}(t)
  \end{matrix} \right ] \\
  \end{equation}
where $t$ is the time from the beginning of the interval.
Thus,  for the $x$ axis, we have the following expressions and for the $y$ axis we have the same with $y$ replacing $x$.
\begin{itemize}
  \item for the \textbf{uniform case}

\begin{eqnarray*}
  x^{(h)}(t) & = &\frac{1}{n}{\mathbf{1}_n^{T}}x(0)\mathbf{1}_n + \sum_{k=2}^n e^{-\lambda^U_k t} ((V^U_k)^T x(0)) V^U_k\\
  x^{(a)}(t) & = &\frac{n_l}{n} u_x t  \mathbf{1}_n\\
  x^{(b)}(t) &= &\left[ \sum_{k=2}^n \frac{1}{\lambda^U_k}(1-e^{-\lambda^U_k t})V^U_k (V^U_k)^T \right ] B u_x
\end{eqnarray*}
where
\begin{itemize}
  \item  $\lambda^U_i; \quad i=1, ...,n$ are the eigenvalues of the Laplacian $L^U$ and  $V^U_i; \quad i=1, ...,n $ are the corresponding right eigenvectors, selected such that the eigenvector corresponding to $\lambda^U_1=0$ is $V^U_1=\displaystyle \frac{1}{\sqrt{n}}\mathbf{1}_n$ and $V^U$, the matrix with columns  $V^U_i$, is orthonormal.
  \item we assumed in the expression for $ x^{(a)}(t)$ that there are $n_l$ agents receiving the exogenous input
\end{itemize}

  \item for the \textbf{scaled case}
  \begin{eqnarray*}
  x^{(h)}(t) & =  & \frac{ \mathbf{d}^T x(0)}{\sum_{i=1}^n d_i} \mathbf{1}_n+ \sum_{i=2}^n e^{-\lambda^S_i t}(W^S_i)^{T} x(0) V^S_i\\
  x^{(a)}(t) & = & \frac{\sum_{i \in N^l} d_i}{\sum_{i=1}^n d_i} u_x  t \mathbf{1}_n\\
  x^{(b)}(t) &= &\left[ \sum_{k=2}^n \frac{1}{\lambda^S_k}(1-e^{-\lambda^S_k t})V^S_k  (W^S_k)^T \right ] B u_x
 \end{eqnarray*}
\end{itemize}
 where
 \begin{itemize}
   \item  $\lambda^S_i; \quad i=1... n$ are the eigenvalues of the Laplacian $L^S$, s.t. $\lambda^S_1=0$
   \item   $V^S$ is a matrix whose columns, $V^S_i$, are the \emph{normalized} right eigenvectors of $L^S$.  In particular,  the normalized right eigenvector corresponding to $\lambda^S_1=0$, is $\displaystyle V^S_1=\frac{1}{\sqrt{n}} \mathbf{1}_n$.
   \item $(W^S)^T$  is the matrix of left eigenvectors, selected s.t.  $(W^S)^T = (V^S)^{-1}$ .  The first row of $(W^S)^T$, denoted by $(W^S_1)^T$, is a left eigenvector of $L^S$ corresponding to $\lambda^S_1=0$ and satisfies Theorem \dref{T-WsT1}:
  \begin{equation*}
  (W^S_1)^T = \frac{\sqrt{n}  \mathbf{d}^T}{\sum_{i=1}^n d_i}
\end{equation*}

\item $d_i$ are the number of neighbors of node $i$ and $ \mathbf{d}$ is a vector with $d_i$ as its $i'th$ element
 \end{itemize}

\msubsection{Interpretation of the asymptotic deviations in the Euclidean space}{Dev-mean}

\subsubsection{Asymptotic position of agent $i$}\label{AsympPos2D}

The  asymptotic positions of agent $i$, in the two-dimensional space, when an external control  $\mathbf{u}=(u_x \quad u_y)^T$  is detected by $n_l$ agents,  will be

\begin{equation}\label{pAsymp}
  p_i(t \rightarrow \infty)=  \left [\begin{matrix}
 \alpha_x +\beta  u_x  t + \gamma_i  u_x\\
 \alpha_y +\beta  u_y  t  + \gamma_i  u_y
  \end{matrix} \right ] \\
  \end{equation}

 where
 \begin{itemize}
   \item $\alpha=(\alpha_x\text{   }\alpha_y)^T$ is the agreement, or gathering, point when there is no external input
   \item $\beta  (u_x\text{   } u_y)^T $ is the collective velocity.
   \item  $\gamma_i  (u_x \text{   } u_y)^T $ are the $x$ and $y$ components of the asymptotic deviation of agent $i$
 \end{itemize}

The values of $\alpha_x, \text{   }\alpha_y, \text{   }\beta$ and $\gamma$, the coefficients of the asymptotic position, are a function  of the assumed influence model, as shown in Table \dref{AsymptoticPosCoef} for a general visibility graph.  Note that $\gamma_i$ is the deviation factor, i.e. $\gamma_i u_x$ is the deviation of agent $i$ in the $x$ direction and similarly for $y$.

 \begin{table}
\centering
\caption{Coefficients of asymptotic position}
\label{AsymptoticPosCoef}
 \begin{tabular}{||c|| c| c||}
 \hline
 & \textbf{Uniform influence} & \textbf{Scaled influence}  \\ [0.5ex]
 \hline\hline
$\alpha_x$ & $\displaystyle \frac{1}{n} \mathbf{1}_n^T x(0)$ &  $\displaystyle \frac{ \mathbf{d}^T x(0)}{\sum_{i=1}^n d_i}$ \\[2ex]
  \hline
$\alpha_y$  &  $\displaystyle \frac{1}{n} \mathbf{1}_n^T  y(0)$ &$ \displaystyle \frac{ \mathbf{d}^T y(0)}{\sum_{i=1}^n d_i}$  \\[2ex]
 \hline
 $\beta$ & $\displaystyle \frac{n_l}{n}$  & $ \displaystyle \frac{\sum_{i \in N^l} d_i}{\sum_{i=1}^n d_i}$ \\[2ex]
  \hline
 $\gamma$ & $\displaystyle \sum_{k=2}^n \left[ \frac{1}{\lambda^U_k}V^U_k (V^U_k)^T \right ] B$ & $\displaystyle \sum_{k=2}^n \left[ \frac{1}{\lambda^S_k}V^S_k (W^S_k)^T \right ] B$ \\[2ex]
  \hline
 \end{tabular}
\end{table}

\subsubsection{Asymptotic deviations}
\begin{figure}
\begin{center}
\includegraphics[scale=0.7]{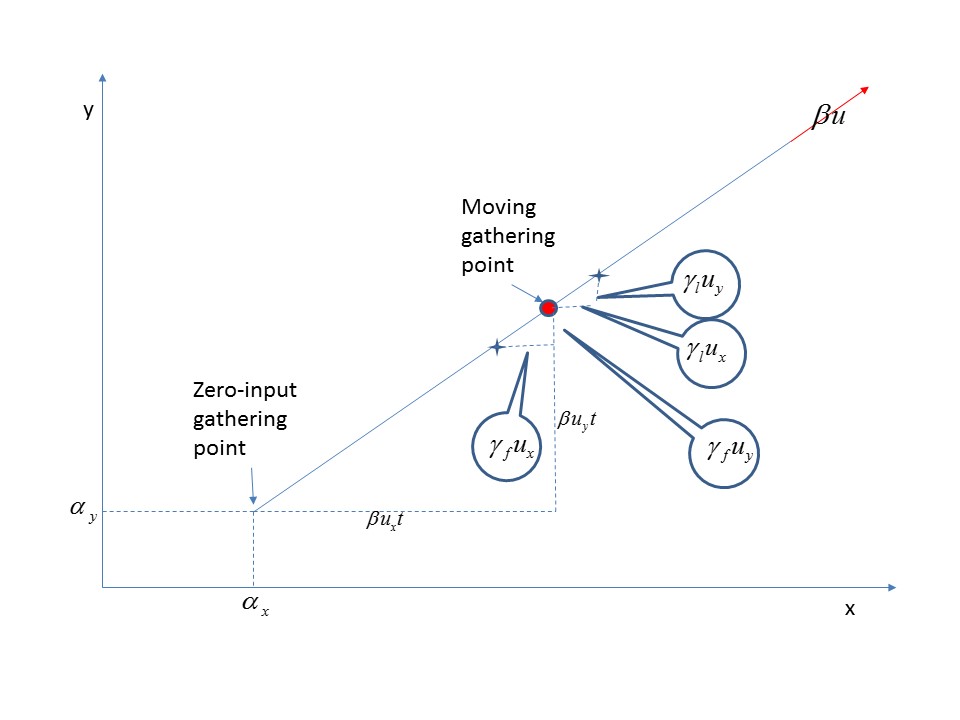}
\caption{Asymptotic dispersion of agents along the direction of $u$}\label{fig-dev}
\end{center}
\end{figure}

The vector of asymptotic deviations, s.t. $\varrho_i = \gamma_i \mathbf{u}$ is the deviation of agent $i$,  in  the $(x, y)$  space, from the (moving) consensus $\mathbf{\alpha}+\beta \mathbf{u} t$, where $\mathbf{\alpha} = (\alpha_x, \alpha_y)$. The agents align along  a line  in the direction of $\mathbf{u}$. The line is anchored at the zero-input gathering, or consensus, point $\mathbf{\alpha}$. Since $\gamma$ is time independent, the asymptotic dispersion of agents along this line is time independent,  as illustrated in Fig. \dref{fig-dev}. The swarm moves with velocity $\beta \mathbf{u}$.

\msubsection{Example of simulation results - Single time interval }{SimEx}

A single time interval of a piecewise constant system is equivalent to  a time-independent configuration with constant exogenous control and leaders. We consider a network of 5 agents and illustrate the group behaviour, for a constant $u$, both in case of incomplete visibility graph and of complete visibility graph. In these examples, the exogenous control is $\mathbf{u} = (10, 2)$. The initial positions $x(0), y(0)$ were once randomly selected in $[-50,50]$, and kept common for all runs.

\subsubsection{Incomplete visibility graph}\label{SimExIncomplete}
\begin{figure}
\begin{center}
\includegraphics[scale=0.6]{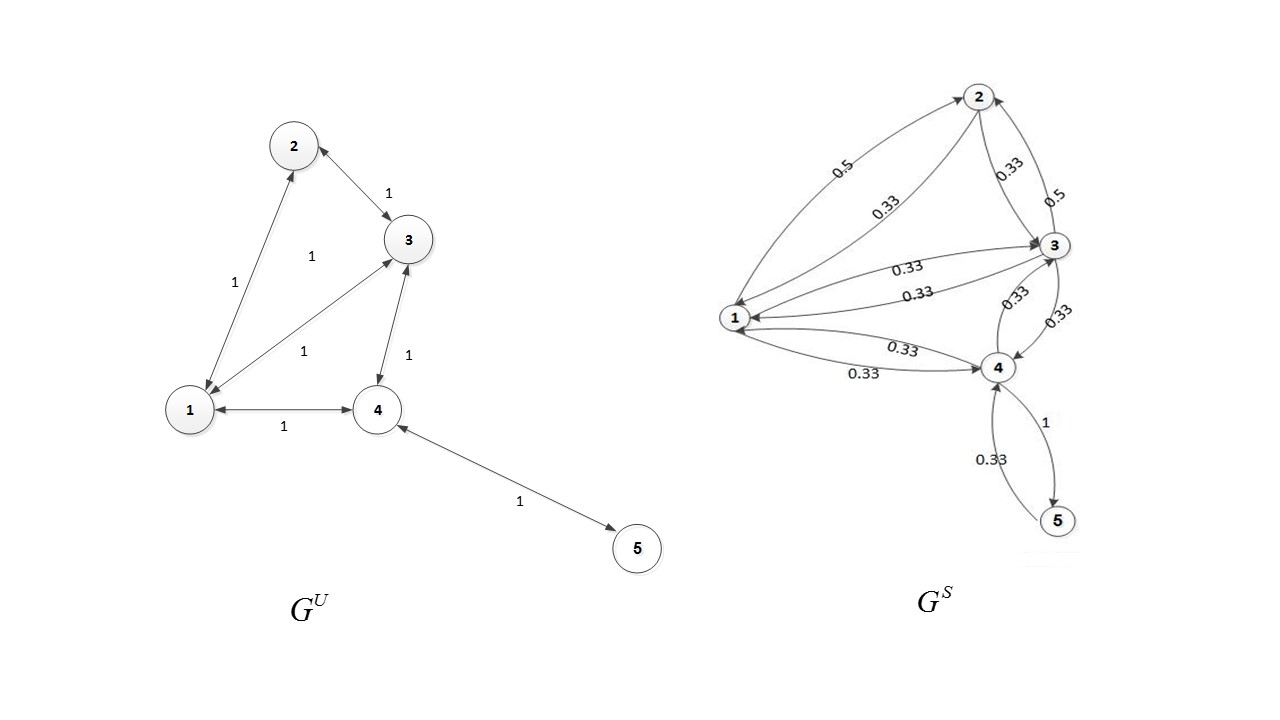}
\caption{Simulated pairwise interaction graph}\label{GraphExSim}
\end{center}
\end{figure}

\begin{figure}
\begin{center}
\includegraphics[scale=0.55]{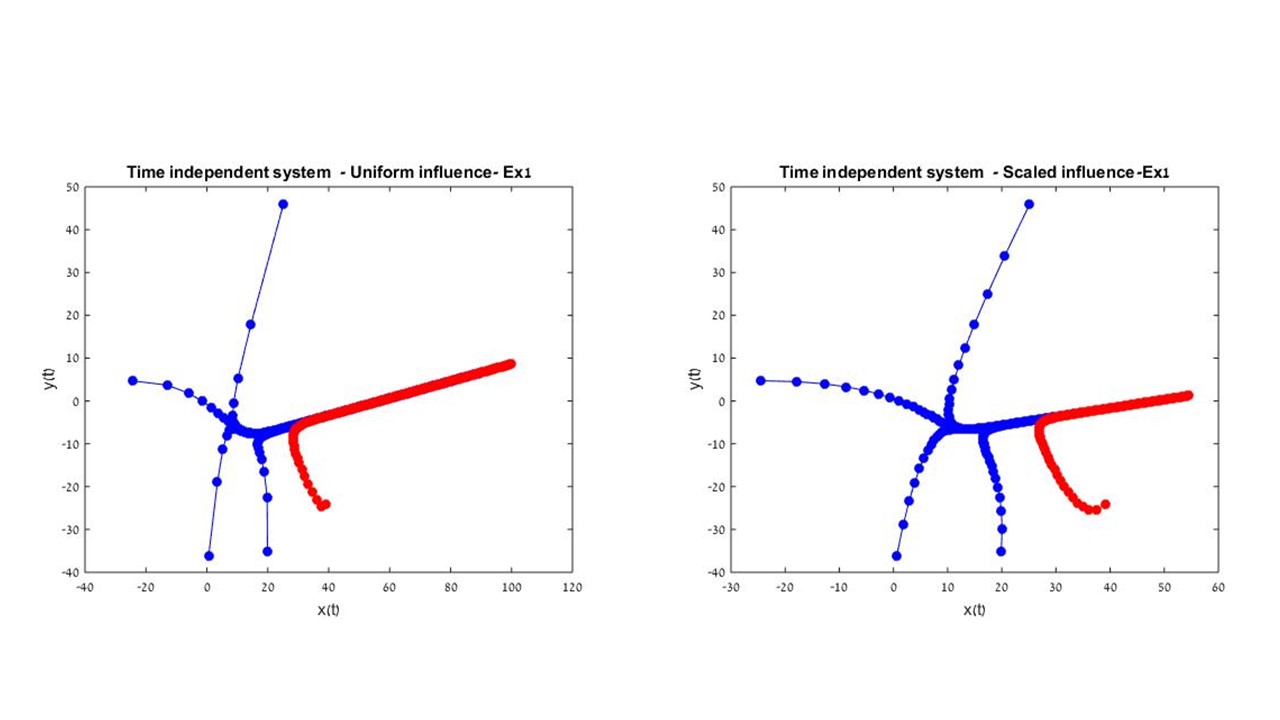}
\caption{Emergent dynamics in Ex1 with uniform and scaled influences}\label{ExSimDyn}
\end{center}
\end{figure}

\begin{figure}
\begin{center}
\includegraphics[scale=0.7]{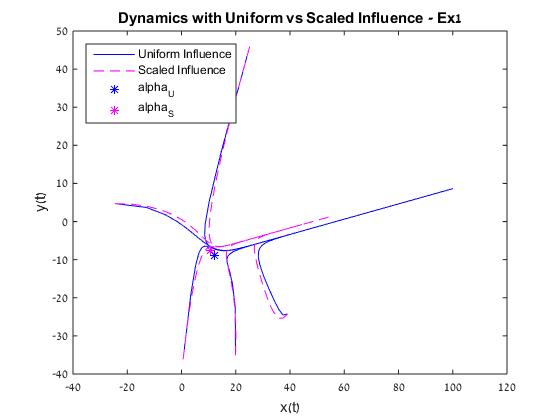}
\caption{Comparative view of emergent dynamics with uniform and scaled influences - Ex1}\label{Ex1Dyn}
\end{center}
\end{figure}

 In the examples in this section, we  illustrate the impact of the influence model applied by the agents and of the leader selection on the agents dynamics when the interaction graphs, $G^U$ and  $G^S$, are as illustrated in Fig. \dref{GraphExSim}.
Fig. \dref{ExSimDyn} shows the emergent dynamics of the agents when agent 5 detects the constant exogenous control, thus is the leader. This example will be named Ex1. In Fig. \dref{ExSimDyn} the leader is colored red and the followers are blue.
The agents are seen to asymptotically align, in both cases, along a line in the direction of $\mathbf{u}$, in this case a line with slope 0.2, as expected. The dots indicate the position of the units at consecutive times, t=1,2,3,... We can also see that, in this example, the collective speed of the agents with scaled influence is considerably lower than that of the agents with scaled influence. While the collective speed in the uniform case is $0.2 |u| $, corresponding to $\displaystyle \beta=\frac{n_l}{n}$ with $n_l=1, n=5$, for the scaled case it is only $0.0833 |u|$, corresponding to  $ \displaystyle \beta = \frac{\sum_{i \in N^l} d_i}{\sum_{i=1}^n d_i}$. \\
Fig. \dref{Ex1Dyn} shows a comparative view of the agents' dynamics in Ex1. Here again we see the difference in the collective speed with uniform vs scaled influence, but we also see that the agents' alignment lines are parallel and each is anchored at the corresponding zero-input gathering point.

\begin{figure}
\begin{center}
\includegraphics[scale=0.7]{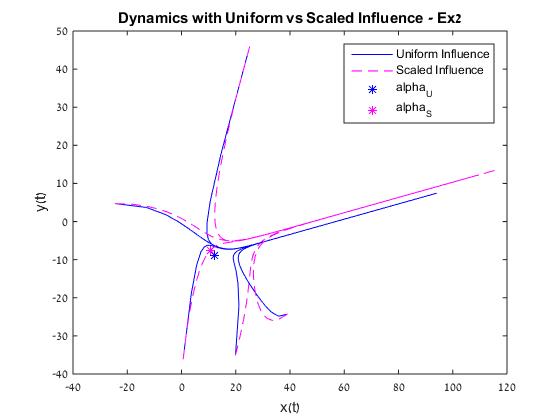}
\caption{Comparative view of emergent dynamics with uniform and scaled influences - Leader is agent 4 - Ex2}\label{Ex2Dyn}
\end{center}
\end{figure}

Note however that while for the uniform case the coefficient of the collective speed is a function of  only the number of leaders, for the scaled case it is also a function of the topology itself, i.e. of the number and distribution of links. Thus, by selecting now agent 4 instead of 5 as leader, we do not change the speed in the uniform case but increase it three times in the scaled case.  This brings the velocities of the agents with scaled influence to be larger than the ones with uniform influence, as illustrated in Fig. \dref{Ex2Dyn}

\begin{figure}
\begin{center}
\includegraphics[scale=0.5]{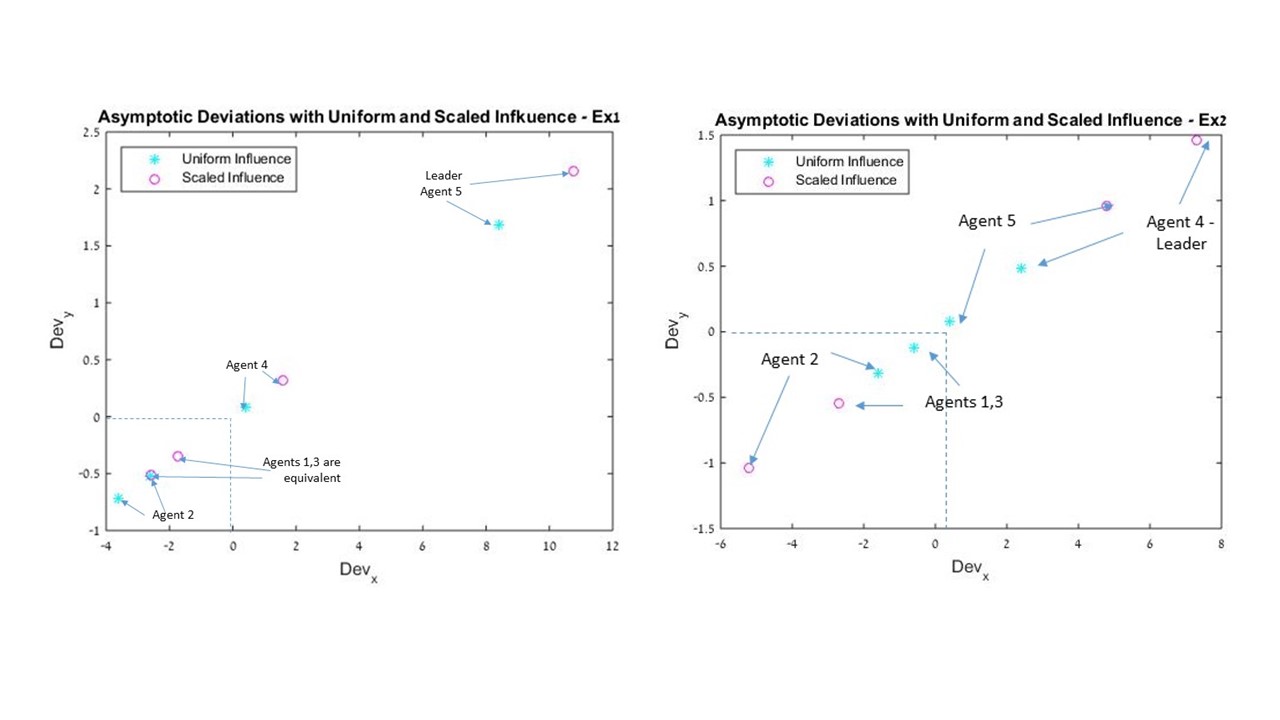}
\caption{Impact of leader selection on agents' asymptotic derivations relative to the moving consensus}\label{ExSimDev}
\end{center}
\end{figure}

\begin{figure}
\begin{center}
\includegraphics[scale=0.6]{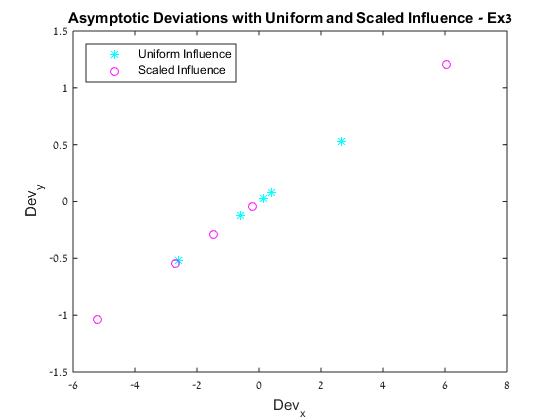}
\caption{Asymptotic derivations relative to the moving consensus when agent 1 is leader}\label{Ex3SimDev}
\end{center}
\end{figure}

Fig. \dref{ExSimDev} shows the asymptotic deviation of the agents relative to the moving gathering point, in both leader cases, agent 4 or agent 5. In both cases, agents 1 and 3 were equivalent, therefore had the same deviation, but this is not always the case. For example, if agent 1 is selected as leader, Ex3, there will be no equivalent agents, as shown in Fig \dref{Ex3SimDev}. Therefore, \textbf{equivalence is not preserved under change of leader}.

\begin{figure}
\begin{center}
\includegraphics[scale=0.6]{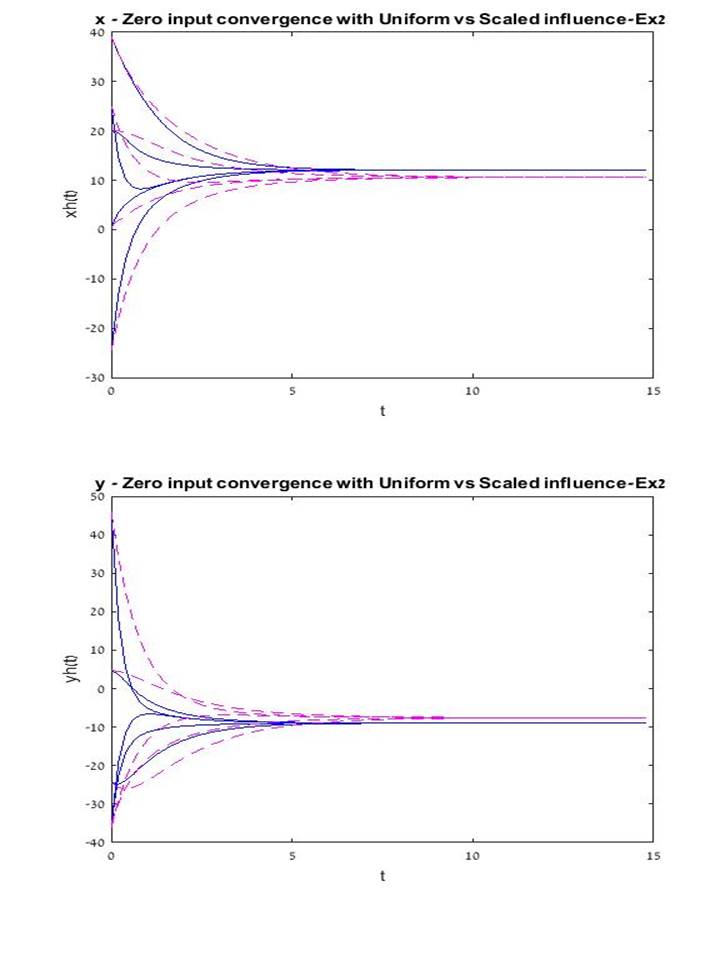}
\caption{Zero Input dynamics as a function of time}\label{ZeroInpDyn}
\end{center}
\end{figure}

Another issue to be considered is that of the impact of the influence model on the time of convergence.
Fig. \dref{ZeroInpDyn} shows the convergence to consensus for uniform and scaled dynamics, when no exogenous control is applied.  We see clearly that the convergence time with scaled influence is longer than that with uniform influence. The convergence time is a function of the first non-zero eigenvalue of the Laplacian, $\lambda_2$. In our examples  one has $\lambda^U_2=0.8299$ and  $\lambda^S_2=0.5657$. Since in these examples we only change leaders, $G^U$ and $G^S$  and the corresponding $\lambda_2$ do not change. Therefore the time of convergence, is identical for all 3 examples. However, we can say that the time of convergence with   scaled influence  is always at least the time of convergence with uniform influence, since
\begin{itemize}
  \item the eigenvalues of the Laplacian of $G^S$ are the eigenvalues of the \textbf{normalized Laplacian} of the corresponding graph with uniform influence, $G^U$. If we denote the eigenvalues of the normalized Laplacian by $\phi^U_i; i=1,.,n$ Then $\lambda^S_i=\phi^U_i; \quad i=1,...n$
  \item As shown by Butler in  \dcite{ButlerPhD}, Theorem 4
  \begin{equation*}
  \frac{1}{d_{max}} \lambda^U_i \leq \phi^U_i \leq \frac{1}{d_{min}} \lambda^U_i
\end{equation*}
where $d_{max}$ is the maximum degree and $d_{min}$ is the minimum degree of a vertex in $G_U$.
\end{itemize}
Thus,  $\lambda^S_2 \leq \lambda^U_2$ for any graph $G^U$ and corresponding $G^S$.

\subsubsection{Complete visibility graph}\label{SimExComplete}
\begin{figure}
\begin{center}
\includegraphics[scale=0.7]{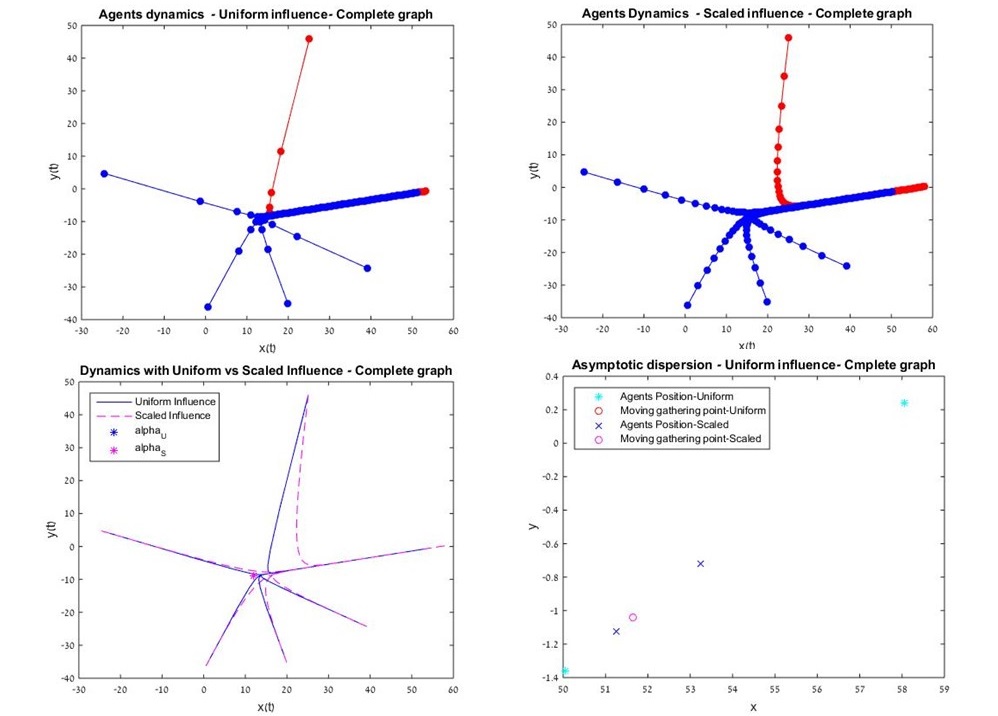}
\caption{Comparative view of emergent dynamics with uniform and scaled influences when the visibility graph is complete }\label{Ex5nComplete}
\end{center}
\end{figure}

In this example we assume a network of 5 agents with complete visibility graphs. Fig. \dref{Ex5nComplete} illustrates the emergent behavior in case of scaled influence vs uniform influence and shows that
\begin{itemize}
  \item the zero input gathering point coincides
  \item the position of the moving gathering point coincides, thus the collective velocity coincides
  \item the dispersion of the agents around the moving gathering point is larger when the influence is scaled, in fact exactly 4 times larger, as expected
  \item the time for convergence to (moving) consensus is larger in case of scaled influence, as expected

\end{itemize}

\newpage

\msec{Multiple time intervals}{MultiIntEx}
In the previous sections we considered a single time interval were the system is time independent, i.e. the visibility graph  and the corresponding Laplacian $L$, the leaders and the exogenous control $u$ are constant along the interval. We showed the dependence of the emergent behavior on the influence model, scaled or uniform, in case of complete and incomplete visibility graphs. We now consider a sequence of time intervals, $[t_k, t_{k+1})$, where a new time interval is triggered by changes in one of the system parameters, the broadcast control $u$, the agents detecting the broadcast control, i.e. the leaders, or the visibility graph and the corresponding Laplacian.
In order for the group behavior along multiple intervals to be a concatenation of dynamics along single intervals, with the end states of one interval becoming the start states of the next interval, we need to ensure that the visibility graph remains strongly connected. In this section we derive sufficient conditions, which are conditions for never losing friends, i.e. for initially adjacent pairs of agents to remain adjacent. Thus, we require $G_k \subseteq G_{k+1}$ and consider  two cases of visibility graphs, each for uniform and scaled influences:
\begin{enumerate}
  \item $G_k$ is complete

  \item $G_k$ is incomplete
\end{enumerate}

Changes in the visibility graph are state dependent, i.e. a link $(i,j)$ exists at time $t$ iff $|p_i(t)-p_j(t)| \leq R$, where $R$ is the visibility, or sensing, range.   In the next sections we derive \textbf{conditions for \emph{never losing neighbors}} and illustrate their effect by simulations. We show that
\begin{enumerate}
  \item if the initial interactions graph is complete and the sensing range is $R$, then

\begin{itemize}
  \item Follower to follower distance and leader to leader distance are monotonically decreasing, therefore initial links are preserved, independently of the value of $u$, for both uniform influence ( Lemma \dref{L-L2Lpreserve}) and scaled influence (Theorem \dref{T-ScaledComplete}a)
  \item Leader to follower links can be proven to be maintained only if the exogenous control is limited to
      \begin{enumerate}
        \item $|u| \leq nR$, for the \textbf{uniform case}, Theorem \dref{T-L2Fpreserve}
        \item $\displaystyle |u| \leq \frac{n}{n-1} R$, for the \textbf{scaled case}, Theorem \dref{T-ScaledComplete}b
      \end{enumerate}
\end{itemize}

Recalling that in  case of complete graphs, all leaders asymptotically move together (one moving gathering point) and all followers move together (at another point gathering point) we note that the distance between any leader to any follower tends to $d_{lf} = (\gamma_l-\gamma_f)|u|$ (cf. section \dref{Dev-mean}) and thus preserving the link requires $d_{lf} \leq R$. Since for the complete visibility case $\gamma_l^S=(n-1)\gamma_l^U$ and $\gamma_f^S=(n-1)\gamma_f^U$  the ratio between the bounds on $u$ shown above becomes evident.\\
In section \dref{LimitedComplete-Ex} we show an example of emergent dynamics when $|u|$ is within bounds and another example where $|u|$ exceeds the derived limit
\item if the initial graph is incomplete then
\begin{itemize}
   \item conditions for never losing friends  are tightly related to the graph topology.\\
    Since an external controller does not know the time-dependent topology these are not useful in practice.
   \item for a general form of incomplete graph, bounds cannot be derived or are too loose to be useful.
 \end{itemize}
 Although useful bounds could not be derived, many simulations show that if the interactions graph starts as an incomplete graph, when inputting $u$ such that $|u| < R$, the agents fast converge to a complete graph, as shown by some examples  in section \dref{IncompleteEx}.
\end{enumerate}

\msubsection{Conditions for maintaining complete graphs }{never-losing2}

Denote by $\delta_{ij}$ the distance between two adjacent agents $i$ and $j$
\begin{equation}\label{dist2}
  \delta_{ij}=\delta_{ji}= |p_i-p_j|= \sqrt{(p_i-p_j)^T(p_i-p_j)}
\end{equation}
Since the movement of the agents is smooth, a necessary and sufficient condition for the link to be always preserved is
 $d{\delta}_{ij}/dt \leq 0$ when $\delta_{ij}=R$, or equivalently $d(\delta_{ij}^2)/dt \leq 0$, when $\delta_{ij}=R$.

 Note that $d(\delta_{ij}^2)/dt$  has the same sign as $d{\delta}_{ij}/dt$ and is defined on all of $\mathbb{R}^n$ while $d\delta_{ij}/dt$ is not defined when $p_i=p_j$.

We have
\begin{equation}\label{deriv-dist}
\frac{d(\delta_{ij}^2)}{dt} =2 \delta_{ij} \dot{\delta_{ij}} = 2(p_i-p_j)^T(\dot{p_i}-\dot{p_j})
\end{equation}

\subsubsection{Uniform influence}\mbox{}\\
 If the visibility graph is a complete graph with uniform influences then $\sigma_{ji}=1; \quad \forall i,j$ and equations (\dref{gen-SelfDyn}), (\dref{gen-LeadDyn})  can be combined  and reformulated as
\begin{equation}\label{gen-dyn}
\dot{p_i}=-n_i p_i+\sum_{k \in N_i}  p_k+b_i u
\end{equation}
where  $p_i = (x_i \quad y_i)^T$ , $u =(u_x \quad u_y)^T$, $N_i$ is the neighborhood of agent $i$, $n_i=|N_i|$ and $b_i= 1$ if $i$ is a leader and 0 if $i$ is a follower.
Since for a complete graph $n_i=n-1; \quad \forall i$, Eq. (\dref{gen-dyn})can be rewritten as
\begin{equation*}
\dot{p_i}=-(n-1) p_i+\sum_{k =1; k \neq i}^n p_k + b_i u
\end{equation*}
and similarly for $\dot{p_j}$, where $n$ is the total number of agents.
Thus
\begin{equation*}
\dot{p}_i-\dot{p}_j=-n(p_i-p_j)+(b_i-b_j) u
\end{equation*}
Denoting $b_{ij}=(b_i-b_j)$, we have
\begin{equation}\dlabel{eq-bij}
  b_{ij} = \begin{cases}0 ; \text{    if both    } i,j \text{  are followers or both are leaders}\\
  1 ; \text{    if   } i \text{  is leader and   } j \text{  is follower}
  \end{cases}
\end{equation}
Substituting in eq. (\dref{deriv-dist}) one obtains
\begin{equation}\label{Uni-dist2}
  \frac{d\delta_{ij}^2}{dt} = -2n \delta_{ij}^2 +2(p_i-p_j)^T b_{ij} u
\end{equation}

\LB{L-L2Lpreserve}
If the visibility graph of $n$ agents with uniform influences is a complete graph, then all leader to leader and  follower to follower links are preserved, independently of the externally applied control $u$.
\LE
\PB
If $i$ and $j$ in (\dref{Uni-dist2})are both leaders or both followers, then $b_{ij}=0$ and thus eq. (\dref{Uni-dist2}) becomes
\begin{equation*}
  \frac{d\delta_{ij}^2}{dt} = -2n \delta_{ij}^2
\end{equation*}
with the solution
\begin{equation*}
 \delta_{ij}^2 (t) = e^{-2nt} \delta_{ij}^2 (0)
\end{equation*}
Thus, $\delta_{ij}(t)$  decreases monotonically from the initial condition.
\PE

We shall consider now the \textbf{case when $i$ is a leader and $j$ is a follower}.
 \TB{T-L2Fpreserve}
If the visibility graph of $n$ agents with uniform influences is a complete graph and the magnitude of the exogenous control is limited to $|u| \leq nR$, then the connection of a leader $i$ and a follower $j$ is never lost.
 \TE
\PB
When $i$ is a leader and $j$ is a follower Eq. (\dref{Uni-dist2}) becomes
\begin{equation}\label{Uni-distL2F}
  \frac{d\delta_{ij}^2(t)}{dt} = -2n \delta_{ij}(t)^2 +2(p_i(t)-p_j(t))^T  u
\end{equation}
But $(p_i-p_j)^T u = \langle (p_i-p_j), u \rangle \leq |p_i-p_j||u| =\delta_{ij}|u|$.  Consider a time $t_1$ when for the first time $\delta_{ij}(t_1)=R$ holds. Since $|u| \leq n R$ for all $t$, we obtain  $\displaystyle \frac{d\delta_{ij}^2}{dt}(t_1) \leq 0$. Therefore, when $i$ is a leader and $j$ is a follower, if  $|u| \leq nR$, then $\delta_{ij}(t_1) \leq R$ and by induction this result holds for all $t$.
\PE

\subsubsection{Scaled influence}\mbox{}\\
If the visibility graph is a complete graph with scaled influences then $\sigma_{ji}=\frac{1}{n-1}; \quad \forall i,j$ and  equations (\dref{gen-SelfDyn}), (\dref{gen-LeadDyn})  can be combined  and reformulated as
\begin{eqnarray}\dlabel{Scaled-dyn}
\dot{p_i} & = &\sum_{k \in N_i} \frac{1}{n-1}( p_k-p_i)+b_i u\\
& = & -p_i+ \frac{1}{n-1}  \sum_{k=1,k \neq i}^n p_k +b_i u\\
& = & -\frac{n}{n-1} p_i + \sum_{k=1}^n p_k+b_i u
\end{eqnarray}
and similarly for $\dot{p_j}$.  Thus, we have
\begin{equation}\dlabel{Deriv-Diff}
  \dot{p_i}-\dot{p_j}=-\frac{n}{n-1}( p_i-p_j) +b_{ij}u
\end{equation}
where $b_{ij}$ as in (\dref{eq-bij}). Therefore, multiplying (\dref{Deriv-Diff}) from the left by $2(p_i-p_j)^T$ we obtain:
\begin{equation}\dlabel{Scaled-deriv}
  \frac{d (\delta_{ij}^2)}{dt}=-2 \frac{n}{n-1} \delta_{ij}^2 + 2 b_{ij} (p_i-p_j)^T u
\end{equation}
The dynamics of $\delta_{ij}(t)$, as expressed by (\dref{Scaled-deriv}), for the case when $i$ and $j$ are both followers or both leaders and for the case when $i$ is a leader and $j$ is a follower are summarized by the following theorem:
\TB{T-ScaledComplete}
If the visibility graph of $n$ agents with scaled influences is a complete graph, then
\hB
\titem{a}	All follower-to-follower links and all leader-to-leader links are monotonically decreasing from the initial conditions, thus these links are preserved, independently of the exogenous control $u$
\titem{b}	If the exogenous control satisfies $|u| \leq  \frac{n}{n-1} R$ then all Leader-to-Follower links are preserved
\hE
\TE
\PB
\tref{a} If $i$ and $j$ are both followers or leaders then $b_{ij}=0$. By substituting in (\dref{Scaled-deriv})and solving the resulting homogenous equation, one obtains
\begin{equation*}
  \delta_{ij}^2(t)=e^{-2 \frac{n}{n-1}t} \delta_{ij}^2(0)
\end{equation*}
Thus, $\delta_{ij}$ monotonically decreases for any two followers or any two leaders and therefore the  link is preserved.\\
\tref{b}  If $i$ is a leader and $j$ is a follower then  $b_{ij}=1$.
Substituting this in eq. (\dref{Scaled-deriv}) and letting again $t_1$ be the first time when $\delta_{ij}(t_1)=R$ we obtain
\begin{equation*}
   \frac{d( \delta_{ij}^2)}{dt}(t_1) \leq -2 \frac{n}{n-1} R^2 + 2R|u|
\end{equation*}
where we used the inequality for inner products $(p_i-p_j)^T u \leq \delta_{ij} |u|$.
If $|u| \leq \frac{n}{n-1} R$ for all $t$, then $\displaystyle \frac{d( \delta_{ij}^2)}{dt}(t_1) \leq 0$, and thus, by induction, the leader-to-follower link is preserved for all $t$.
\PE

\subsubsection{Simulation examples - Effect of $|u|$ on complete graph preservation}\label{LimitedComplete-Ex}
\begin{figure}
\begin{center}
\includegraphics[scale=0.6]{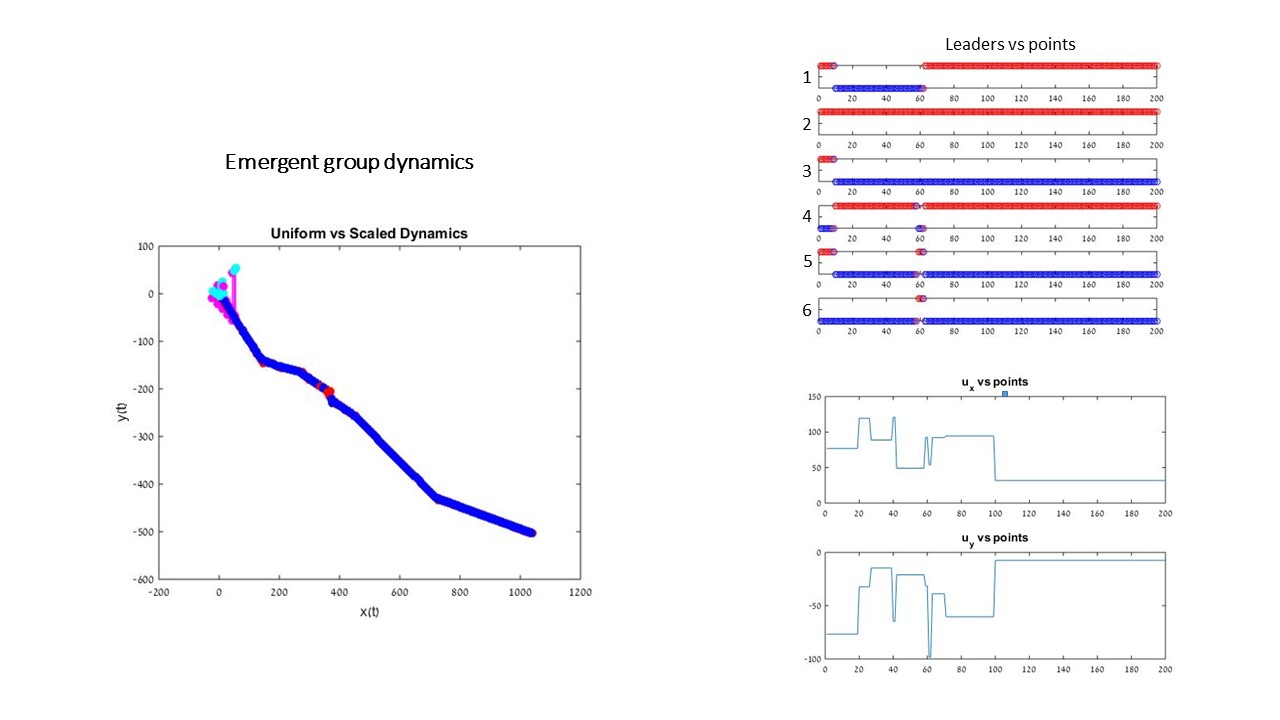}
\caption{Group dynamics with limited visibility and high $|u|$ -Ex1}\label{fig-LimitedVisDyn-Ex1}
\end{center}
\end{figure}

We illustrate the emergent behavior of a group of 6 agents with initially complete visibility graph, $R=50$,  $u$ and leaders randomly selected, as shown. In the first example, Ex1, where $u_x, u_y$ have random values in the range $[100, 100]$, at $t=5sec$, the restriction on $|u|$ for the scaled case is not satisfied while for the uniform case it is satisfied. Thus,  when the scaled influence is applied, the graph splits in two parts (after ~ 5 sec), leaders forming one component and followers forming the other component. When the split occurs, the agents dynamics simulation is stopped. Thus, in Fig. \dref{fig-LimitedVisDyn-Ex1}, the dynamics with scaled influence (cyan and magenta) stopped soon after the beginning of the run (at t=5 sec) while the dynamics with uniform influence (blue and red) evolved for the whole requested period (40 sec).

When the range of $u_x, u_y$ is reduced to within the limits, all links are preserved, as illustrated in Ex2, where $u_x \in [-20,20]$, $u_y \in [-10,10]$. The leaders were again randomly selected.
\begin{figure}
\begin{center}
\includegraphics[scale=0.6]{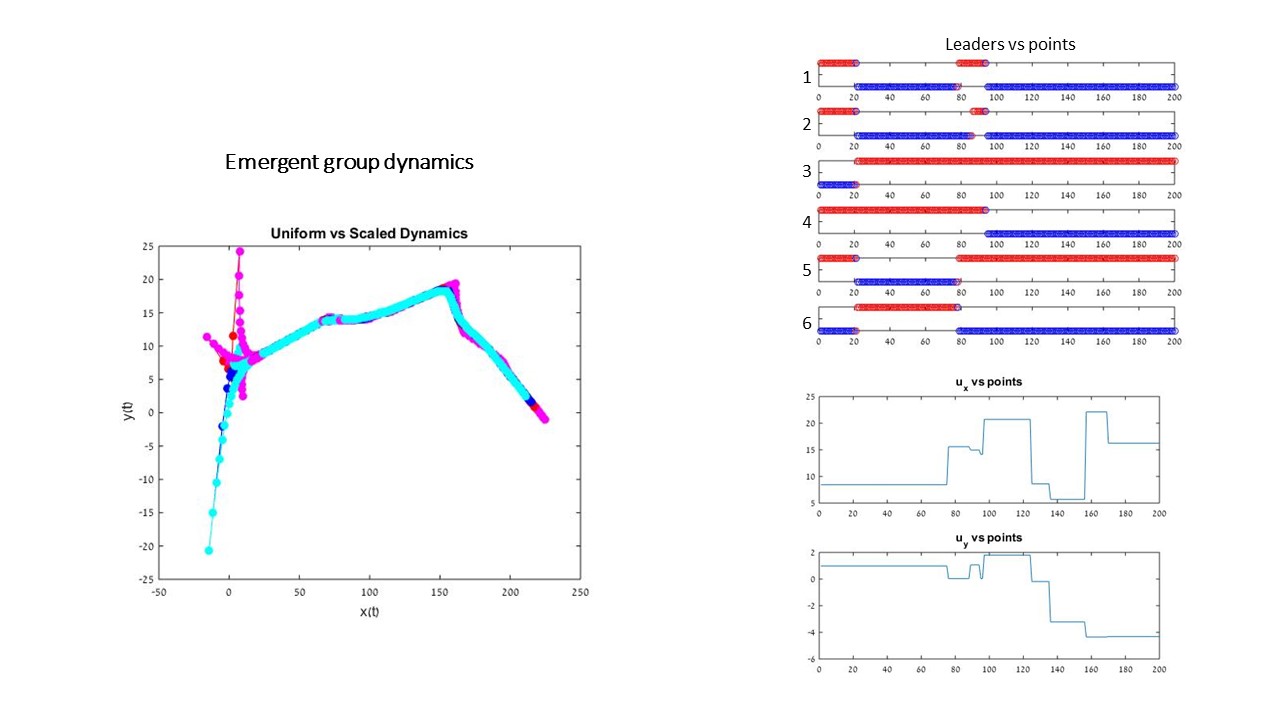}
\caption{Group dynamics with limited visibility and $|u|$ within limits - Ex2}\label{fig-LimitedVisDyn-Ex2}
\end{center}
\end{figure}

In this case the initial complete graph is preserved for both the scaled and the uniform influence and the agents complete the run in both cases.

\msubsection{Conditions for never losing friends when visibility graph is incomplete}{LimitedIncomplete}
In this section we show that for a general case of incomplete graphs, the "never losing friends" requirement imposes stringent conditions on the topology. Moreover, we show, by examples, that for specific topologies these conditions are too stringent and the property can be proven under relaxed restrictions.\\
We employ the following notations:
\begin{itemize}
   \item $\mathbb{N}^f$ denotes the set of followers
   \item $\mathbb{N}^l$ denotes the set of leaders
   \item $n_f = |\mathbb{N}^f|$  is the number of followers
   \item $n_l= |\mathbb{N}^l|$ is  the number of leaders
   \item $n = n_f+n_l$ is the total number of agents
   \item the set of followers adjacent to an agent $i$, leader or follower, is denoted by  $\mathbb{N}_i^f $\\
   $\mathbb{N}_i^f \subseteq \mathbb{N}^f$
   \item the set of leaders adjacent to an agent $i$, leader or follower, is denoted by  $\mathbb{N}_i^l $\\
  $\mathbb{N}_i^l \subseteq \mathbb{N}^l$
   \item $n_{il}$ is the number of leaders agent $i$ is connected to, $n_{il}=|\mathbb{N}_i^l |$
   \item $n_{if}$ is the number of followers agent $i$ is connected to, $n_{if}=|\mathbb{N}_i^f |$
   \item $\mathbb{N}_i$ is the neighborhood of $i$, $\mathbb{N}_i = \mathbb{N}_i^f \bigcup \mathbb{N}_i^l$
   \item $n_i$ is the size of the neighborhood of $i$, $n_i = n_{il} + n_{if}$
   \item $\mathbb{N}_{(ij)}^l$ denotes the set of leaders adjacent to both $i$ and $j$
   \item $n_{(ij)l}$ is the number of leaders that have a link to both $i$ and $j$
   \item $\mathbb{N}_{(ij)}^f$ denotes the set of followers adjacent to both $i$ and $j$
   \item $n_{(ij)f}$ is the number of followers that have a link to both $i$ and $j$

\end{itemize}

Can we find conditions on the topology and on $|u|$ s.t. any two nodes, $i$ and $j$, initially connected, i.e. satisfying  $\delta_{ij}(0) = |p_i(0)-p_j(0)| \leq R$ will remain connected, i.e. will  satisfy $\frac{d}{dt}\delta_{ij}^2 \leq 0$ when $\delta_{ij}(t)=R$ ? We consider $t_1$, the first time when for one or more links holds $\delta_{ij}(t_1)=R$ and derive conditions for $\frac{d}{dt}\delta_{ij}^2(t_1) \leq 0$ for each link type and each influence type.

\subsubsection{General incomplete topology  - Uniform case}\label{IncompleteGeneral}

If each agent applies the movement equation with uniform influence, then we have
\begin{itemize}
  \item for followers
  \begin{equation}\dlabel{ML-FolDynU}
      \dot{p}_i  =  -\sum_{k \in \mathbb{N}_i^f} (p_i-p_k) - \sum_{k \in \mathbb{N}_i^l} (p_i-p_k); \quad i \in \mathbb{N}^f
\end{equation}
  \item for leaders
  \begin{equation}\dlabel{ML-LeadDynU}
      \dot{p}_i=-\sum_{k \in \mathbb{N}_i^l} (p_i-p_k) - \sum_{k \in \mathbb{N}_i^f} (p_i-p_k) + u; \quad i \in \mathbb{N}^l
\end{equation}
\end{itemize}

\begin{enumerate}
  \item If $i$ and $j$ are \textbf{\emph{both followers}}, then applying (\dref{ML-FolDynU}) to $i$ and $j$ we obtain
\begin{equation}\dlabel{eq-F2F-U}
   \begin{aligned}
   \dot{p}_i - \dot{p}_j   = & -\sum_{k \in \mathbb{N}_i^f} (p_i-p_k) - \sum_{k \in \mathbb{N}_i^l} (p_i-p_k) \\
   & +\sum_{k \in \mathbb{N}_j^f} (p_j-p_k) + \sum_{k \in \mathbb{N}_j^l} (p_j-p_k)
   \end{aligned}
\end{equation}
Separating now the set of neighbors common to $i$ and $j$ from the set of private neighbors to $i$ or $j$ and using
\begin{eqnarray*}
  \mathbb{N}_i^f & = &  \mathbb{N}_{ij}^f +  \mathbb{N}_i^f \backslash \mathbb{N}_{ij}^f\\
   \mathbb{N}_i^l & = &  \mathbb{N}_{ij}^l +  \mathbb{N}_i^l \backslash \mathbb{N}_{ij}^l
\end{eqnarray*}
and similarly for $j$, we obtain
\begin{equation*}
   \begin{aligned}
   \dot{p}_i - \dot{p}_j   = & -n_{(ij)f} p_i + \sum_{k \in \mathbb{N}_{ij}^f} p_k -\sum_{k \in \mathbb{N}_i^f \backslash \mathbb{N}_{ij}^f}  (p_i-p_k) \\
    & -n_{(ij)l} p_i + \sum_{k \in \mathbb{N}_{ij}^l} p_k -\sum_{k \in \mathbb{N}_i^l \backslash \mathbb{N}_{ij}^l}  (p_i-p_k) \\
    & +n_{(ij)f} p_j - \sum_{k \in \mathbb{N}_{ij}^f} p_k +\sum_{k \in \mathbb{N}_j^f \backslash \mathbb{N}_{ij}^f}  (p_j-p_k) \\
    & + n_{(ij)l} p_j -\sum_{k \in \mathbb{N}_{ij}^l} p_k +\sum_{k \in \mathbb{N}_j^l \backslash \mathbb{N}_{ij}^l}  (p_j-p_k)\\
    = & -(n_{(ij)f}+n_{(ij)l})(p_i-p_j) - \sum_{k \in \mathbb{N}_i^f \backslash \mathbb{N}_{ij}^f}  (p_i-p_k) -\sum_{k \in \mathbb{N}_i^l \backslash \mathbb{N}_{ij}^l}  (p_i-p_k)\\
     & +\sum_{k \in \mathbb{N}_j^f \backslash \mathbb{N}_{ij}^f}  (p_j-p_k)+\sum_{k \in \mathbb{N}_j^l \backslash \mathbb{N}_{ij}^l}  (p_j-p_k)
   \end{aligned}
\end{equation*}

Consider now the time $t_1$, the first time when one or more links satisfy $\delta_{ij}(t_1)=R$ and let the considered follower to follower link be among them. Then we have $\delta_{ij}(t_1)=R$ and $\delta_{ik}(t_1) = |p_i(t_1)-p_k(t_1)| \leq R; \quad \forall k \in \mathbb{N}_i^f \quad \text{and   } k \in \mathbb{N}_i^l $ and similarly for $j$.
Recalling that $\displaystyle \frac{d}{dt}\delta_{ij}^2  =  2(p_i-p_j)^T ( \dot{p}_i - \dot{p}_j)$ we obtain at
\begin{equation}\dlabel{F2Fderiv}
   \begin{aligned}
  \frac{1}{2} \frac{d}{dt}\delta_{ij}^2(t_1)  = & -(n_{(ij)f}+n_{(ij)l})R^2 +\sum_{k \in \mathbb{N}_i^f \backslash \mathbb{N}_{ij}^f} (p_i-p_j)^T (p_k-p_i) +\sum_{k \in \mathbb{N}_i^l \backslash \mathbb{N}_{ij}^l} (p_i-p_j)^T (p_k-p_i)\\
     & +\sum_{k \in \mathbb{N}_j^f \backslash \mathbb{N}_{ij}^f} (p_i-p_j)^T (p_j-p_k)+\sum_{k \in \mathbb{N}_j^l \backslash \mathbb{N}_{ij}^l} (p_i-p_j)^T (p_j-p_k)
   \end{aligned}
\end{equation}
   Using now $V_1^T V_2 \leq |V_1| |V_2|$    we can write eq. (\dref{F2Fderiv}) as
\begin{equation}\dlabel{F2Fderiv2}
   \begin{aligned}
  \frac{1}{2} \frac{d}{dt}\delta_{ij}^2  \leq & -(n_{(ij)f}+n_{(ij)l})R^2 +(n_{if}-n_{(ij)f}) R^2+ (n_{il}-n_{(ij)l}) R^2\\
     & + (n_{jf}-n_{(ij)f}) R^2+ (n_{jl}-n_{(ij)l}) R^2\\
     =& [ n_i+n_j-3 n_{(ij)f} - 3 n_{(ij)l}]R^2
   \end{aligned}
\end{equation}
where we used
\begin{equation*}
  n_i =n_{if} + n_{il}
\end{equation*}
and similarly for $j$.\\
  Thus, if $n_i+n_j \leq 3 n_{(ij)f} + 3 n_{(ij)l}$ is satisfied then $ \frac{d}{dt}\delta_{ij}^2(t_1) \leq 0$ when $i, j \in \mathbb{N}^f$.

  \item If $i$ and $j$ are \textbf{\emph{both leaders}}, then applying (\dref{ML-LeadDynU}) to $i$ and $j$ we obtain
\begin{equation}\dlabel{eq-L2L-U}
   \begin{aligned}
   \dot{p}_i - \dot{p}_j   = & -\sum_{k \in \mathbb{N}_i^f} (p_i-p_k) - \sum_{k \in \mathbb{N}_i^l} (p_i-p_k) + u\\
   & +\sum_{k \in \mathbb{N}_j^f} (p_j-p_k) + \sum_{k \in \mathbb{N}_j^l} (p_j-p_k) -u
   \end{aligned}
\end{equation}
 Since $u$ is common to $i$ and $j$ eq. (\dref{eq-L2L-U}) reduces to eq. (\dref{eq-F2F-U}) and therefore we obtain the same condition for never losing neighbors:
 if $n_i+n_j \leq 3 n_{(ij)f} + 3 n_{(ij)l}$ is satisfied then $ \frac{d}{dt}\delta_{ij}^2(t_1) \leq 0$ when $i, j \in \mathbb{N}^l$ and at $t_1$  $\delta_{ij}=R$ and $\delta_{ik} \leq R$, $\delta_{jk} \leq R \quad \forall k \neq i, k \neq j$

  \item If \textbf{\emph{$i$ is leader and $j$ is follower}}, then
  from (\dref{ML-LeadDynU}) for $i$ and (\dref{ML-FolDynU}) for $j$ we obtain
  \begin{equation*}
   \begin{aligned}
   \dot{p}_i - \dot{p}_j   = & -\sum_{k \in \mathbb{N}_i^f} (p_i-p_k) - \sum_{k \in \mathbb{N}_i^l} (p_i-p_k) + u \\
   & +\sum_{k \in \mathbb{N}_j^f} (p_j-p_k) + \sum_{k \in \mathbb{N}_j^l} (p_j-p_k)
   \end{aligned}
\end{equation*}
which, following the same technique as above, reduces to
\begin{equation*}
   \begin{aligned}
   \dot{p}_i - \dot{p}_j   =  & -(n_{(ij)f}+n_{(ij)l})(p_i-p_j) +\sum_{k \in \mathbb{N}_i^f \backslash \mathbb{N}_{ij}^f}  (p_i-p_k) -\sum_{k \in \mathbb{N}_i^l \backslash \mathbb{N}_{ij}^l}  (p_i-p_k) + u\\
     & +\sum_{k \in \mathbb{N}_j^f \backslash \mathbb{N}_{ij}^f}  (p_j-p_k)+\sum_{k \in \mathbb{N}_j^l \backslash \mathbb{N}_{ij}^l}  (p_j-p_k)
   \end{aligned}
\end{equation*}
\end{enumerate}
\begin{equation}\dlabel{L2Fderiv}
   \begin{aligned}
  \frac{1}{2} \frac{d}{dt}\delta_{ij}^2  \leq & -(n_{(ij)f}+n_{(ij)l})R^2 +(n_{if}-n_{(ij)f}) R^2+ (n_{il}-n_{(ij)l}) R^2 + |u|R\\
     & + (n_{jf}-n_{(ij)f}) R^2+ (n_{jl}-n_{(ij)l}) R^2\\
     =& [ n_i+n_j-3 n_{(ij)f} - 3 n_{(ij)l}]R^2 +|u|R
   \end{aligned}
\end{equation}
Thus, if $3 ( n_{(ij)f} +  n_{(ij)l} ) \geq (n_i+n_j)  $ and $|u| \leq [ 3 ( n_{(ij)f} +  n_{(ij)l} ) - (n_i+n_j)]R$ then for $i \in \mathbb{N}^l$ and $j \in \mathbb{N}^f$, $\delta_{ij}^2 \leq 0$ when  $\delta_{ij}=R$.\\
All of the above results can be summarized by the following theorem:
\TB{IncompleteGen}
Given a group of agents with connected visibility graph and uniform influence, any link $(i,j)$ satisfying the following conditions will be preserved
\begin{enumerate}
  \item ( $i \in N^f$ and $j \in N^f$) or ($i \in N^l$ and $j \in N^l$) and $n_i+n_j \leq 3 n_{(ij)f} + 3 n_{(ij)l}$, independently of the exogenous control

  \item if $i \in N^l$ and $j \in N^f$  and $n_i+n_j \leq 3 n_{(ij)f} + 3 n_{(ij)l}$
  then an input $u$ that satisfies
  \begin{equation*}
    |u| \leq[ 3 (n_{(ij)f}+ n_{(ij)l})-(n_i+n_j)] R
  \end{equation*}
  will ensure  the link preservation
\end{enumerate}
where
\begin{itemize}
  \item $n_{(ij)l}$ is the number of leaders that have a link to both $i$ and $j$
  \item $n_{(ij)f}$ is the number of followers that have a link to both $i$ and $j$
  \item $n_i$ is the number of nodes adjacent to $i$
  \item $n_j$ is the number of nodes adjacent to $j$
\end{itemize}

\TE
Note that these conditions are not useful to us since the controller is unaware of the time varying and random values needed in the quantity limiting the control speed $|u|$, in order to ensure the preservation of all initial visibility links.

\paragraph{Effect of assuming a Specific topology on conditions for  never losing neighbors}\mbox{}\\

In this section we show that if a specific topology is assumed, then the bounds derived in section \dref{IncompleteGeneral}, for a general incomplete graph with uniform influences, can be tightened. We illustrate the effect on the example shown in Fig. \dref{fig-Ex1-Incomplete}. A more general example, although with some specific features, is shown in Appendix \dref{incomplete}, where all leaders form a complete graph and all followers form a complete graph.

  \begin{figure}
\begin{center}
\includegraphics[scale=0.6]{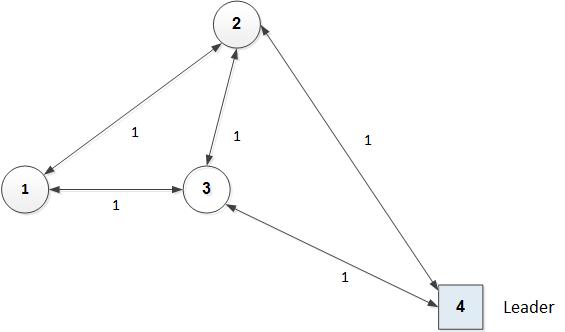}
\caption{Ex1 - specific case of incomplete graph}\label{fig-Ex1-Incomplete}
\end{center}
\end{figure}
 If we consider link $(4,3)$ and apply theorem \dref{IncompleteGen}, we have:
 \begin{itemize}
   \item $n_4=2$
   \item $n_3=3$
   \item $n_{(34)f}=1$
   \item $n_{(34)l}=0$
 \end{itemize}
Thus the condition   $n_3+n_4 \leq 3 n_{(34)f} + 3 n_{(34)l}$ does not hold and there is no $u$ that will ensure that link $(4,3)$ is preserved. However, if we consider the particular structure of the graph we obtain:
\begin{equation*}
\begin{aligned}
  \dot{p}_4-\dot{p}_3 = & -(p_4-p_3)-(p_4-p_2)+u + (p_3-p_4)+(p_3-p_2)+ (p_3-p_1)\\
             = & -3 (p_4-p_3) + (p_3-p_1) + u
\end{aligned}
\end{equation*}
Using the same technique as above, we obtain
\begin{equation*}
  \frac{1}{2} \frac{d}{dt}\delta_{34}^2  \leq  - 2 R^2 + |u| R
\end{equation*}
Thus, for this particular, incomplete, topology,  $|u| \leq 2 R$ ensures the preservation of link $(4,3)$.

\subsubsection{ General incomplete topology - Scaled case}\mbox{}

If each agent applies the movement equation with scale influence, then we have
\begin{itemize}
  \item for followers
  \begin{equation}\dlabel{ML-FolDynS}
      \dot{p}_i  =  -\frac{1}{n_i}\sum_{k \in \mathbb{N}_i^f} (p_i-p_k) -\frac{1}{n_i} \sum_{k \in \mathbb{N}_i^l} (p_i-p_k); \quad i \in \mathbb{N}^f
\end{equation}
  \item for leaders
  \begin{equation}\dlabel{ML-LeadDynS}
      \dot{p}_i=-\frac{1}{n_i}\sum_{k \in \mathbb{N}_i^l} (p_i-p_k) -\frac{1}{n_i} \sum_{k \in \mathbb{N}_i^f} (p_i-p_k) + u; \quad i \in \mathbb{N}^l
\end{equation}
\end{itemize}

 \begin{enumerate}
   \item  $i$ and $j$ are \textbf{followers}\\
   Applying  (\dref{ML-FolDynS}) to $i$ and $j$ one can write:
   \begin{equation}\label{F2Fs}
      \dot{p}_i -  \dot{p}_j =  \frac{1}{n_i}\sum_{k \in \mathbb{N}_i^f} (p_k-p_i) +\frac{1}{n_i} \sum_{k \in \mathbb{N}_i^l} (p_k-p_i)+\frac{1}{n_j}\sum_{k \in \mathbb{N}_i^f} (p_j-p_k) +\frac{1}{n_j} \sum_{k \in \mathbb{N}_i^l} (p_j-p_k)
   \end{equation}

   \begin{eqnarray*}
      \frac{d}{dt}\delta_{ij}^2 & = & 2(p_i-p_j)^T ( \dot{p}_i - \dot{p}_j)\\
      & = & 2 \left[ \frac{1}{n_i}\sum_{k \in \mathbb{N}_i^f} (p_i-p_j)^T (p_k-p_i) +\frac{1}{n_i} \sum_{k \in \mathbb{N}_i^l} (p_i-p_j)^T (p_k-p_i) \right ]\\
      & + & 2 \left[ \frac{1}{n_j}\sum_{k \in \mathbb{N}_j^f} (p_i-p_j)^T (p_j-p_k) +\frac{1}{n_j} \sum_{k \in \mathbb{N}_i^l}(p_i-p_j)^T (p_j-p_k) \right ]
   \end{eqnarray*}
   Using now
   \begin{itemize}
     \item $V_1^T V_2 \leq |V_1| |V_2|$
     \item $\delta_{ij} = R$
     \item $\delta_{ik} = |p_i-p_k| \leq R; \quad \forall k \in \mathbb{N}_i^f \quad \text{and   } k \in \mathbb{N}_i^f $ and similarly for $j$
   \end{itemize}
we obtain
\begin{equation}\label{dF2Fs}
   \frac{d}{dt}\delta_{ij}^2 \leq 2 \left[ \frac{1}{n_i} \left ( n_{if} R^2 + n_{il} R^2 \right ) + \frac{1}{n_j} \left ( n_{jf} R^2 + n_{jl} R^2 \right ) \right]
\end{equation}

Since $n_i = n_{if} + n_{il}$, and similarly for $j$, eq. (\dref{dF2Fs}) becomes
\begin{equation}\label{dF2Fs2}
   \frac{d}{dt}\delta_{ij}^2 \leq 4 R^2
\end{equation}
Note that equation \dref{dF2Fs2} does not ensure that $\frac{d}{dt}\delta_{ij}^2 \leq 0$ when the distance between two followers approaches the visibility range $R$.  As such it is not useful, since it does not ensure that this distance does not increase beyond $R$.

   \item $i$ and $j$ are \textbf{leaders}\\

Since $u$ is common to all leaders we obtain the same bound on leader to leader link as on follower to follower link, (\dref{dF2Fs2})

   \item $i$ is a \textbf{follower and} $j$ is a \textbf{leader}\\
Following the same procedure as above, we obtain
\begin{equation}\label{dL2Fs2}
   \frac{d}{dt}\delta_{ij}^2 \leq 4 R^2 + 2|u|R
\end{equation}

Thus, without any assumptions on the topology of the graph, the property of never losing friends when applying the protocol with scaled influence, cannot be proven.
 \end{enumerate}

\begin{figure}
\begin{center}
\includegraphics[scale=0.6]{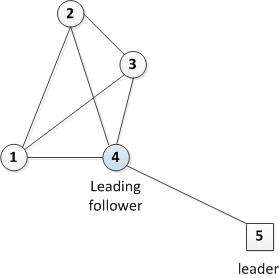}
\caption{Illustration of incomplete graph-case 1}\label{fig-case1}
\end{center}
\end{figure}

\paragraph{Specific cases of topology - Uniform vs Scaled influence}\mbox{}
Although it seems from the above that scaled influence is weaker than uniform influence in never losing neighbors, we will show here that there are specific cases where visibility link preservation with uniform influence can be proven only under the assumption of certain initial configurations, i.e. under "conditional topology" conditions, while scaled influence relaxes these conditions.
We consider  the case of a single leader with a single link to a complete sub-graph of followers, as illustrated in Fig. \dref{fig-case1}.
 and show that
\begin{itemize}
  \item if the uniform protocol is applied, then some very strict constraints on the initial states are required in order to ensure the property of never lose neighbors for $n_f > 2$
  \item if the scaled protocol  is applied, then  for $ |u|\leq \frac{n}{n-1} R$ all initial visibility links are preserved for any $n>2 \quad \text{or} \quad n_f>1$.

\end{itemize}

\subparagraph{Uniform influence with a-single-leader-to-a-single-follower connection}
The topology considered here belongs to the class of incomplete graphs with the followers forming a complete subgraph and the leaders forming a complete subgraph, discussed in Appendix \dref{incomplete}.
Assuming the leader to be agent $n$ and its adjacent follower to be agent $n-1$ we have:
\begin{itemize}
  \item $n_{il}=0$  for $i=1,...,n-2$
  \item $n_{il}=1$ for  $i = n-1$
  \item $n_{if}=1$  for $i=n$
  \item $n_{if}=n-2$ for  $i=1,...,n-1$
  \item $n_{(ij)l}=0$, for all $i,j$.
\end{itemize}

Thus, conditions (\dref{F2F-connect})-(\dref{L2F-preserve}) in Appendix \dref{incomplete} become:
\begin{itemize}
  \item The leader to leader condition (\dref{L2L-connect}) is not applicable
  \item The condition for follower to follower connection preservation (\dref{F2F-connect}): since $n_{il} + n_{jl} \leq 1$, we obtain : $n_f\geq 1$, obvious.
  \item The conditions for the leader to follower connection preservation (\dref{L2F-preserve}) yields: $n<4$  (namely $n \leq 3$ or equivalently $n_f\leq 2$) and thus (\dref{u-connect}) yields $|u| \leq R (4-n)$
\end{itemize}
Therefore, a single leader with a single connection to followers cannot be proven to drive the followers without losing the connection  \textbf{unless the number of followers $n_f \leq 2$} and the exogenous control satisfies $|u| \leq R (4-n)$.

\subparagraph{Conditional initial links preservation without limiting the number of followers}

By conditional initial links preservation we mean that the initial links can be proven to be maintained only when the \textbf{initial states are limited to certain configurations}.  We have shown above that for the single leader with single leader-follower connection, the initial links can be proven to be preserved only when the number of followers is limited to two.
Here we show that with certain initial configurations, the restriction on the number of followers is removed.  In particular, we show that there exists an exogenous control $u$ such that for certain initial configurations, the link between the leader and the leading-follower is preserved for any number of followers.

In the following two lemmas we look at a graph where agent $n$ is \underline{a single leader  with a single link} to a follower labelled $n-1$, which will be called "leading follower".  We assume that the followers subgraph is initially complete and denote the visibility range by $R$.
\LB{lemma1}
Suppose that the following initial condition holds:
\begin{equation*}
\delta_{n-1,i}(0)  <  \frac{R}{n-1}; \text{    } i=1,...,n-2
\end{equation*}
Then for all times $t$ we have that:
\begin{equation*}
\delta_{ij}(t)  <  \frac{2R}{n-1}; \text{    } i,j=1,...,n-2
\end{equation*}
\LE
\PB
By the triangle inequality the following holds:
\begin{equation*}
\delta_{ij}(0)  <  \frac{2R}{n-1}; \text{    } i,j=1,...,n-2
\end{equation*}
The Lemma follows from the fact that \emph{the distance between non-leading followers is monotonically decreasing}.  This is seen from the fact that, given that the followers subgraph is complete, we have for $i,j=1,...n-2$:
\begin{eqnarray*}
  \dot{p}_i &=& -(n-1)p_i+\sum_{k=1}^{n-1}p_k \\
  \dot{p}_j &=& -(n-1)p_j+\sum_{k=1}^{n-1}p_k
\end{eqnarray*}
and thus
\begin{eqnarray*}
  \frac{d}{dt} \delta_{ij}^2 (t)&=& 2(p_i(t)-p_j(t))^T(\dot{p}_i(t)-\dot{p}_j(t))\\
   &=& -2(n-1) \delta_{ij}^2(t)
\end{eqnarray*}
with the solution
\begin{equation}\dlabel{Conditional-F2F}
  \delta_{ij}^2 (t)=e^{-2(n-1)t}\delta_{ij}^2(0)
\end{equation}
\PE

\LB{lemma2}
Suppose that $|u|\leq \frac{n}{n-1} R$ and that
the following initial conditions hold:
\begin{eqnarray*}
  \delta_{n,n-1}(0) & < & R \\
  \delta_{n-1,i}(0) & < & \frac{R}{n-1}; \text{    } i=1,...,n-2
\end{eqnarray*}
Then for all times $t$ hold:
\hB
\titem{a} $\delta_{n,n-1}(t)  \leq R$
\titem{b} $\displaystyle \delta_{n-1,i}(t) \leq \frac{R}{n-1}; \text{    } i=1,...,n-2$

\hE
\LE
\PB
We shall prove the Lemma by contradiction.  Suppose \tref{a}, \tref{b} do not hold and let $t_1$ be \textbf{the first time when \tref{a} and/or \tref{b} is contradicted by one or more links}, namely that
 \begin{eqnarray*}
   \delta_{n,n-1}(t_1^+) &>& R \\
  \text{and/or  } \delta_{n-1,i}(t_1^+) & > & \frac{R}{n-1}
 \end{eqnarray*}
  Note that until time $t_1$ both \tref{a} and \tref{b} hold for all links and since all $\delta$'s are continuous functions, at time $t_1$ holds $\delta_{n-1,i}(t_1) \leq \frac{R}{n-1}$ and $\delta_{n,n-1}(t_1)  \leq  R; \text{    } i=1,...,n-2$.  Consider \textbf{any one} of the links that contradicts \tref{a} or \tref{b} at time $t_1$.  If link  $(n,n-1)$ contradicts \tref{a}, then
  \begin{eqnarray}
    \delta_{n-1,n}(t_1) &=&R\label{n1n-t1} \\
    \delta_{n-1,i}(t_1) & \leq & \frac{R}{n-1}; \quad i= (1, ...., n-2)\label{n1i-t1} \\
    \frac{d }{dt}\delta_{n-1,n}^2(t_1)& > & 0\label{not-a-t1}
  \end{eqnarray}

Starting from
\begin{eqnarray*}
  \dot{p}_n &=& p_{n-1}-p_n+u \\
 \dot{p}_{n-1}  &=& \sum_{i=1}^{n-2} (p_i-p_{n-1})+p_n-p_{n-1}
\end{eqnarray*}
we obtain
\begin{eqnarray*}
  \frac{d }{dt}\delta_{n-1,n}^2(t_1)& =& 2(p_{n-1}(t_1)-p_n(t_1))^T \left [ \sum_{i=1}^{n-2} \left ( p_i(t_1)-p_{n-1}(t_1) \right ) + 2\left ( p_n(t_1)-p_{n-1}(t_1) \right )-u \right ]\\
  & = & -4 \delta_{n,n-1}^2(t_1)+2 \sum_{i=1}^{n-2} \left( (p_n(t_1)-p_{n-1}(t_1))^T (p_{n-1}(t_1)-p_i(t_1)) \right ) +2 (p_n(t_1)-p_{n-1}(t_1))^T u \\
  & \leq & -4 R^2+2 (n-2)\frac{R^2}{n-1} + 2 R |u|
\end{eqnarray*}
where we used (\dref{n1n-t1}), (\dref{n1i-t1}) and the property of inner products $V_1^T V_2 \leq |V_1||V_2|$.
Since $ |u|\leq \frac{n}{n-1} R$ , we have

\begin{equation*}
  \frac{d }{dt}\delta_{n-1,n}^2(t_1) \leq -2 R^2 < 0
\end{equation*}
contradicting (\dref{not-a-t1}), i.e. the assumption that \tref{a} does not hold.

Now suppose that the considered link is $(n-1,i)$ for some follower $i \in 1,...,n-2$.  At  time $t_1$ holds
\begin{eqnarray}
  \delta_{n,n-1}(t_1) & \leq & R\label{b1-t1} \\
  \delta_{n-1,i}(t_1)& = & \frac{R}{n-1}; \quad i \in 1,...,n-2\label{b2-t1}  \\
  \delta_{n-1,j}(t_1) & \leq & \frac{R}{n-1}; \quad j= 1,...,n-2 ; \quad j \neq i\label{b3-t1}
\end{eqnarray}
If \tref{b} is contradicted by link $(n-1,i)$ for the first time at $t_1$ then $\displaystyle \frac{d}{dt}\delta_{n-1,i}^2(t_1) > 0 $ will hold.
\begin{eqnarray*}
  \dot{p}_{n-1}  &=& \sum_{j=1}^{n-2} (p_j-p_{n-1})+p_n-p_{n-1}\\
  & = & -(n-1)p_{n-1}+\sum_{j=1}^{n-1} p_j-p_{n-1} +p_n\\
  \dot{p}_i  & = & \sum_{j=1}^{n-1} (p_j-p_i)\\
    & = & - (n-1) p_i + \sum_{j=1}^{n-1} p_j
\end{eqnarray*}


 \begin{eqnarray*}
   \frac{d }{dt}\delta_{n-1,i}^2(t) &= & 2(p_{n-1}(t)-p_i(t))^T [-2((n-1)(p_{n-1}(t)-p_i(t))+(p_n(t)-p_{n-1}(t))\\
    & \leq & -2(n-1) \delta_{n-1,i}^2+2\delta_{n-1,i} \delta_{n,n-1}(t)
 \end{eqnarray*}
 where we used again the inequality for inner products.
 Considering now  the above at time $t_1$ and using(\dref{b1-t1}), (\dref{b2-t1}),  we obtain
 \begin{eqnarray*}
   \frac{d }{dt}\delta_{n-1,i}^2(t_1) & \leq & 2 \frac{R^2}{n-1}-2 \frac{(n-1)R^2}{(n-1)^2}\\
   & \leq & 0
 \end{eqnarray*}
contradicting the assumption that \tref{b} does not hold at time $t_1$ for link $(n-1,i)$.

\PE

From the previous two Lemmas it follows that:

\TB{cond-preserve}
Let agent $n$ be a single leader  with a single link to a follower labelled $n-1$. Assume that followers subgraph is initially complete and denote the visibility range by $R$.
Suppose that $|u|\leq \frac{n}{n-1} R$ and that the following initial conditions hold:
\begin{eqnarray*}
  \delta_{n,n-1}(0) & < & R\\
  \delta_{n-1,i}(0) & < & \frac{R}{n-1}; \text{    } i=1,...,n-2
\end{eqnarray*}

Then neighbors are never lost, i.e. all initial links are preserved.

\TE

\subparagraph{Scaled influence with a-single-leader-to-a-single-follower connection}

We assume as before that the followers form a complete graph.  The agents are labeled s.t. agent $n$ is the leader and $n-1$ is the leading follower. There are no constraints on the initial conditions, i.e. $\delta_{ij}(0) \leq R; \quad \forall i \sim j$. Recall that all agents apply the scaled protocol
\begin{equation*}
  \dot{p}_i = \frac{1}{n_i}\sum_{j \in N_i} (p_j-p_i)+b_i u
\end{equation*}
where
\begin{itemize}
  \item $p_i = (x_i \quad y_i)^T$
  \item $N_i$ is the neighborhood of $i$ and $n_i=|N_i|$
  \item $b_i$ is 1 if $i=n$, i.e. $i$ is the leader, and 0 otherwise
\end{itemize}

\TB{T-ScaledSingle}
Let $n$ agents with scaled influence and with visibility range $R$  have a-single-leader-to-a-single-follower connection and complete followers subgraph. If we label the leader by $n$ and the leading follower by $n-1$, then
\hB
\titem{a} \underline{for $ i,j=1,...n-2$},	$\delta_{ij}(t)$ is monotonically decreasing, thus if  $\delta_{ij}(0) \leq 0 $ then $\delta_{ij}(t) <0$ for all $t$, independently of the external control, $u$
\titem{b}	if  $ |u|\leq \frac{n}{n-1} R$ and
\begin{eqnarray*}
   \delta_{n,n-1}(0) & < & R \\
   \delta_{n-1,i}(0) & <  & R; \quad i=1,....n-2
  \end{eqnarray*}
  then
  \begin{eqnarray*}
  \delta_{n,n-1}(t) & \leq & R \\
  \delta_{n-1,i}(t) & \leq & R; \quad i=1,....n-2
  \end{eqnarray*}
  for all $t$
\hE

\TE
\PB
\underline{Property \tref{a}} -    As before, we consider $\displaystyle \frac{d}{dt}\delta_{ij}^2(t); \quad i,j=1,...,n-2$  and use
  \begin{eqnarray*}
    \frac{d}{dt}\delta_{ij}^2(t) &= & 2(p_i(t)-p_j(t))^T(\dot{p}_i(t)-\dot{p}_j(t))  \\
     \dot{p}_i(t) & = & -\frac{n-1}{n-2}p_i(t)+\frac{1}{n-2}\sum_{k=1}^{n-1}p_k(t) \\
     \dot{p}_j (t)& = & -\frac{n-1}{n-2}p_j(t)+\frac{1}{n-2}\sum_{k=1}^{n-1}p_k(t)
  \end{eqnarray*}
 to obtain
 \begin{equation*}
    \frac{d}{dt}\delta_{ij}^2(t)=-2 \frac{n-1}{n-2}\delta_{ij}^2(t)
 \end{equation*}
 with the solution
 \begin{equation}\dlabel{Scaled-F2F}
  \delta_{ij}^2 (t)=e^{- 2  \frac{n-1}{n-2} t}\delta_{ij}^2(0)
\end{equation}
which is monotonically decreasing\\
\underline{Property \tref{b}} - We shall prove this property again by contradiction. Suppose that  $t_1$ is the first time that this property is contradicted by one or more links, namely an external control $ |u|\leq \frac{n}{n-1} R$  is applied and
\begin{eqnarray*}
  \delta_{n,n-1}(t_1^+) & > & R \\
\text{and/or  }  \delta_{n-1,i}(t_1^+) & > & R; \quad i=1,....n-2
  \end{eqnarray*}
Since $t_1$ is the first time that the above holds and all links sizes are continuous functions, for all $t \leq t_1$ property \tref{b} holds, thus
\begin{eqnarray*}
  \delta_{n,n-1}(t_1) & \leq & R \\
  \delta_{n-1,i}(t_1) & \leq & R; \quad i=1,....n-2
  \end{eqnarray*}
Consider any one of the links that contradicts \tref{b} at time $t_1$.
\begin{enumerate}
  \item Assume that the link that contradicts \tref{b} at time $t_1$ is $(n,n-1)$, then let
  \begin{eqnarray}
  \delta_{n,n-1}(t_1) & = & R\label{dist-t1-nn1}\\
  \delta_{n-1,i}(t_1) & \leq & R; \quad i=1,....n-2\label{dist-t1-n1i}
  \end{eqnarray}
  and show that $ \frac{d }{dt}\delta_{n,n-1}^2(t_1) > 0$ \textbf{does not hold}.
  \begin{eqnarray*}
    \dot{p}_n &=& p_{n-1}-p_n+u \\
    \dot{p}_{n-1} &=& \frac{1}{n-1} \left [ \sum_{j=1}^{n-2}(p_j-p_{n-1})+(p_n-p_{n-1}) \right ]
  \end{eqnarray*}
  \begin{equation*}
   \dot{p}_n- \dot{p}_{n-1}=-2 (p_n-p_{n-1})+\frac{1}{n-1} \sum_{j=1}^{n-2}(p_{n-1}-p_j)+u
  \end{equation*}

  Multiplying from left by $2(p_n-p_{n-1})^T$ and using the inequality for inner products, we obtain
\begin{equation*}
  \frac{d }{dt}\delta_{n,n-1}^2(t) \leq -4\delta_{n,n-1}^2(t)+  \frac{2}{n-1}\sum_{j=1}^{n-2} \delta_{n,n-1}(t) \delta_{n-1,j}(t)+2  \delta_{n,n-1}(t)|u|
\end{equation*}
Since at $t_1$ eqs. (\dref{dist-t1-nn1}), (\dref{dist-t1-n1i}) hold, we have
\begin{equation*}
  \frac{d }{dt}\delta_{n,n-1}^2(t_1) \leq -4R^2+2\frac{n-2}{n-1}R^2+2 R |u|
\end{equation*}
Thus, if $|u| \leq \frac{n}{n-1}R$, then $ \frac{d }{dt}\delta_{n,n-1}^2(t_1) \leq 0$, contradicting the assumption that statement \tref{b} of the Theorem does not hold for link $(n,n-1)$.
  \item Consider now a link $(n-1,i)$, for any $i=1,...,n-2$ and show that it cannot be the one that first contradicts \tref{b} at time $t_1$. As for \tref{a}, we have at $t_1$
      \begin{eqnarray*}
  \delta_{n,n-1}(t_1) & \leq & R \\
  \delta_{n-1,i}(t_1) & \leq & R; \quad i=1,....n-2
  \end{eqnarray*}
  For any link $j; \quad j \in 1,....,n-2$ assumed to contradict statement \tref{b} of the Theorem, we have to show that when $ \delta_{n-1,j}(t_1) = R$, while
   \begin{eqnarray}
  \delta_{n,n-1}(t_1) & \leq & R\label{scaled1-t1} \\
  \delta_{n-1,i}(t_1) & \leq & R; \quad i \neq j\label{scaled2-t1}
  \end{eqnarray}
  if $|u| \leq \frac{n}{n-1}R$, then $ \frac{d }{dt}\delta_{n-1,j}^2(t_1) > 0$ does not hold, contradicting the assumption that statement \tref{b} of the Theorem does not hold for link $(n-1,j)$.
      We have
      \begin{eqnarray*}
        \dot{p}_{n-1} &=& \frac{1}{n-1} \left [ \sum_{j=1}^{n-2}(p_j-p_{n-1}) +(p_n-p_{n-1}) \right ]\\
         \dot{p}_{j} &=& \frac{1}{n-2} \left [ \sum_{k=1}^{n-2}(p_k-p_j) +(p_{n-1}-p_j) \right ]
      \end{eqnarray*}
      \begin{eqnarray*}
        \frac{d}{dt} \delta_{n-1,j}^2(t) &=& 2(p_{n-1}(t)-p_j(t))^T (\dot{p}_{n-1}(t)-\dot{p}_j(t)) \\
         &=& \frac{2}{n-1} \left [ \sum_{k=1}^{n-2}(p_{n-1}(t)-p_j(t))^T(p_k(t)-p_{n-1}(t)) + (p_{n-1}(t)-p_j(t))^T(p_n(t)-p_{n-1}(t)) \right ]\\
          & & - \frac{2}{n-2}\left [ \sum_{k=1}^{n-2}(p_{n-1}(t)-p_j(t))^T(p_k(t)-p_j(t)) + \delta_{n-1,k}^2(t)\right ]
      \end{eqnarray*}

Using now in the above equation, at $t=t_1$, $ \delta_{n-1,j}(t_1) = R$, (\dref{scaled1-t1}), (\dref{scaled2-t1}) and the inequality for inner products $V_1^T V_2 \leq |V_1||V_2|$, we obtain
   \begin{eqnarray*}
     \frac{d}{dt} \delta_{n-1,i}^2(t_1) & \leq & -\frac{2}{n-1}R^2 \\
      & < & 0
   \end{eqnarray*}
again contradicting the assumption that statement \tref{b} of the Theorem does not hold for this link.

\end{enumerate}

\PE

\subsubsection{Some simulation results with incomplete initial interaction graph}\label{IncompleteEx}
\begin{figure}
\begin{center}
\includegraphics[scale=0.5]{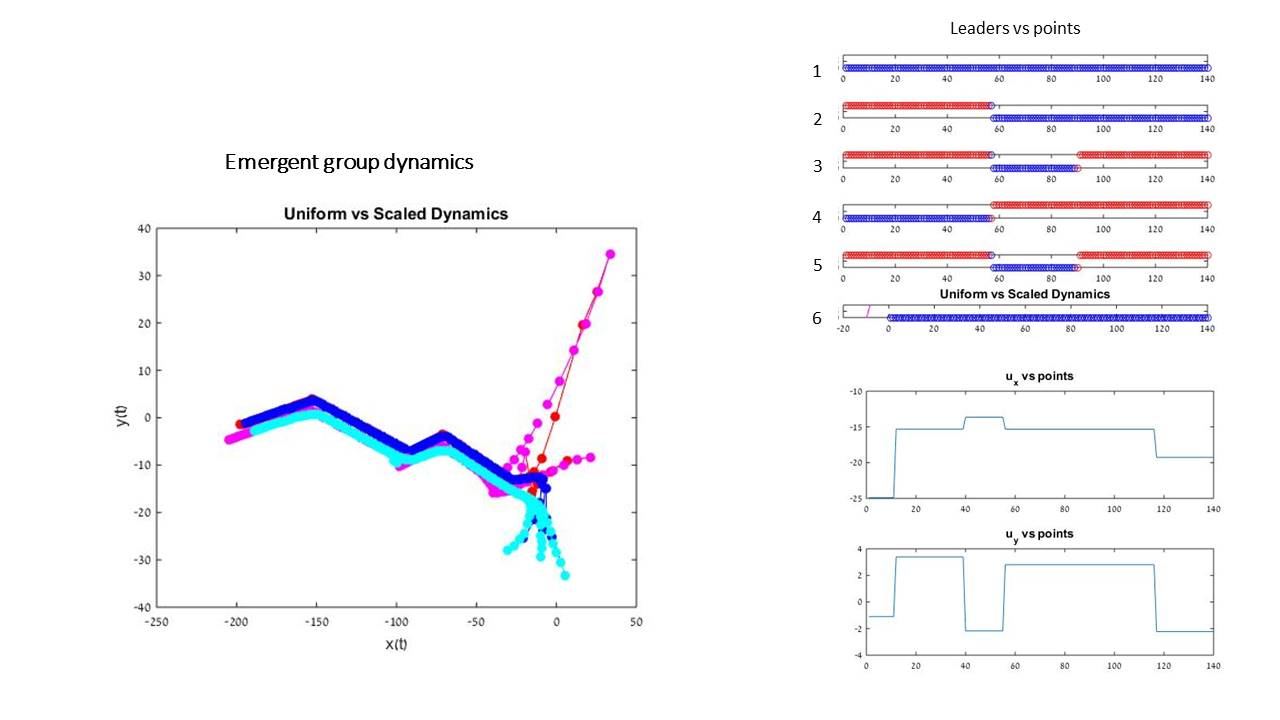}
\caption{Dynamics with Incomplete initial graph - Ex1}\label{fig-IncompleteEx1}
\end{center}
\end{figure}

In this section two examples are shown where the agents initial interaction topology is incomplete. Although we could not find analytic limits on $|u|$ such that  the property of never lose neighbors is ensured we ran both cases, and many others, with $|u|< R$ and in both cases the agents converged to a complete graph which was afterwards preserved.
\paragraph{Incomplete initial interaction graph - Ex1}
\begin{itemize}
  \item n=6
  \item initial number of links = 10
  \item $u$ and leaders as shown in Fig. \dref{fig-IncompleteEx1}
\end{itemize}

\begin{figure}
\begin{center}
\includegraphics[scale=0.7]{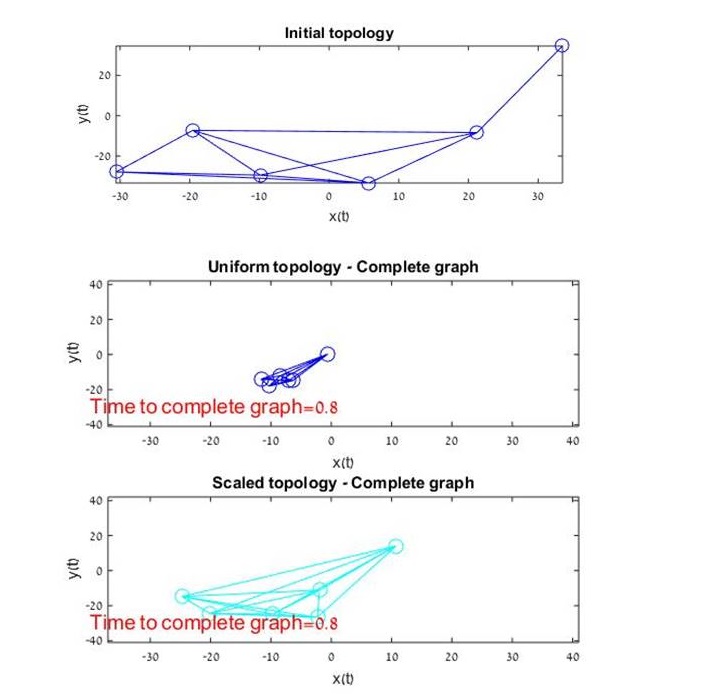}
\caption{Convergence to complete graph - Ex1}\label{fig-GraphsIncompleteEx1}
\end{center}
\end{figure}

\begin{figure}
\begin{center}
\includegraphics[scale=0.4]{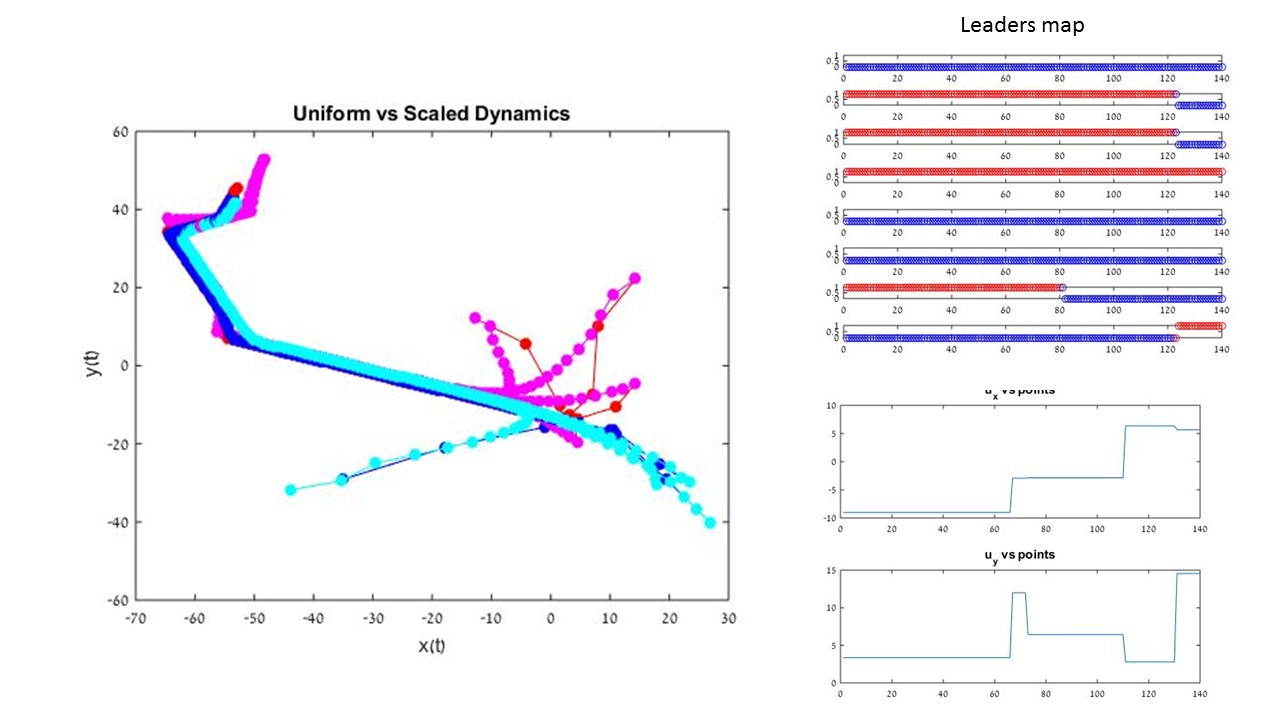}
\caption{Dynamics with initially incomplete graph - Ex2}\label{fig-IncompleteEx2}
\end{center}
\end{figure}

\paragraph{Incomplete initial interaction graph - Ex2}
\begin{itemize}
  \item n=8
  \item initial number of links = 12
  \item $u$ and leaders randomly selected, as shown in Fig. \dref{fig-IncompleteEx2}
\end{itemize}

\begin{figure}
\begin{center}
\includegraphics[scale=0.4]{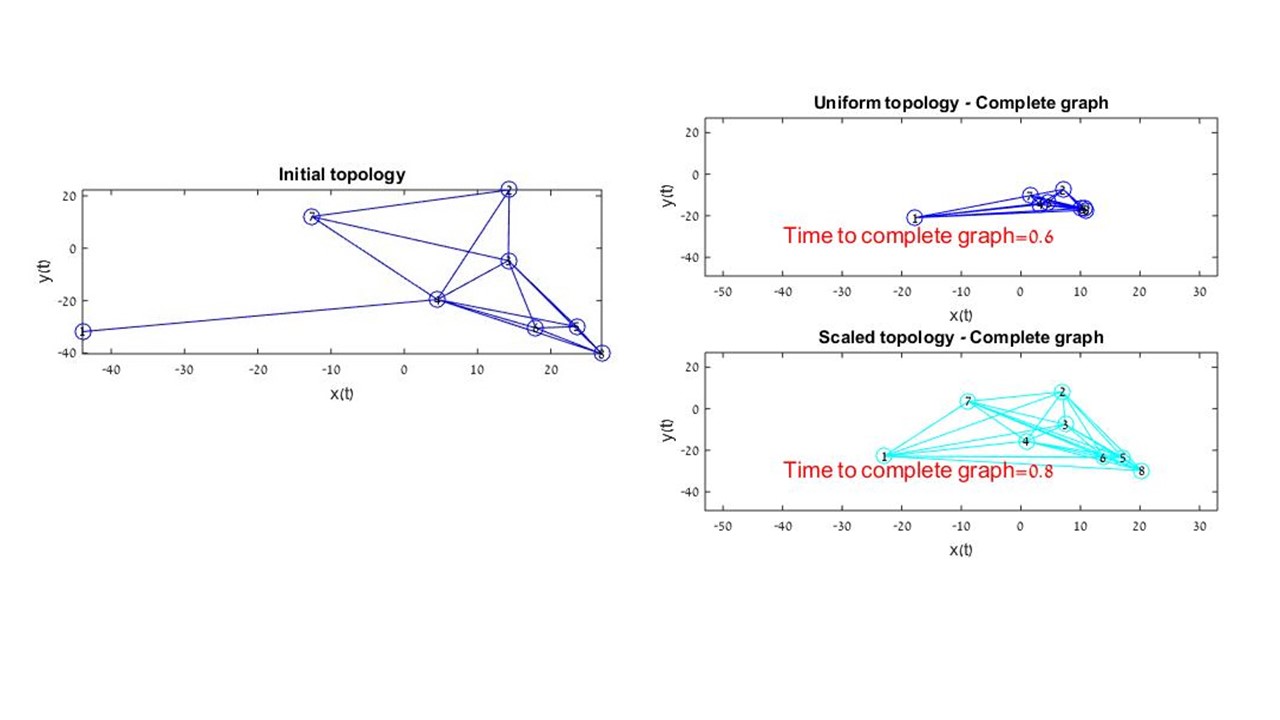}
\caption{Convergence to complete graph - Ex2}\label{fig-GraphsIncompleteEx2}
\end{center}
\end{figure}

\newpage

\msec{Summary and directions for future research}{future}

In this report we introduced a model for controlling swarms of identical, simple, oblivious, myopic agents by  broadcast velocity control that is received by a random set of agents in the swarm.  The agents detecting the broadcast control are the ad-hoc leaders of the swarm, while they detect the exogenous control.  All the agents, modeled as single integrators, apply a local linear gathering control, based on the weighted relative position to all neighbors. The weights are the neighbors' influence on the agent.  The leaders superimpose the received exogenous control, a desired velocity $u$. We considered two models of  neighbors influence, uniform and scaled by the size of agent's neighborhood.  We have shown that if the the system is piecewise constant, where in each time interval the  system evolves as a time-independent dynamic linear system with a connected visibility graph, then in each such interval, $[t_k, t_{k+1})$,  the swarm tends to asymptotically align on a line in the direction of $u(t_k)$, anchored at the zero-input gathering point, $\alpha(t_k)$ and moves with a collective velocity that is a fraction of the desired velocity. We denote this fraction by $\beta(t_k)$. If the visibility graph in the interval is complete, then $\alpha(t_k)$ for both influence models is the same, the average of all agents' positions at the beginning of the interval, and $\displaystyle \beta(t_k)=\frac{n_l(t_k)}{n} $.  However, if the visibility graph in the interval is incomplete then $\alpha(t_k)$  and $\beta(t_k)$ are not the same for the two models. Moreover, in the scaled case they are a function of the topology of the graph, the in-degree of its nodes and of the selected leaders. Since we assumed that in each interval the visibility graph is connected we need conditions to ensure that a connected graph remains connected. We showed that if the graph is complete then restrictions on $|u|$, pending the influence model, will ensure that it remains complete. However, when the graph is incomplete, conditions for never losing neighbors are tightly related to the graph topology and therefore not useful in practice. Never losing neighbors might be too stringent a requirement. We note that although conditions, independent of specific topologies, for never losing friends in an incomplete graph were not found, in practice all simulations that we ran showed convergence to complete graphs which were afterwards preserved if $|u|$ was within bounds.\\
In future research, we intend to extend the dynamic model to  double integrators, i.e. acceleration controlled agents. Also, we are currently considering the same paradigm of stochastic broadcast control in conjunction with non-linear gathering processes, as for example \dcite{Bellaiche}, and connectedness preserving gathering processes, as for example \dcite{DJ2010}.

\newpage
\appendix
\addcontentsline{toc}{section}{Appendices}

\msec{Algebraic representation of Graphs }{preliminary}

Graphs are broadly adopted in the multi-agent literature to encode interactions in networked systems. In this appendix some useful facts from algebraic graph theory are presented. Graphs and algebraic graph theory have proven to be powerful tools when working with agent networks. Given a multi-agent system, the network can be represented by a directed or an undirected graph  $G=(V,E)$, where  $V$ is a finite set of vertices,  representing agents, and $E$ is the set of edges, $E \in [V \times V]$, representing inter-agent information exchange links.\footnote{
Vertices are also referred to as nodes and the two terms will be used interchangeably}
 A simple graph contains no self-loops, namely there is no edge from a node to itself. If the graph is undirected then the edge set $E$ contains unordered pairs of vertices. In directed graphs (digraphs) the edges are ordered pairs of vertices. Graphs also admit representations in terms of matrices. Examples of such matrices are:
\begin{itemize}
  \item Adjacency and degree
  \item Incidence matrix
  \item Laplacian
\end{itemize}
These will be discussed in the next sections for undirected graphs as well as for  digraphs.

\msubsection{Undirected graphs}{Graphs}

We denote an unordered graph by $G^U$ and label a node $v_i$ by $i$. Then, recalling that in unordered graphs edges contain unordered pairs of vertices, when an edge exists between vertices $i$ and $vj$, we refer to them as adjacent, and denote this relationship by $i \sim{\ } j$. In this case, edge $E_{ij}=(i,j)$ is called incident with vertices $i$ and $j$.
The neighborhood of the vertex $i$, denoted by $N_i$, is the set of all vertices that are adjacent to $i$. A path of length $m$ in $G^U$ is given by a sequence of distinct vertices such that for $k = 0,1,…, m - 1$, the vertices ${i+k}$ and ${i+k+1}$ are adjacent. In this case, $i$ and ${i+m}$ are referred to as the end vertices of the path.  We say that the graph is connected if for every pair of vertices in $V$  there is a path with those vertices as its end vertices. If this is not the case, the graph is called disconnected.
We refer to a connected graph as having one connected component. A disconnected graph has more than one component.
The number of neighbors of each vertex $i$ is its degree, denoted by ${{d}_{i}}$.  The degree matrix $\Delta $ of a graph $G^U$ is a diagonal matrix with elements $\Delta_{ii}={{d}_{i}}$, where $d_i$ is the degree  of vertex $i$.  Any simple graph can be represented by its adjacency matrix.  For an undirected, unweighed graph $G^U$, the adjacency matrix   $A^U$, is a symmetric matrix with 0,1 elements, such that

\begin{equation}\label{def-A}
A^U_{ij} =
\begin{cases}
1 & \text{if } v_i \sim{\ } v_j
\\
0 & \text{otherwise }
\\
\end{cases}
\end{equation}

Another matrix representation of a graph, is the Laplacian. The most straightforward definition of the graph Laplacian associated with an undirected graph $G^U$ is
\begin{equation}\label{def-L}
  L^U=\Delta -A^U
\end{equation}
Thus, the sum of each row and of each column of the Laplacian  $L^U$ is zero.\\
Note: $L^U, A^U$ are shortcuts for $L(G^U), A(G^U)$, the Laplacian and adjacency matrix respectively associated with the undirected, unweighted, graph $G^U$, to be used whenever the context is clear.

An alternate and useful definition of $L^U$ is by using the incidence matrix $\in {{\mathbb{R}}^{n\times m}}$, see Theorem  \dref{T-Laplacian}, where $n$ is the number of vertices and $m$ is the number of edges.  For an undirected graph we select an arbitrary orientation for all edges and define the incidence matrix $B$ as:

\begin{equation}\label{def-incidence}
B_{ik} =
\begin{cases}
+1 & \text{if } i \text{  is the head of edge } k
\\
-1 & \text{if } i \text{  is the tail of edge } k
\\
0  &  \text{otherwise}
\\
\end{cases}
\end{equation}
where  $i\in V$  ; $k\in E $.

Some basic properties of  $L^U$, the Laplacian associated with an undirected graph,  are presented in the next section.

\subsubsection{Properties of  the Laplacian associated with an undirected graph}

\TB{T-Laplacian}
The Laplacian $L^U$,  associated with an undirected graph, satisfies the property $L^U=B{{B}^{T}}$, regardless of the edge orientation selection.
\TE

\PB
We need to prove that
\begin{equation*}
{{\left[ B{{B}^{T}} \right]}_{ik}} =
\begin{cases}
d_i & \text{if } i=j
\\
-1 & \text{if } i \neq j \text{ and } i \sim j
\\
0  &  \text{otherwise}
\\
\end{cases}
\end{equation*}
where $i \in V , j \in V$.  We have

\begin{equation*}
  {{\left[ B{{B}^{T}} \right]}_{ii}}=\sum\limits_{j=1}^{m}{B_{ij}^{2}}\text{ = (number of entries }\ne \text{0 in row }i)={{d}_{i}}
\end{equation*}

\begin{equation*}
  {{\left[ B{{B}^{T}} \right]}_{ik}}=\sum\limits_{j=1}^{m}{B_{ij}^{{}}}B_{kj}^{{}}=
  \begin{cases}
-1 & \text{if } i \sim k
\\
0  & \text {otherwise}
\\
  \end{cases}
\end{equation*}

\TB{T-properties}

For an undirected graph,
\hB
\titem{a}	the associated Laplacian $L^U$ is real and symmetric
\titem{b}	$L^U$ is positive semi-definite
\titem{c}	the eigenvalues of $L^U$ are real and nonnegative
\titem{d}  there is always an orthonormal basis for $\mathbf{R}^n$ consisting of real eigenvectors of $L^U$.
\hE
\TE
\PB
The symmetry is obvious, from (\dref{def-A}) and (\dref{def-L}). The positive semi-definiteness is due to ${{x}^{T}}Lx={{x}^{T}}B{{B}^{T}}x=||{{B}^{T}}x|{{|}^{2}}\ge 0$.
Property \tref{c} follows from \tref{a},  \tref{b} and \tref{d} follows from \tref{a}. If the eigenvalues of $L^U$ are distinct then all we need to do is find an eigenvector for each eigenvalue and if
necessary normalize it by dividing by its length. If there are repeated roots, then
it will usually be necessary to apply the Gram{Schmidt process to the set of basic
eigenvectors obtained for each repeated eigenvalue. Moreover, due to \tref{a},  \tref{c} these eigenvectors can be selected to be real, since:
\begin{itemize}
  \item if $v$ is an eigenvector of a real symmetric matrix $A$, associated with the real eigenvalue $\lambda$ , then $Av=\lambda v$ with $v \neq 0$.
  \item if $v=a+ib$ then $ A(a+ib)=\lambda (a+ib)$  from which follows that  $Aa=\lambda a$ and $Aa=\lambda a$
  \item since $v \neq 0$  either $a \neq 0$ or $b \neq 0$, thus either $a$ or $b$ is a real eigenvector of $A$.
  \item If both $a, b \neq 0$ then one can choose the real eigenvector of $A$ corresponding to the real eigenvalue $\lambda$.

\end{itemize}

In view of Theorem \dref{T-properties}, the eigenvalues of $L^U$, denoted by ${\lambda^U_{1}},{\lambda^U_{2}},...,{\lambda^U_{n}},$  referred to as \emph{spectrum} of $L^U$ , can be ordered as:
\begin{equation}\dlabel{eq-eigen}
  0={\lambda^U_{1}}\le {\lambda^U_{2}}\le ....\le {\lambda^U_{n}}
\end{equation}
The vector of ones, denoted by $\mathbf{1}_n$ is an eigenvector associated with the zero eigenvalue $\lambda^U_1$.

\TB{T-Connected1}

\hB
\titem{b} The multiplicity of the zero eigenvalue equals the number of components of $G^U$.
\titem{a} A graph $G^U$ is connected  iff ${\lambda^U_{2}}>0$.
\titem{c} If the graph $G^U$ is connected, then
\begin{equation}\dlabel{eq-Xpositive}
z^T L^U z >0 \text{  for any  } z \not\in \mathbf{span}\{\mathbf{1}_n\}
\end{equation}
\hE
\TE

\PB
Let $c$ be the number of connected components of $G^U$ and let $N^{(1)}, \cdots N^{(c)}$ be the collection of nodes in each of those components.
Note that if for some $n$-dimensional vector $z$ holds $B^T z = 0$ and $z_i \ne 0$, then $z_j = z_i $ for any neighbor $j$ of $i$ and by extension $z_j = z_i$ for any node $j$ in the same connected component as $i$.  Therefore, the null space of $B^T$ is spanned by the $c$ linearly independent vectors $Z^{(1)},\cdots, Z^{(c)}$ defined by
\begin{equation}\dlabel{eq-ZN}
Z_i^{(k)} = \begin{cases} 1 & \text{  if  }  i \in N^{(k)} \\
0 & \text{  otherwise  } \\
\end{cases}
\text{ }\text{ }\text{ }\text{ }k = 1, \cdots , c \text{ } ;\text{ } i = 1,...,n
\end{equation}

Now $B^T z = 0$ implies $BB^T z = 0$ and from Theorem \dref{T-Laplacian} follows $L^U z=0$.  Viceversa, if $L^U z = 0$ then $BB^T = 0$, hence $0 = z^T B B^T z = \parallel B^T z \parallel^2$, thus $B^T z = 0 $.  Therefore , the null space of $L^U$ is identical to the null space of $B^T$, which from (\dref{eq-ZN}) has degree $c$.
Since the multiplicity of the zero eigenvalue of $L^U$ is the degree of its null space, \dref{T-Connected1b}) follows.

The graph $G^U$ is connected iff $c=1$, hence \dref{T-Connected1a}).

Since $L^U$ is positive semi-definite and  $\mathbf{span}\{\mathbf{1}_n\}$ is the null space of $L^U$ when $c=1$, follows \dref{T-Connected1c}).
\vspace{12pt}

\subsubsection{Algebraic connectivity}
The eigenvalue $\lambda^U_2$ of the Laplacian associated with a graph $G^U$ is referred to as the \emph{algebraic connectivity} of $G^U$ and denoted by $a(G^U)$ (see \dcite{Fiedler}). A brief summary of   the properties of $a(G^U)$ follows:
\begin{itemize}
 \item Since $L^U$ is positive semi-definite one can write (using Courant's theorem):
 \begin{equation}\label{eq-lambda2}
   a(G^U)=\lambda^U_2= \min_{x \in X} \frac{x^T L^U x}{x^T x}
 \end{equation}
 where $X$ is the set of vectors s.t. $x^T \mathbf{1}_n=0$

  \item $a(G^U)$ is non-decreasing for graphs with the same set of vertices, i.e. if ${{G^U}_{1}}=(V,{{E}_{1}})\text{ and }{{G^U}_{2}}=(V,{{E}_{2}})$ such that ${{E}_{1}}\subseteq {{E}_{2}}$ then  $a(G^U_1)\le a(G^U_2) $
   \item  let $G^U_1$ arise from $G^U$ by removing $k$ vertices and all adjacent edges. Then $a(G^U_1) \geq a(G^U) - k$
   \item $\displaystyle a(G^U) \leq   \frac{n}{(n-1)} \min_i  d_i \leq \frac{2|E|}{(n-1)} $ where $n$  is the number of vertices in $G^U$, $d_i$ is the degree of vertex $i$, i.e. number of neighbors of $i$, and $|E|$ is the number of edges
   \item If $G^U$ is a graph with $n$ vertices which is not complete then $a(G^U) \leq n-2$.
   \item If $G^U$ is a complete graph with $n$ vertices, $K_n$, then $a(K_n) = n$.

\end{itemize}

Proofs for the above properties and additional properties appear in \dcite{Fiedler}.

\subsubsection{Normalized Laplacian associated with a graph $G^U$}
The normalized Laplacian associated with a graph $G^U$, denoted by $\Gamma$, is closely related to the Laplacian $L^U$  defined in section \dref{Graphs}.
\begin{equation}\label{normalizedL}
  \Gamma= \Delta^{-1/2} L^U \Delta^{-1/2} = I- \Delta^{-1/2} A^U \Delta^{-1/2}
\end{equation}
where $\Delta$ is the degree matrix of $G^U$ and $A^U$ its adjacency matrix (\dref{def-A}).
Entry-wise we have
\begin{equation*}
    \Gamma_{ij}=
    \begin{cases}
1 & \text{if } i=j
\\
\frac{-1}{\sqrt{d_i d_j}} & \text{ if  } i \sim j
\\
0 & \text{otherwise }
\\
\end{cases}
  \end{equation*}
where $d_i$ is the degree of vertex $i$ in $G^U$.

Denoting the eigenvalues of $\Gamma$ by  $\lambda^\Gamma_i; \quad 1 \geq i \leq n$, ordered s.t. $\lambda^\Gamma_1 \leq \lambda^\Gamma_2 \leq .... \leq \lambda^\Gamma_n$, one has \textbf{for a connected graph} $G^U$ on $n$ vertices(ref. \dcite{Chen}, Theorem 1.1):
\begin{enumerate}
  \item $\lambda^\Gamma_1=0$, with corresponding eigenvector $ \Delta^{1/2} \mathbf{1}_n$
  \item $\displaystyle \sum_{i=1}^n \lambda^\Gamma_i = n$
  \item For $n \geq 2, \quad \lambda^\Gamma _2 \leq \frac{n}{n-1}$ and $\lambda^\Gamma _n \geq \frac{n}{n-1}$ with equality holding if and only if $G^U$ is complete
  \item For a graph which is not a complete graph, we have $\lambda^\Gamma_2 \leq 1 $.
  \item For all $i \leq  n$, we have $\lambda^\Gamma_i \leq 2$ with $\lambda^\Gamma_n = 2$ if and only if $G^U$ is a nontrivial bipartite graph, as shown by Chung in \dcite{Chung}.

\end{enumerate}

\paragraph{Relationship of eigenvalues of normalized Laplacian to eigenvalues of standard Laplacian}\mbox{}\\
The standard Laplacian associated with a graph $G^U$ is defined by (\dref{def-L}) while the normalized Laplacian associated with the same graph, $\Gamma$, is defined by (\dref{normalizedL}).
Butler shows in \dcite{ButlerPhD}, Theorem 4, that
\begin{equation}\label{Norm2StdEig}
  \frac{1}{d_{max}} \lambda^U_i \leq \lambda^\Gamma_i \leq \frac{1}{d_{min}} \lambda^U_i
\end{equation}
where $d_{max}$ is the maximum degree and $d_{min}$ is the minimum degree of a vertex in $G^U$, $ \lambda^U_i$ are the eigenvalues of $L^U$ and $ \lambda^\Gamma_i$ are the eigenvalues of $\Gamma$, s.t. $\lambda^U_1=\lambda^\Gamma_1=0$. Moreover,  $0 \leq \lambda^\Gamma_i \leq 2$ while $0 \leq \lambda^U_i \leq 2 d_{max}$.

\subsubsection{Eigenvalues of the standard and normalized Laplacian upon deleting an edge}\label{interlaced}
Let $G^U$ be a connected graph, and let $G^U_1 = G^U-e$, where $e$ is an edge of $G^U$, s.t. there are no isolated vertices in $G^U_1$.  Let's denote the eigenvalues of the standard Laplacian  associated with $G^U_1$, $L(G^U_1)$, by $\theta_i$ while retaining the notation $\lambda^U_i$ for the eigenvalues of the standard Laplacian associated with $G^U$, $L(G^U)$, ordered  s.t.
\begin{equation*}
  0=\lambda^U_1 < \lambda^U_2 \leq \lambda^U_3 \leq ....\leq \lambda_n
\end{equation*}
and
\begin{equation*}
  0=\theta_1<\theta_2 \leq \theta_3 \leq ....\leq \theta_n
\end{equation*}
Then the eigenvalues of $L(G^U_1)$ interlace the eigenvalues of $L(G^U)$ (Ref. \dcite{Chen}, Theorem 2.2):
\begin{equation}\label{inter-standard}
  \lambda^U_n \geq \theta_n \geq \lambda^U_{n-1} \geq .......\geq \lambda^U_2 \geq \theta_2 > \lambda^U_1=0
\end{equation}
Moreover, one has (ref. \dcite{Mer1998})
\begin{equation*}
  \sum_{i=1}^n \lambda^U_i = 2+\sum_{i=1}^n \theta_i
\end{equation*}
Let's consider now the eigenvalues of the \emph{normalized Laplacians} associated with $G^U$ and $G^U_1$,  $\Gamma(G^U)$ and $\Gamma(G^U_1)$ respectively. If we denote  by $\lambda^\Gamma_i$ the eigenvalues of $\Gamma(G^U)$  and  by $\phi_i$ the eigenvalues of $\Gamma(G^U_1)$, sorted s.t.
\begin{equation*}
  0=\lambda^\Gamma_1 < \lambda^\Gamma_2  \leq ....\leq \lambda^\Gamma_n \leq 2
\end{equation*}
\begin{equation*}
  0=\phi_1 < \phi_2  \leq ....\leq \phi_n \leq 2
\end{equation*}
 then (Ref. \dcite{Chen}, Theorem 2.3) the eigenvalues of $\Gamma(G^U)$ do \textbf{not} simply interlace the eigenvalues of $\Gamma(G^U_1)$, since $\displaystyle \sum_{i=1}^n \lambda^\Gamma_i = \sum_{i=1}^n \phi_i=n$. Instead the following relationship,  (\dref{inter-normalized}), holds:
\begin{equation}\label{inter-normalized}
  \lambda^\Gamma_{i-1} \leq \phi_i \leq  \lambda^\Gamma_{i+1}; \quad i=1,...,n
\end{equation}
where we set $\lambda^\Gamma_0=0$ and $\lambda^\Gamma_{n+1}=2$

\subsubsection{Example of eigenvalues of standard and normalized Laplacians }
This example comes to illustrate
\begin{enumerate}
  \item the relationship of eigenvalues of a normalized Laplacian to those of the standard Laplacian associated with the same graph
  \item the change in eigenvalues of the standard Laplacian vs the normalized Laplacian upon deleting an edge

\end{enumerate}

Consider three graphs $G_1, G_2, G_3$, s.t. $G_2=G_1-e, G_3=G_2-e$ as shown in Fig. \dref{fig-GraphsForRelEigen} and let $L(G_i); \quad i=1,2,3$ be the standard Laplacian and $\Gamma(G_i)$ the normalized Laplacian associated with $G_i$.
\begin{figure}
\begin{center}
\includegraphics[scale=0.4]{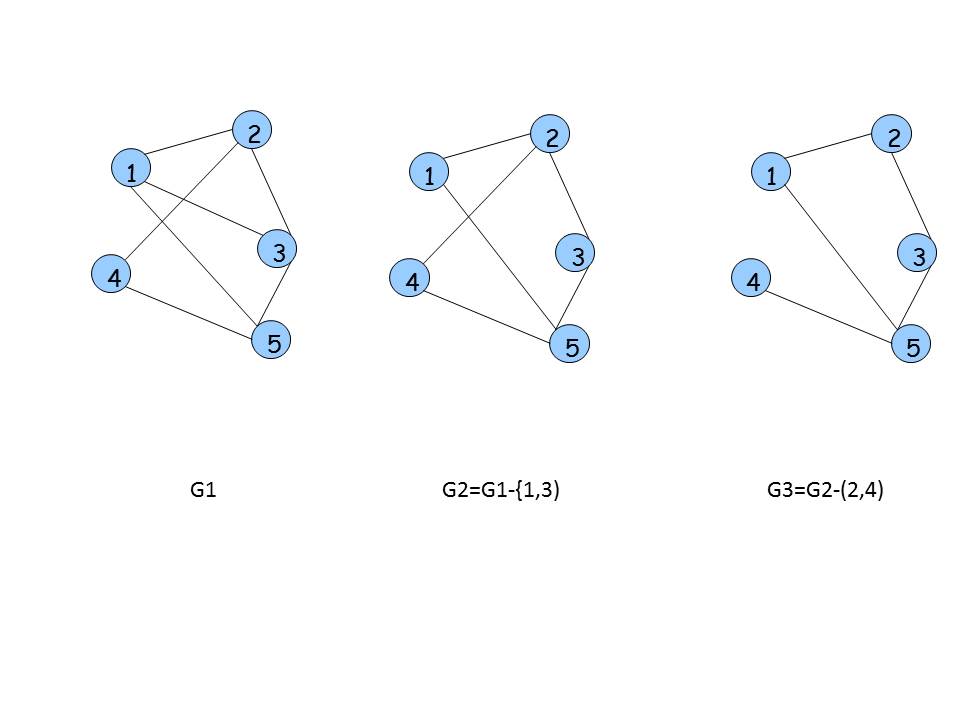}
\caption{Graphs used for the illustration of eigenvalues of standard and normalized Laplacians }\label{fig-GraphsForRelEigen}
\end{center}
\end{figure}

\begin{table}
\centering
\caption{Ex- Eigenvalues of standard and normalized Laplacians }
\label{EigRelation}
 \begin{tabular}{||c| c| c| c||}
 \hline
 &$G_1$ & $G_2$ & $G_3$  \\ [0.5ex]
 \hline\hline
 eig$(L(G_i)$ & 0; 2; 3; 4; 5 &0; 2; 2; 3; 5; &0; 0.83; 2; 2.7; 4.48 \\
 eig$(\Gamma(G_i)$ & 0; 0.862; 1; 1.33; 1.805 & 0; 1; 1; 1; 2 & 0; 0.59; 1; 1.41; 2\\
 $d_{\max}(G_i)$ &3 & 3 &  3 \\
 $d_{\min}(G_i)$ &2 & 2 & 1  \\
  eig$(L(G_i)/d_{\max}(G_i)$ &0; 0.67; 1; 1.33; 1.67 &0; 0.67; 0.67; 1; 1.67 &0; 0.28; 0.67; 0.9;  1.49 \\
  eig$(L(G_i)/d_{\min}(G_i)$ &0; 1; 1.5; 2; 2.5 &0; 1; 1; 1.5; 2.5 & 0; 0.83; 2; 2.7; 4.48\\[1ex]
 \hline
 \end{tabular}
\end{table}

By considering Table \dref{EigRelation} we observe that the presented examples illustrate (\dref{Norm2StdEig}), (\dref{inter-standard}), (\dref{inter-normalized}).

\msubsection{Directed graphs}{Digraph}
A directed graph (or digraph), denoted by $D= (V, E)$, is a graph whose edges are ordered pairs of vertices. For the ordered pair $(i, j) \in E$, when vertices $v_i, v_j$ are labelled $i, j$,  $i$ is said to be the tail  of the edge, while $j$ is its head.
Notions of adjacency, neighborhood and connectedness can be extended in the context of digraphs, e.g.:
\begin{itemize}
     \item The adjacency matrix for directed weighted graphs is defined as
  \begin{equation}\label{DirAdj}
    A^D_{ij}=
    \begin{cases}
\sigma_{ji} & \text{if } (j, i) \in E(D)
\\
0 & \text{otherwise }
\\
\end{cases}
  \end{equation}
  where $\sigma_{ji}$ is the strength of the influence of $j \text{  on  } i$
  \item The set of neighbors of node $i$, denoted by $N_i$, is defined as $N_i=\{j \in V : A^D_{ij}>0 \}$
   \item The in-degree $InDegree$ and out-degree $OutDegree$ of node $i$ are defined as
  \begin{eqnarray*}
    InDegree_i  &=& \sum_{j=1}^n A^D_{ij} \\
    OutDegree_i &=& \sum_{j=1}^n A^D_{ji}
  \end{eqnarray*}
  \begin{itemize}
    \item If the digraph is unweighted, i.e. $A^D_{ij}$ are binary, then $ InDegree_i=|N_i|$.
    \item If $InDegree_i=OutDegree_i$, i.e. the total weight of edges entering the node and leaving the same node are equal, then node $i$ is called balanced
    \item If all nodes in the digraph are balanced then the digraph is called balanced
  \end{itemize}
  \item The in-degree matrix $\Delta^D $ of a digraph $D$ is an $n \times n$ diagonal matrix s.t.  $\Delta^D_{ii}=(in-deg)_i$.
   \item The Laplacian associated with the digraph $D$, $L^D$, is defined as
   \begin{equation}\label{L-def}
    L^D=\Delta^D -A^D
   \end{equation}
   where $\Delta^D$ is the in-degree matrix and $A^D$ defined as in (\dref{DirAdj})
   \item The incidence matrix for a digraph can be defined analogously to (\dref{def-incidence}) by skipping the pre-orientation that is needed for undirected graphs.
 \item Connectedness in digraphs:
 \begin{itemize}
   \item A digraph is called strongly connected if for every pair of vertices there is a directed path between them.
   \item The digraph is called weakly connected if it is connected when viewed as a graph, that is, a disoriented digraph.
 \end{itemize}
\end{itemize}
\textbf{Definition:}\\
\begin{itemize}
  \item A digraph has a rooted out-branching if there exists a vertex $r$ (the root) such that for every other vertex $i \neq r \in V$ there is a directed path from $r$ to $i$. In this case, every  $i \neq r \in V$ is said to be reachable from $r$.
  \item In strongly connected digraphs each node is a root.
\end{itemize}

\subsubsection{Properties of Laplacian matrices associated with digraphs}

\TB{T-PropertiesDirL}
 The Laplacian associated with a strongly connected digraph $D$ of order $n$, denoted by $L^D$ and defined as in (\dref{L-def}) has the following properties:
\hB
\titem{a}	$L^D$ has an eigenvalue $\lambda^D_1 = 0$ with an associated right eigenvector of ones, $\mathbf{1}_n$
\titem{b}	 $rank(L^D)=n-1$, i.e. the algebraic multiplicity of the zero eigenvalue is 1.
 \titem{c} The remaining (non-zero) eigenvalues of $L^D$ have a strictly positive real part
 \titem{d} The left eigenvector of $L^D$ corresponding to $\lambda^D_1=0$ is
$\mathbb{1}_n^T$ \emph{if and only if the graph is balanced}
\hE
\TE
\PB\\
Property \tref{a} - by construction, since the sum of each row of L is 0\\
Property \tref{b}-  Lemma 2 in \dcite{OS-M2007}  \\
Property \tref{c} - By Gershgorin's theorem (see Appendix \dref{App-Gersh}),  since $\displaystyle InDegree_i = L^D_{ii} = \sum_{j \neq i}^n L^D_{ij}$ and since every eigenvalue of  $L^D$ must be within a distance $\displaystyle  \sum_{j \neq i}^n L^D_{ij}$ from $L^D_{ii}$ for some $i$  all the eigenvalues of $L^D$ are located in a disk centered at $InDegree_{max}+0j$ in the complex plane, where $InDegree_{max}$ is the maximum in-degree of any node in $D$. Thus, the real part of the eigenvalues of $-L^D$ are non-positive\\
Property \tref{d} -  Proposition 4 in \dcite{OS-M2003}
\PE

\subsubsection{Directed symmetric graph - Scaled influences}\dlabel{ScaledGraph}
A special form of a directed graph is the symmetric, usually non-balanced, graph with scaled influences. In this graph
\begin{itemize}
  \item for each edge entering a node there is an edge exiting the same node (symmetric graph)
   \item the weight of the directed edge $(j,i)$, from $j$ to $i$ was defined as $\sigma_{ji} = \frac{1}{|N_i|}; \quad \forall j \in N_i$,  the scaled influence of $j$ to $i$
    \item the  InDegree does not necessarily equal the OutDegree  for all nodes (non-balanced)
\end{itemize}
Such a digraph will be referred to as a scaled graph and will be denoted by $S$.
Since $S$ is a digraph it inherits all the properties presented in Section \dref{Digraph} but has some additional ones, stemming from its special structure. The matrices associated with $S$ are:
\begin{itemize}
  \item The adjacency matrix defined as
  \begin{equation}\label{ScaledAdj}
    A^S_{ij}=
    \begin{cases}
\frac{1}{|N_i|} & \text{if } j \in N_i
\\
0 & \text{otherwise }
\\
\end{cases}
  \end{equation}
  \item The InDegree matrix, $\Delta^S = I$
  \item The Laplacian matrix associated with $S$
  \begin{equation}\label{def-Ls1}
    L^S=\Delta^S- A^S=\Delta^{-1} L^U
  \end{equation}
  where $\Delta$ and $L^U$ are respectively the degree matrix and the Laplacian associated with the undirected graph $G^U$ corresponding to the scaled graph $S$.

\end{itemize}

\LB{L-scaledEigen}
Let $\lambda_i^S$ be the $i-th$ eigenvalue of the non-symmetric $L^S$ and $V_i^S$ be a  corresponding right eigenvector, namely
\begin{equation}\dlabel{LD-eigen}
  L^S V_i^S= \lambda_i^S V_i^S = \Delta^{-1}L^U V_i^S
\end{equation}. Then
\begin{description}
\item[a)] $\lambda_i^S$ is also an eigenvalue of the symmetric, normalized Laplacian, $\Gamma$,  defined by (\dref{normalizedL}) of the corresponding undirected graph
\item[b)] $V_i^\Gamma = \Delta^{1/2} V_i^S$ is an eigenvector of $\Gamma$, associated with $\lambda_i^S$
\item[c)] The eigenvectors of $L^S$ span the entire space $\mathbb{R}^n$.
\end{description}
\LE
\PB
After some simple algebra, eq. (\dref{LD-eigen}) leads to
\begin{equation}\dlabel{xxxx}
  \left (\Delta^{-1/2}L^U \Delta^{-1/2}\right) \left (\Delta^{1/2} V_i^S \right)= \lambda_i^S   \left ( \Delta^{1/2} V_i^S \right)
\end{equation}
Thus $\lambda_i^S$ is also an eigenvalue of the symmetric matrix $\Gamma=\Delta^{-1/2}L^U V^{-1/2}$, the normalized Laplacian, and  $V_i^\Gamma = \Delta^{1/2} V_i^S$ is a corresponding right eigenvector.
From this we conclude that:
\begin{enumerate}
  \item All eigenvalues of $L^S$ are real and non-negative
  \item The algebraic multiplicity of all eigenvalues of $L^S$ equals their geometric multiplicity
  \item The eigenvectors of $L^S$ corresponding to $\lambda^S$ span the same subspace of $\mathbb{R}^n$ as the eigenvectors of $\Gamma$ corresponding to the same $\lambda^S$ (since $V^\Gamma= \Delta^{1/2} V^S$ and $\Delta^{1/2}$  non-singular)
   \item   The collection of the eigenvectors of $L^S$ corresponding to \emph{all} its eigenvalues spans the entire $\mathbb{R}^n$

\end{enumerate}
\PE

From Lemma \dref{L-scaledEigen} follows that  $L^S$ is diagonizable, i.e. can be written as
  \begin{equation}\dlabel{LsDecompose}
    L^S=V^S \Lambda^S (V^S)^{-1}
  \end{equation}
  where $\Lambda^S $ is a diagonal matrix consisting of the eigenvalues of $L^S$, which are real but not necessarily distinct, and  $V^S$ is a matrix whose columns are the \emph{normalized} right eigenvectors of $L^S$.  In particular, we note that since the sum of the elements of any row of $L^S$ is 0, the normalized right eigenvector corresponding to $\lambda_S=0$, is $V^S_1=\frac{1}{\sqrt{n}} \mathbf{1}_n$.

\LB{L-Ws1Vs1}

Denote  $(W^S)^T=(V^S)^{-1}$. Then

\begin{itemize}
  \item Each row of $(W^S)^T$ is a left eigenvector of $L^S$
  \item The first row $(W^S_1)^T$ of $(W^S)^T$ is a left eigenvector of $L^S$ corresponding to $\lambda_S=0$ and satisfies  $(W^S_1)^T V^S_1=1$

\end{itemize}
\LE
\PB
 By multiplying (\dref{LsDecompose}) from the left by $(V^S)^{-1}$ and substituting it with $(W^S)^T$, we obtain
 \begin{equation*}
(W^S)^T L^S= \Lambda^S (W^S)^T
  \end{equation*}
namely the rows of $(W^S)^T$ are right eigenvectors of $L^S$.
The relationship $(W^S_1)^T V^S_1=1$ follows immediately from the definition of $(W^S)^T$.
\PE

\TB{T-WsT1}
Denote by $\mathbf{d}$ the vector of degrees of vertices in the undirected graph $G^U$ corresponding to the digraph $S$, i.e. $\mathbf{d}=diag(\Delta)=\Delta \cdot \mathbf{1}_n$. Then $(W^S_1)^T$, is given by
\begin{equation}\dlabel{w1d}
  (W^S_1)^T= \frac{\sqrt{n} \cdot \mathbf{d}^T}{\sum_{i=1}^n d_i}
\end{equation}
where $d_i$ is the $i$-th element of $\mathbf{d}$.
\TE
\PB
The vector $(W^S_1)^T$ is a left eigenvector of $L^S$ associated with $\lambda_1^S = 0$ and similarly to Lemma \dref{L-scaledEigen}
\begin{equation*}
  (W_1^\Gamma)^T= (W_1^S)^T  \Delta^{-1/2}
\end{equation*}
is a left eigenvector of $\Gamma$ associated with the same $\lambda_1^\Gamma = 0$.

Since $\Gamma$ is symmetric, $(W^\Gamma_1)$ is also a \emph{right} eigenvector of $\Gamma$ associated with $\lambda_1^\Gamma = 0$ and thus, from Lemma \dref{L-scaledEigen}, follows that
\begin{equation*}
  (W^\Gamma_1)^T \in span( (V^S_1)^T  \Delta^{1/2})
\end{equation*}
or
\begin{equation*}
  (W^S_1)^T \in span((V^S_1)^T  \Delta) = span(\mathbf{1}_n^T  \Delta) = span(\mathbf{d}^T)
\end{equation*}
where we have used  $V_1^S=\frac{1}{\sqrt{n}} \mathbf{1}_n$.
Recalling that $(W^S_1)^T V_1^S = 1$, we get (\dref{w1d}).

\PE

\newpage
\msec{About matrices}{matrices}
\msubsection{Algebraic and geometric multiplicity of eigenvalues}{EigMultiplicity}

Let $\lambda_i$ be an eigenvalue of an arbitrary $n\times n$ matrix $A$.  The \emph{algebraic multiplicity} of $\lambda_i$ is its multiplicity as a root of the characteristic polynomial $det(A-\lambda I)$, that is, the largest integer $k$ such that $(\lambda - \lambda_i)^k$ divides evenly that polynomial.  The \emph{geometric multiplicity} of an eigenvalue $\lambda_i$ is the dimension of the eigenspace associated to $\lambda_i$, i.e. the number of linearly independent eigenvectors with that eigenvalue.

\LB{L-geom}
The algebraic multiplicity of an eigenvalue is larger than or equal to its geometric multiplicity.
\LE
\PB
From the above, the geometric multiplicity = $n-rank(A-\lambda I)$.
From \dcite{HJbook} Theorem 1.2.18 we have that if the algebraic multiplicity of $\lambda$ is $k$ then $rank(A-\lambda I) \geq n-k$. Thus, the geometric multiplicity is $ \leq n-n+k=k$,
\PE

Let $\mathbf{1}^{\bot}$ denote the subspace of $\mathbf{R}^{n}$ perpendicular to $\mathbf{1}_n$.  Clearly, the subspace $\mathbf{1}^{\bot}$ has dimension $n-1$.

\msubsection{Gershgorin's theorem}{App-Gersh}
This section follows ref.\dcite{Gersh}.
\TB{T-GershEig}
Every eigenvalue of a matrix $A_{n \times n}$ satisfies
\begin{equation*}
  |\lambda-A_{ii}| \leq \sum_{j\neq i} |A_{ij}|; \text{    } i \in \{1,2,....,n\}
\end{equation*}
\TE
For proof see ref.\dcite{Gersh}. \\
In analyzing this theorem we see that every eigenvalue of the matrix $A$ must be
within a distance $d$ of $A_{ii}$ for some $i$. Since in general eigenvalues are elements of $\mathcal{C}$, we can visualize an eigenvalue as a point in the complex plane, where that point has to be
within distance $d$ of $A_{ii}$ for some $i$.\\
\textbf{Definition} - Gershgorin's disc\\
Let $d_i= \sum_{j\neq i} |A_{ij}|$.
Then the set $D_i=\{z \in \mathcal{C}:|z-A_{ii}|\leq d_i$ is called the $i'th$ Gershgorian disc of $A$. This disc is the interior plus the boundary of a circle with radius $d_i$,centered at $A_{ii}$. Thus, for a matrix $A_{n \times n}$ there are $n$ discs in the complex plane, each centered on one of the diagonal entries of the matrix $A$. Theorem \dref{T-GershEig} implies that every eigenvalue must lie within one of these discs. However it does not say that within each disc there is an eigenvalue.\\
\textbf{Definition} - Disjoint discs\\
A Subset G of the Gershgorin discs is called a disjoint group of discs if no disc in the
group G intersects a disc which is not in G.
\TB{Eig-Disjoint}
If a matrix $A_{n \times n}$  has a disjoint Gershgorin disc, $P$, created from a row with a real diagonal element then the eigenvalue within disc $P$ is real.
\TE

\msubsection{Positive and non-negative matrices}{Pos-nonnegative}
If $A=[a_{ij}] \in \mathbb{R}^{n \times m}$ and $B=[b_{ij}] \in \mathbb{R}^{n \times m}$ then we denote
\begin{eqnarray*}
  A &\geq & 0 \quad \text{if all} \quad a_{ij}\geq 0 \quad \text{and} \quad A >  0 \quad \text{if all} \quad a_{ij}> 0 \\
  A & \geq & B \quad \text{if} \quad A-B  \geq 0 \quad \text{and} \quad A  >  B \quad \text{if} \quad A-B  > 0
\end{eqnarray*}
If $A\geq 0$ we say that $A$ is a non-negative matrix and if $A > 0$ we say that $A$ is a positive matrix. A non-negative matrix such that its rows sum to one is called a stochastic matrix.

One of the cornerstones of the theory of nonnegative matrices is the Perron-Frobenius theorem. The main part of the Perron-Frobenius theorem is summarized next (Theorem 8.2.8 in \dcite{HJbook}):
\TB{T-Perron}
Let $A>0$. Then
\hB
\titem{a} $\rho(A) >0$ where $\rho(A)$ is the spectral radius of $A$
\titem{b} $\rho(A)$ is an algebraically simple eigenvalue of $A$
\titem{c}there is a unique real vector $x$ such that $Ax=\rho x$ and $x_1+x_2+....x_n=1$; this vector is positive
\titem{d} there is a unique real vector $y$ such that $y^T A=\rho y^T$ and $x^T y=1$; this vector is positive
\titem{e} $\|\lambda\|<\rho(A)$ for every eigenvalue of $A$ such that $\lambda \neq \rho(A)$
\titem{f} ($\rho(A)^{-1} A)^m \rightarrow x y^T \quad \text{as} \quad m \rightarrow \infty$
\hE
\TE
Perron's theorem is generalized to non-negative matrices with the additional condition of irreducibility (see definition and properties in section \dref{RedMat} of this Appendix), Theorem 8.4.4 in \dcite{HJbook}.

\TB{T-PerronFrob} Let $A_{n \times n}\quad n\geq 2$ be non-negative and irreducible. Then
\hB
\titem{a} $\rho(A) >0$
\titem{b} $\rho(A)$ is an algebraically simple eigenvalue of $A$
\titem{c}there is a unique real vector $x$ such that $Ax=\rho x$ and $x_1+x_2+....x_n=1$; this vector is positive
\titem{d} there is a unique real vector $y$ such that $y^T A=\rho y^T$ and $x^T y=1$; this vector is positive
\hE
\TE

\msubsection{Reducible and irreducible matrices}{RedMat}
\textbf{Irreducible Matrix} – A matrix, $A$, is irreducible (or ergodic) if any state can be reached from any other state in a finite number of time steps.\\
\textbf{Reducible Matrix} – A matrix, $A$, is reducible if it is not possible to reach all states of the model from all other states in a finite number of time steps. \\

A positive matrix is automatically irreducible, because every state can be reached after only one time step.  The same reasoning applies to any matrix that is positive when taken to some power (see “Primitive Matrix”). \\
\textbf{Primitive Matrix} – A matrix, $A$, is primitive (or aperiodic) if all elements of the matrix are simultaneously positive when the matrix is raised to a high enough power. Primitive matrices are always irreducible.

\newpage
\msec{Eigenvalues based square matrix decomposition}{App-decomp}
Following \dcite{HJbook}, let $\mathbf{M}_n$ denote the class of all $n \times n$ matrices.
Then
\begin{itemize}
  \item Any matrix $A \in \mathbf{M}_n$ is similar to an essentially unique  Jordan matrix,i.e. \begin{itemize}
        \item there exists a matrix $T$ such that $J=T^{-1}A T$ where $J$ is the Jordan canonical form
        \item Jordan canonical form is a block diagonal matrix defined by
        \begin{equation}\label{JordanForm}
          J= \left (
\begin{matrix}
J_{k_1}(\lambda_1) &  0  & 0 &\ldots & 0\\
0  &  J_{k_2}(\lambda_2) & 0 &\ldots & 0\\
\vdots & \vdots & \ddots & & \vdots\\
0  &   0       &\ldots & 0& J_{k_r}(\lambda_r)
\end{matrix}  \right )
        \end{equation}
        where
        \begin{itemize}
          \item $\lambda_i$ are the eigenvalues of $A$, not necessarily distinct
          \item $J_{k_i}(\lambda_i)$ is a Jordan block associated with $\lambda_i$
          \item  $J_k(\lambda)$ is a matrix of the form
        \begin{equation}\label{JordanBlock}
          J_k(\lambda)= \left (
\begin{matrix}
\lambda &  1  & 0 &\ldots & 0\\
0  &  \lambda & 1 &\ldots & 0\\
\vdots & \vdots & \ddots & & \vdots\\
0  &   0       &\ldots & 0& \lambda
\end{matrix} \right )
        \end{equation}
         \item The number of Jordan blocks associated with an eigenvalue $\lambda_i$ is the geometric multiplicity of $\lambda_i$.
         \item The sum of the dimensions of all Jordan blocks associated with $\lambda_i$ is the algebraic multiplicity of of $\lambda_i$
         \item If every $J_i$ has dimension $1 \times 1$, then A is called diagonalizable, which is true if and only if every eigenvalue has same algebraic and geometric multiplicities.

        \end{itemize}
   \item The Jordan matrix $J$ is uniquely determined by $A$ up to permutation of its blocks.
   \item If $A$ is real and has only real eigenvalues, then $T$ can be chosen to be real.

      \end{itemize}
 \item  If $A \in \mathbf{M}_n$ with distinct eigenvectors (not necessarily distinct eigenvalues) then $A=V \Lambda V^{-1}$, where $\Lambda$ is a diagonal matrix formed from the eigenvalues of $A$, and the columns of $V$ are the corresponding eigenvectors of $A$.\\
     A matrix $A \in \mathbf{M}_n$ always has $n$ eigenvalues, which can be ordered (in more than one way) to form a diagonal matrix $\Lambda \in \mathbf{M}_n$ and a corresponding matrix of nonzero columns $V$ that satisfies the eigenvalue equation $AV=V\Lambda$. If the $n$ eigenvectors are distinct then $V$ is invertible, implying the decomposition $A=V \Lambda V^{-1}$.\\
     Comment: The condition of having $n$ distinct eigenvalues is sufficient but not necessary. The necessary and sufficient condition is for each eigenvalue to have geometric multiplicity equal to its algebraic multiplicity.
 \item If $A$ is real-symmetric its $n$  (possibly not distinct) eigenvalues are all real with geometric multiplicity which  equals the algebraic multiplicity.   $V$ is always invertible and can be made to have normalized columns. Then the equation $VV^T=I$ holds, because each eigenvector is orthonormal to the other. Therefore the decomposition (which always exists if $A$ is real-symmetric) reads as: $A=V \Lambda V^T$. This is known as the \emph{the spectral theorem}, or \emph{symmetric eigenvalue decomposition} theorem.
 \end{itemize}

\newpage
\msec{Matrix exponential}{exp-M}
Definition: The matrix exponential $e^{\mathbf{A}}$ of $\mathbf{A}$ is the series
\begin{equation}\label{expMseries}
  e^{\mathbf{A}}= \mathbf{I}+\sum_{k=1}^\infty \frac{\mathbf{A}^k}{k!}
\end{equation}
\msubsection{Properties}{expM-Properties}
\begin{enumerate}
  \item Property \dref{Diff-expM} is proven by term by term differentiation of the series and factoring out $\mathbb{A}$.
  \begin{equation}\label{Diff-expM}
  \frac{d}{dt}e^{\mathbf{A}t}= \mathbf{A}e^{\mathbf{A}t}=e^{\mathbf{A}t}\mathbf{A}
\end{equation}
i.e $\mathbf{A}$ and $ e^{\mathbf{At}}$ commute.
  \item If $\mathbf{A}$ and $\mathbf{B}$ commute then
  \begin{equation*}
    e^{\mathbf{A}+\mathbf{B}}= e^{\mathbf{A}}e^{\mathbf{B}}
  \end{equation*}
  In particular $\mathbf{A}$ and $\mathbf{B}$  commute if one of them is a scalar matrix, i.e. it has the form of the $\beta \mathbf{I}$
  \item For any $\beta \in \mathbb{R}$ and $\mathbf{v} \in \mathbb{R}^n$
 \begin{equation*}
   e^{\beta \mathbf{I}} \mathbf{v}=  e^{\beta}\mathbf{v}
 \end{equation*}
 The proof follows from the definition.
 \item If $\lambda$  is an eigenvalue of matrix $A$ with eigenvector $\mathbf{v}$ then
 \begin{equation*}
   e^{\mathbf{A}t} \mathbf{v}=e^{\lambda t} \mathbf{v}
 \end{equation*}
 \PB
\begin{eqnarray*}
  \mathbf{A}t &=& \lambda \mathbf{I}+ \mathbf{A}t-\lambda \mathbf{I} \\
  e^{\mathbf{A}t} \mathbf{v}&=& e^{(\lambda \mathbf{I}t+ \mathbf{A}t-\lambda \mathbf{I}t)} \mathbf{v} \\
  &=& e^{\lambda t} e^{( \mathbf{A}-\lambda \mathbf{I})t}  \mathbf{v}\\
  &=& e^{\lambda t} \left (\mathbf{I}+ ( \mathbf{A}-\lambda \mathbf{I})t + \frac{( \mathbf{A}-\lambda \mathbf{I})^2 t^2}{2!}+.... \right ) \mathbf{v}\\
   &=& e^{\lambda t} \left (\mathbf{I}+ 0+0+.....\right ) \mathbf{v}\\
   &=& e^{\lambda t} \mathbf{v}
\end{eqnarray*}
 \PE
\end{enumerate}

\msubsection{Generalized matrix exponential}{Gen-expM}
The existence of the Jordan form allows us to generalize the matrix exponential in (\dref{expMseries}). \\
Note: A diagonal matrix is a special case of Jordan form.
\begin{equation}\label{powerA}
 A^k=(T A T^{-1})^k = T J^k T^{-1}
\end{equation}
where
\begin{equation}\label{powerJ}
          J^k= \left (
\begin{matrix}
 J_1^k  \\
  &  J_2^k\\
 &  & \ddots & \\
& & && J_r^k
\end{matrix} \right )
\end{equation}

and if $J_i$ is an $s \times s$  Jordan block associated with $\lambda_j$ then
 \begin{equation}\label{powerJordanBlock}
          J_i^k= \left (
\begin{matrix}
\lambda_j^k &  k \lambda_j^{k-1}  & k^2 \lambda_j^{k-2} &\ldots & k^{s-1} \lambda_j^{k-s-1}\\
0  & \lambda_j^k & k \lambda_j^{k-1} & k^2 \lambda_j^{k-2} &\ldots & \\
0 & 0 & \lambda_j^k & k \lambda_j^{k-1}  &\ldots & \\
\vdots & \vdots & \ddots & \ddots &\\
0  &  0    &\ldots & 0 & \lambda_j^k
\end{matrix} \right )
        \end{equation}

Thus
\begin{equation}\label{Gen-expAt}
  e^{tA}=\sum_{k=0}^\infty \frac{t^k A^k}{k!} = Te^{tJ}T^{-1}
\end{equation}
where
\begin{equation}\label{expJt}
          e^{Jt}= \left (
\begin{matrix}
 e^{J_1 t}  \\
  &  e^{J_2 t}\\
 &  & \ddots & \\
& & && e^{J_r t}
\end{matrix} \right )
\end{equation}
and
\begin{equation}\label{expJordanBlock}
          e^{J_i t}= \left (
\begin{matrix}
e^{\lambda_j t} &  t e^{\lambda_j t} & \frac{t^2}{2!} e^{\lambda_j t} &\ldots & \frac{t^{s-1}}{(s-1)!} e^{\lambda_j t}\\
0  & e^{\lambda_j t} & t e^{\lambda_j t} & \frac{t^2}{2!} e^{\lambda_j t} &\ldots & \\
0 & 0 & e^{\lambda_j t} & t e^{\lambda_j t}   &\ldots & \\
\vdots & \vdots & \ddots & \ddots &\\
0  &  0    &\ldots & 0 & e^{\lambda_j t}
\end{matrix} \right )
        \end{equation}

\newpage
\msec{Incomplete graph - case of complete subgraphs of followers and leaders }{incomplete}
 We consider the case when the subset of leaders and the subset of followers initially form complete subgraphs, but not all leaders are connected to all followers.  A leader can be connected to more than one follower and a follower can be connected to more than one leader, as illustrated in Fig. \dref{fig-case3}.\\
 \begin{figure}[H]
\begin{center}
\includegraphics[scale=0.6]{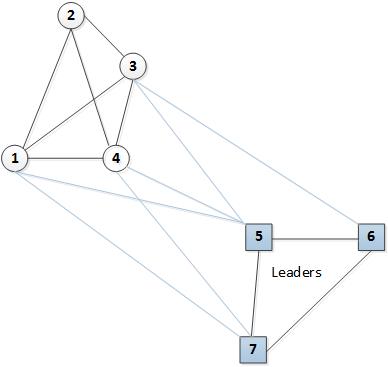}
\caption{Illustration of considered incomplete graph}\label{fig-case3}
\end{center}
\end{figure}

We employ the following notations:
\begin{itemize}
   \item $\mathbb{N}^f$ denotes the set of followers
   \item $\mathbb{N}^l$ denotes the set of leaders
   \item $n_f = |\mathbb{N}^f|$  is the number of followers
   \item $n_l= |\mathbb{N}^l|$ is  the number of leaders
   \item $n = n_f+n_l$ is the total number of agents
   \item $\mathbb{N}_i^f$ denotes the set of followers adjacent to agent $i$
   \item $\mathbb{N}_i^l$ denotes the set of leaders adjacent to agent $i$
   \item $n_{il}$ is the number of leaders agent $i$ is connected to
   \item $n_{if}$ is the number of followers agent $i$ is connected to
   \item $\mathbb{N}_{(ij)}^l$ denotes the set of leaders adjacent to both $i$ and $j$
   \item $n_{(ij)l}$ is the number of leaders that have a link to both $i$ and $j$
   \item $\mathbb{N}_{(ij)}^f$ denotes the set of followers adjacent to both $i$ and $j$
   \item $n_{(ij)f}$ is the number of followers that have a link to both $i$ and $j$
   \item $\mathbb{N}_i$ is the neighborhood of $i$, $\mathbb{N}_i = \mathbb{N}_i^f \bigcup \mathbb{N}_i^l$
   \item $n_i$ is the size of the neighborhood of $i$, $n_i = n_{il} + n_{if}$
\end{itemize}

\msubsection{Uniform influence}{UniSpecialCase}
The followers dynamics can then be written as
\begin{equation}\dlabel{ML-FolDyn}
  \dot{p}_i=-\sum_{k \in \mathbb{N}^f} (p_i-p_k) - \sum_{k \in \mathbb{N}_i^l} (p_i-p_k); \quad i \in \mathbb{N}^f
\end{equation}
while the leaders dynamics can be written as
\begin{equation}\dlabel{ML-LeadDyn}
  \dot{p}_i=-\sum_{k \in \mathbb{N}^l} (p_i-p_k) - \sum_{k \in \mathbb{N}_i^f} (p_i-p_k) + u; \quad i \in \mathbb{N}^l
\end{equation}

\TB{Incomplete}
With the above model, suppose the following hold:
\begin{enumerate}
  \item for every pair of followers $i$ and $j$ that are initially neighbors holds
  \begin{equation}\dlabel{F2F-connect}
    n_{il}+n_{jl} \leq n_f + 3  n_{(ij)l}
  \end{equation}
  \item for every pair of leaders $i$ and $j$ that are initially neighbors holds
  \begin{equation}\dlabel{L2L-connect}
    n_{if}+n_{jf} \leq n_l + 3  n_{(ij)f}
  \end{equation}
  \item for any follower $i$ that is initially adjacent to a leader $j$ holds
  \begin{equation}\dlabel{L2F-preserve}
    n_{il}+n_{jf} - n/2 > 0
  \end{equation}
\end{enumerate}
Then an input $u$ that satisfies
\begin{equation}\dlabel{u-connect}
  |u| \leq 2R \cdot(n_{il}+n_{jf} - n/2)
\end{equation}
ensures the property of \emph{never lose neighbors}.
\TE

\PB
We shall prove the Theorem by contradiction.  Suppose the Theorem is not true, and consider \emph{the event when the statement of the theorem is contradicted  \textbf{for the first time}}.  Let $(i,j)$ be the link where this happens, namely $(i,j)$ reaches distance $R$, while all other connections are less than or equal to $R$ and $d(\delta_{ij}^2) /dt> 0$.  We shall prove that this is a contradiction.
\begin{itemize}
  \item \textbf{Suppose $i$ and $j$ are both followers}.
  Both $i$ and $j$ have links to all other followers.  Also, recalling our notations,$i$ and $j$ have links to leaders in sets $\mathbb{N}_i^l, \mathbb{N}_j^l$ respectively and
\begin{eqnarray*}
  \mathbb{N}_{(ij)}^l &=& \mathbb{N}_i^l\cap \mathbb{N}_j^l\subseteq \mathbb{N}^l \\
  n_{(ij)l}&=& |\mathbb{N}_{(ij)}^l|
\end{eqnarray*}
We have $\delta_{ij}=|p_i-p_j|=R$. Since this is the first time a neighbor is lost, all other connections are less than or equal $R$.  Recall that $\frac{d}{dt} \delta_{ij}^2 = 2(p_i-p_j)^T(\dot{p}_i-\dot{p}_j)$.  We have
\begin{equation*}
  \dot{p}_i-\dot{p}_j = -\sum_{k \in \mathbb{N}^f} (p_i-p_k) - \sum_{k \in \mathbb{N}_i^l} (p_i-p_k) + \sum_{k \in \mathbb{N}^f} (p_j-p_k) + \sum_{k \in \mathbb{N}_j^l} (p_j-p_k)
\end{equation*}
Substituting
\begin{eqnarray*}
  \mathbb{N}_i^l &=& \mathbb{N}_{(ij)}^l+\mathbb{N}_i^l \backslash \mathbb{N}_{(ij)}^l \\
  \mathbb{N}_j^l &=& \mathbb{N}_{(ij)}^l+\mathbb{N}_j^l \backslash \mathbb{N}_{(ij)}^l
\end{eqnarray*}
one obtains after some algebra
\begin{equation*}
  \dot{p}_i-\dot{p}_j = -(n_f+n_{(ij)l})(p_i-p_j)  - \sum_{k \in \mathbb{N}_i^l \backslash \mathbb{N}_{(ij)}^l  } (p_i-p_k) + \sum_{k \in \mathbb{N}_j^l \backslash \mathbb{N}_{(ij)}^l } (p_j-p_k)
\end{equation*}
Thus,
\begin{eqnarray*}
  \frac{d}{dt} \delta_{ij}^2 &= &-2(n_f+n_{(ij)l})\delta_{ij}^2+2  \left [ \sum_{k \in \mathbb{N}_i^l \backslash \mathbb{N}_{(ij)}^l  }(p_i-p_j)^T(p_k-p_i) + \sum_{k \in \mathbb{N}_j^l \backslash \mathbb{N}_{(ij)}^l }(p_i-p_j)^T (p_j-p_k) \right]\\
  & \leq & -2(n_f+n_{(ij)l})\delta_{ij}^2 +2  \left [\sum_{k \in \mathbb{N}_i^l \backslash \mathbb{N}_(ij)^l  }\delta_{ij}|p_k-p_i|) + \sum_{k \in \mathbb{N}_j^l \backslash \mathbb{N}_{(ij)}^l }\delta_{ij} |p_j-p_k| \right]\\
  & \leq & -2(n_f+n_{(ij)l})\delta_{ij}^2 +2  \left [(n_{il}-n_{(ij)l})\delta_{ij}R + (n_{jl}-n_{(ij)l})\delta_{ij}R \right ]
\end{eqnarray*}

Since $\delta_{ij}=R$, we have
\begin{equation*}
  \frac{d}{dt} \delta_{ij}^2 \leq 2 R^2 \left [ -n_f+n_{il}+n_{jl}-3n_{(ij)l} \right ]
\end{equation*}
and condition  (\dref{F2F-connect})  leads to  $\frac{d}{dt} \delta_{ij}^2 \leq 0$ .

  \item \textbf{Suppose $i$ and $j$ are both leaders}

The leaders dynamics is given by eq. (\dref{ML-LeadDyn}).  Both $i$ and $j$ have links to all other leaders and links.  They have links to followers in sets $\mathbb{N}_i^f, \mathbb{N}_j^f$ respectively.  Recalling our notations,
\begin{eqnarray*}
  \mathbb{N}_{(ij)}^f &=& \mathbb{N}_i^f\cap \mathbb{N}_j^f\subseteq \mathbb{N}^f \\
  n_{(ij)f}&=& |\mathbb{N}_{(ij)}^f|
\end{eqnarray*}
Following the same technique as in the previous section, we obtain that since $\delta_{ij}=R$
\begin{equation*}
  \frac{d}{dt} \delta_{ij}^2 \leq 2 R^2 \left [ -n_l+n_{if}+n_{jf}-3n_{(ij)f} \right ]
\end{equation*}
and condition (\dref{L2L-connect})  leads to  $\frac{d}{dt} \delta_{ij}^2 \leq 0$.

  \item \textbf{Suppose $i$ is a follower and $j$ is a leader}.

Then $i$ follows dynamics (\dref{ML-FolDyn}) and $j$ follows dynamics (\dref{ML-LeadDyn}).
\begin{eqnarray*}
  \frac{d}{dt} \delta_{ij}^2 & = & 2(p_i-p_j)^T \left ( -\sum_{k \in \mathbb{N}^f} (p_i-p_k) - \sum_{k \in \mathbb{N}_i^l} (p_i-p_k) + \sum_{k \in \mathbb{N}^l} (p_j-p_k) + \sum_{k \in \mathbb{N}_j^f} (p_j-p_k) -u \right)\\
  & = & 2(p_i-p_j)^T \left ( a + b -u  \right)
\end{eqnarray*}
where
\begin{eqnarray*}
  a &=&  -\sum_{k \in \mathbb{N}_j^f} (p_i-p_k)- \sum_{k \in \mathbb{N}^f \backslash \mathbb{N}_j^f } (p_i-p_k) - \sum_{k \in \mathbb{N}_i^l} (p_i-p_k)  \\
  b &=&  \sum_{k \in \mathbb{N}_j^l} (p_j-p_k)+ \sum_{k \in \mathbb{N}^l \backslash \mathbb{N}_j^l} (p_j-p_k) + \sum_{k \in \mathbb{N}_j^f} (p_j-p_k)
\end{eqnarray*}
After some more calculations we obtain
\begin{eqnarray*}
  \frac{d}{dt} \delta_{ij}^2 & = &-2(n_{jf}+n_{il})\delta_{ij}^2+ 2(p_i-p_j)^T \left ( - \sum_{k \in \mathbb{N}^f \backslash \mathbb{N}_j^f } (p_i-p_k) + \sum_{k \in \mathbb{N}^l \backslash \mathbb{N}_i^l} (p_j-p_k) -u  \right )\\
  & \leq & -2(n_{jf}+n_{il})\delta_{ij}^2+ 2 \left ( \sum_{k \in \mathbb{N}^f \backslash \mathbb{N}_j^f }\delta_{ij} |(p_i-p_k)| + \sum_{k \in \mathbb{N}^l \backslash \mathbb{N}_i^l} \delta_{ij}|(p_j-p_k)| + \delta_{ij} |u|  \right )\\
   & \leq & -2(n_{jf}+n_{il})R^2+2(n_f-n_{jf})R^2+2(n_l-n_{il})R^2+2R|u|\\
   & \leq & 2R \left [ -2 \left ( n_{jf}+n_{il}-\frac{n}{2} \right)R +|u| \right ]
\end{eqnarray*}
where
\begin{eqnarray*}
  n_{jf} &=& |\mathbb{N}_j^f| \\
  n_{il} &=&  |\mathbb{N}_i^l|
\end{eqnarray*}

If conditions \dref{L2F-preserve} and (\dref{u-connect}) hold, then obviously  $\frac{d}{dt} \delta_{ij}^2 \leq 0$

\end{itemize}

Thus, if conditions (\dref{F2F-connect}) to (\dref{u-connect}) hold, then all initial neighbors are preserved.

\PE

\msec{Effect of edge addition}{EdgeAdd}
Adding edges to a graph preserves its connectedness and increases its density. Denote $ G_1 = G+e$, where $e$ is the edge added to $G$. If we denote by $\lambda_i(G)$ the eigenvalues of the Laplacian corresponding to $G$,  we have (ref. Appendix \dref{interlaced})
\begin{itemize}
\item  \textbf{for uniform influence}

\begin{equation*}
  \lambda^U_n(G_1) \geq \lambda^U_n(G) \geq \lambda^U_{n-1}(G_1) \geq .......\geq \lambda^U_2(G_1) \geq \lambda^U_2(G) > \lambda^U_1(G_1)= \lambda^U_1(G)=0
\end{equation*}
   \begin{equation*}
  \sum_{i=1}^n \lambda^U_i(G_1) = 2+\sum_{i=1}^n \lambda^U_i(G)
\end{equation*}
 Thus, the algebraic connectivity, $\lambda_2^U$ is non-decreasing as edges are added to the graph.
   \item for a graph with \textbf{scaled influence}, the eigenvalues of the Laplacian are not necessarily non-decreasing when links are added. Instead, the eigenvalues satisfy:
  \begin{itemize}
    \item For any strongly connected graph, $G^S$, we have
    \begin{equation*}
  0=\lambda_1(G^S) < \lambda_2(G^S)  \leq ....\leq \lambda_n(G^S) \leq 2
\end{equation*}
\begin{equation*}
  \sum_{i=1}^n \lambda_i(G^S) = \sum_{i=1}^n \lambda_i(G^S_1)=n
\end{equation*}
\item If $G_1^S, G^S$ is are graphs with scaled influence corresponding to $G_1^U, G^U$, then the eigenvalues $\lambda_i(G^S)$ do \textbf{not} simply interlace the eigenvalues of $\lambda_i(G^S_1)$. Instead the following relationship holds:
        \begin{equation*}
  \lambda_{i-1}(G^S) \leq \lambda_{i}(G_1^S) \leq  \lambda_{i+1}(G^S); \quad i=1,...,n
\end{equation*}
where we use $\lambda_0(G^S)=0$ and $\lambda_{n+1}(G^S)=2$
 \end{itemize}
  \end{itemize}

\newpage

\end{document}